%% file: iclr2027_conference_rebuttal.tex
\documentclass{article} %
\usepackage[x11names,dvipsnames]{xcolor}
\usepackage{iclr2027_conference,times}

\usepackage{url}
\usepackage{fontawesome5}

\usepackage[utf8]{inputenc} %
\usepackage[T1]{fontenc}    %
\usepackage[backref=page]{hyperref}       %
\usepackage{url}            %
\usepackage{booktabs}       %
\usepackage{threeparttable}
\usepackage{amsfonts}       %
\usepackage{nicefrac}       %
\usepackage{microtype}      %
\usepackage{xcolor}         %

\usepackage[utf8]{inputenc} %
\usepackage[T1]{fontenc}    %
\usepackage{hyperref}       %
\usepackage{hyperref}       %
\usepackage{url}            %
\usepackage{booktabs}       %
\usepackage{enumitem}
\usepackage{mathtools}
\usepackage{todonotes}
\usepackage{amssymb,fge}
\usepackage{thm-restate}
\usepackage{wrapfig}
\usepackage{bbm}
\usepackage{mathrsfs}
\usepackage{soul}
\usepackage{array}
\usepackage{multirow}
\usepackage{algorithm}
\usepackage{tikz}
\usetikzlibrary{arrows.meta}\usepackage{subcaption}
\usepackage{wrapfig}
\usetikzlibrary{positioning,arrows.meta,shapes.geometric}

\usepackage{natbib}
\usepackage{nicefrac}       %
\usepackage{microtype}      %
\usepackage{xcolor}         %
\usepackage{colortbl}
\usepackage{enumitem}
\usepackage{xspace}
\usepackage{lineno}

\PassOptionsToPackage{final,pagebackref=true}{hyperref}
\usepackage{hyperref} %
\usepackage{wasysym}
\usepackage{acronym}
\usepackage{xfrac}
\usepackage{amsmath, amsthm, mathtools}

\usepackage[nameinlink]{cleveref}
\usepackage[normalem]{ulem}
\usepackage{changepage}
\usepackage{float}
\usepackage{makecell}
\usepackage{pifont}
\usepackage{tikz}
\usepackage{empheq}
\usepackage{booktabs}

\usepackage{svg}
\usepackage{centernot}

\Crefname{assumption}{Assumption}{Assumptions}

\definecolor{sectiongray}{gray}{0.90}
\definecolor{refgray}{gray}{0.55}
\usepackage{algorithm}
\usepackage{graphicx}
\usepackage{tabularx}

\usepackage{subcaption}

\makeatletter
\newcommand{\finishwrap}{%
  \par
  \ifnum\c@WF@wrappedlines>1
    \addtocounter{WF@wrappedlines}{-1}%
    \vskip\value{WF@wrappedlines}\baselineskip
    \global\c@WF@wrappedlines=0
    \WF@finale
  \fi
}
\makeatother

\newcommand{\pmt}[2]{\ensuremath{#1\,{\scriptstyle\pm\,#2}}}
\newcommand{\bpmt}[2]{%
  \ensuremath{\mathbf{#1}\,{\scriptstyle\boldsymbol{\pm}\,\mathbf{#2}}}%
}

\definecolor{rewardcell}{HTML}{DDEFF7}

\usepackage{algpseudocode}
\usepackage{tcolorbox}
\usepackage{titletoc}
\theoremstyle{plain}

\hypersetup{
	colorlinks=true,       %
	linkcolor=RoyalBlue,        %
	citecolor=RoyalBlue,        %
	filecolor=magenta,     %
	urlcolor=RoyalBlue         
}

\graphicspath{{figures/}}
\input{math_commands}

\setcitestyle{round}

\title{Simple Endpoint Reward Finetuning}

\author{%
Romeo Passaro$^{1,2}$\thanks{Equal Contribution.}
\hspace{0.01em}
\thanks{Correspondence to \texttt{\{rpassaro,fruscio\}@aithyra.at}, and \texttt{ishan.nath@stats.ox.ac.uk}.}
\hspace{0.01em},
Francesco M. Ruscio $^{2,3,4 *\dagger}$,
Ishan Nath$^{1 *\dagger}$,
Jeongjae Lee$^5$,
\\
\textbf{Michael M. Bronstein$^{1,2}$,} \textbf{Avishek Joey Bose$^{3,6\dagger}$,} \textbf{Alexander Tong$^2$}\\
$^1$University of Oxford, $^2$AITHYRA, $^3$Imperial College London, $^4$TU Wien, \\ $^5$KAIST, $^6$Mila -- Qu\'ebec AI Institute\\[3pt]
{\small\normalfont
\faGithub \enspace
\href{https://github.com/francescoshox/simple-endpoint-reward-finetuning}{\texttt{https://github.com/francescoshox/simple-endpoint-reward-finetuning}}
}
}

\iclrfinalcopy %
\begin{document}

\maketitle

\begin{abstract}
\looseness=-1
Reward fine-tuning adapts pretrained diffusion and flow models to meet downstream objectives by optimizing a specified reward. Rewards are evaluated on generated samples, while fine-tuning must improve the dynamics at intermediate noisy states. Propagating reward gradients backward through sampling trajectories provides a training signal at these states, but incurs substantial computational cost. Existing first-order alternatives reduce this cost through approximations, often using few-step sample predictions and evaluating reward gradients off-policy. To address these limitations, we introduce \namelong (\nameshort), a simple regression objective that turns endpoint reward gradients into training targets at noisy states. \nameshort generates a sample, computes its reward gradient, and independently re-noises the sample to obtain intermediate training states, where the model’s velocity is regressed toward the reference velocity shifted by the endpoint reward gradient.  We motivate this simple objective through stochastic optimal control, replacing costly gradient transport with a principled approximation to the true reward gradient direction. Each endpoint reward gradient can supervise multiple independently re-noised states, decoupling endpoint generation from optimization. When post-training Stable Diffusion 3.5 Medium on PickScore and HPSv2.1, \nameshort achieves favorable trade-offs among reward improvement, image quality, and output diversity compared with existing methods. \nameshort is shown to be highly efficient, substantially improving training efficiency relative to other first order finetuning methods while remaining effective with as few as five sampling steps per training rollout.
\end{abstract}

\section{Introduction}

Diffusion and flow-matching models~\citep{ho2020denoising,song2021scorebased,lipman2023flowmatching,esser2024scaling,peluchetti2021,albergo2025stochastic,liu2022flow} have become a dominant paradigm for generative modeling in continuous domains, powering modern text-to-image (T2I) generation~\citep{rombach2022high,esser2024scaling}, video generation~\citep{ho2022video,blattmann2023align}, and other multimodal applications~\citep{liu2023audioldm, bao2023one}. These models learn dynamics that transport a simple source distribution to the data distribution, and generate samples by numerically simulating the learned dynamics~\citep{song2021scorebased,lipman2023flowmatching}.

\looseness=-1
Pretraining large generative models, including language and flow models, typically focuses on accurately modeling the data distribution rather than directly optimizing the performance of downstream
tasks~\citep{brown2020languagemodelsfewshotlearners, lipman2023flowmatching,esser2024scaling}.
Deployed models are therefore often \emph{reward-finetuned} to improve task-specific performance, such as prompt adherence, or alignment with human preferences~\citep{ouyang2022training,xu2023imagereward}.

\looseness=-1
For diffusion and flow models, rewards are typically defined on clean samples~\citep{clark2024directly}, yet fine-tuning must improve the
dynamics at intermediate noisy states. When the reward is differentiable, its gradient can be backpropagated through time from the generated samples along the sampling trajectory, providing a training signal at these
states~\citep{clark2024directly,prabhudesai2023aligning}.  However, generating a sample typically requires many neural function evaluations of the learned dynamics~\citep{tong_conditional_2023,geng2025mean,frans2025one}. Backpropagating through the entire trajectory therefore incurs substantial computational and memory costs, and can yield unstable gradients~\citep{bergmeister2026reinforceadjointmatchingscaling,
clark2024directly,prabhudesai2023aligning}. Several reward fine-tuning methods seek to propagate reward information from generated samples to noisy states without the full cost of backpropagation through time (BTT). Some approaches reduce this cost by simplifying gradient propagation or shortening sampling
trajectories~\citep{clark2024directly,prabhudesai2023aligning,wu2024deep,liang2026leapalign}, while others use adjoint techniques derived from stochastic optimal
control (SOC)~\citep{domingoenrich2025adjoint,shin2026efficient}. Additional approaches learn auxiliary networks to estimate intermediate
guidance directions~\citep{liu2025value,jia2025reward}, or construct
intermediate reward signals through local denoising and image
reconstruction~\citep{shen2025directlyaligningdiffusiontrajectory,kang2026diffusion}. Finally, some methods obtain inexpensive estimates of the guidance direction at noisy steps using the single step denoiser predictions, which for images are informative mainly at low-noise states~\citep{xu2023imagereward,tan2026easytune}, and which require separate reward evaluations at different noise levels.

\looseness=-1
Existing methods thus face a trade-off between the cost of propagating reward gradients through the dynamics and the reliability of inexpensive guidance estimates. This raises a natural
question:
\begin{mdframed}[style=MyFrame2]
    \begin{quote}
    Can we formulate a cheap and efficient reward fine-tuning objective that uses only reward gradients at fully generated samples, while still approximating the reward-ascent direction at noisy intermediate states?
\end{quote}
\end{mdframed}
\begin{figure}[t]
    \centering
    \resizebox{\linewidth}{!}{\input{figures/serf_overview-2-2}}
    \caption{\nameshort regresses the velocity at re-noised generated samples towards the reference velocity shifted by the clean-sample reward gradient.}
    \vspace{-15pt}
    \label{fig:reward-anchor-geometry}
\end{figure}
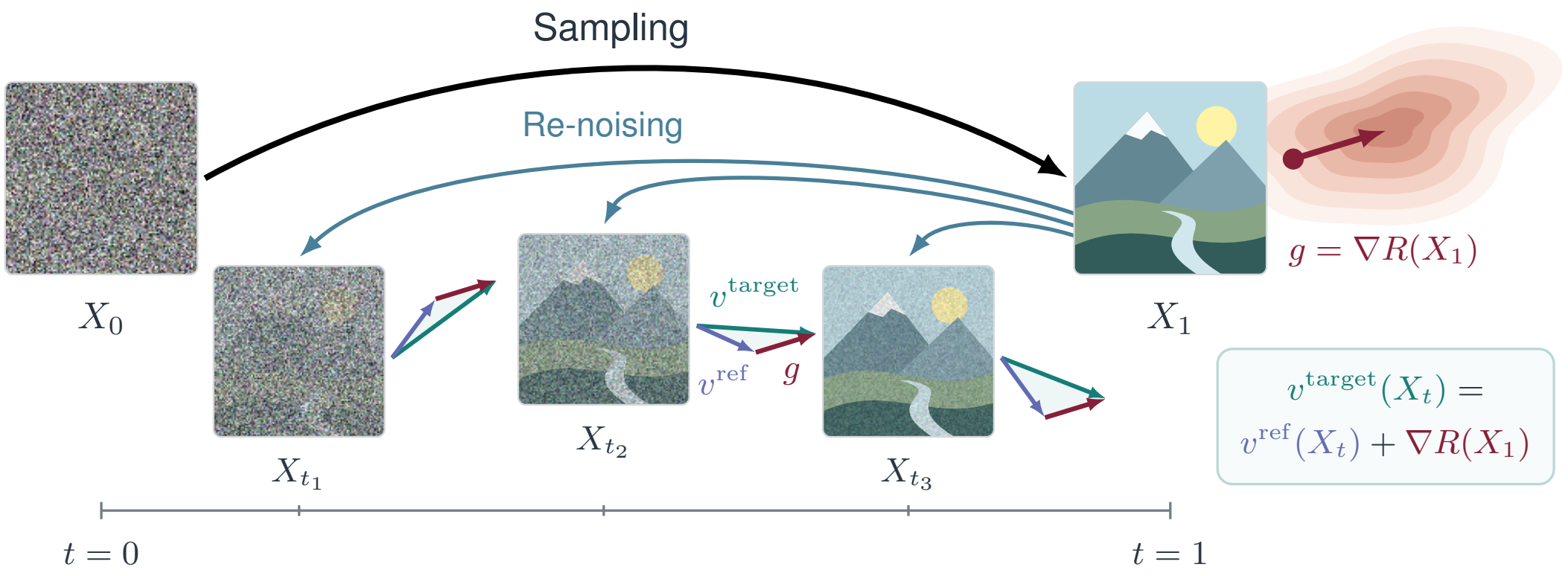

\looseness=-1
To answer this question, we introduce \namelong (\nameshort), a first-order reward fine-tuning method based on a simple regression objective. Rather than explicitly calculating the update direction at a given noisy state, we first generate clean samples with an ODE sampler and compute their reward gradients to define a reward direction. We then independently re-noise each sample using the same noise-corruption procedure used during pretraining. At these noisy states, we regress the model's velocity toward the pretrained reference velocity shifted by the endpoint reward gradient (see \Cref{fig:reward-anchor-geometry}). Reward gradients are thus evaluated on-policy at clean fully generated samples, while optimization requires no BTT. Each endpoint reward gradient can supervise multiple independently re-noised states, amortizing generation and reward evaluation and decoupling the number of training targets from the number of training rollout steps. The objective is agnostic to the choice of endpoint sampler and requires no storage of the sampling trajectory. We derive the objective from stochastic optimal control by approximating the transport of endpoint reward gradients through the dynamics, showing \nameshort arises naturally as an approximation of the reward-ascent direction.

\looseness=-1
We summarize our main contributions as follows:
\begin{itemize}[topsep=0pt, itemsep=0pt, parsep=0pt, leftmargin=*]
    \item \looseness=-1 \textbf{Simple and efficient fine-tuning} We introduce \nameshort, a simple regression objective that uses reward gradients at fully generated samples to fine-tune flow models without BTT. These gradients shift the pretrained velocity to form regression targets at noisy states. By constructing these states through the pretraining corruption procedure, \nameshort decouples training from the sampling trajectory and allows each endpoint reward gradient to be reused across multiple states.
    
    \item \looseness=-1 \textbf{Theoretical analysis.}
    Starting from SOC, we express reward ascent as regression and
    approximate gradient transport to obtain \nameshort's target.
    Under explicit assumptions, we prove positive alignment with the
    exact expected reward gradient and show that independent conditional
    sampling yields better gradient estimator covariance than
    shared-trajectory sampling under the same SDE joint law.
    
    \item \looseness=-1 \textbf{Empirical performance.}
    We evaluate \nameshort by fine-tuning Stable Diffusion~3.5
    Medium~\citep{esser2024scaling} on
    HPSv2.1~\citep{wu2023hpsv2} and
    PickScore~\citep{kirstain2023pickapic}, comparing against
    a comprehensive set of baselines. We find better trade-offs
    between optimized reward, held-out reward scores, image
    fidelity, and diversity, while improving training efficiency
    over competing first-order methods. We further validate \nameshort on GenEval and on JPEG compressibility. Finally, \nameshort remains effective with as few as five ODE sampling steps per training rollout.
\end{itemize}

\section{Background}

\subsection{Flow and diffusion models}
\looseness=-1
Flow matching learns a velocity field that transports Gaussian noise to data~\citep{lipman2023flowmatching,esser2024scaling}. Let $X_0\sim p_0=\mathcal N(0,I)$ and $X_1\sim p_{\mathrm{data}}$ be independent. For the linear interpolation $X_t=tX_1+(1-t)X_0$, with $t\sim p(t)$, the training objective is
\begin{equation}
\mathcal L_{\mathrm{FM}}(v)
=\mathbb E_{t,X_1,X_0}
\left\|v_t(X_t)-(X_1-X_0)\right\|^2.
\label{eq:flow_matching}
\end{equation}
Its minimizer is $v_t^{\mathrm{ref}}(x)=\mathbb E[X_1-X_0\mid X_t=x]$. The ODE $\diff X_t=v_t^{\mathrm{ref}}(X_t)\diff t$, initialized with $X_0\sim p_0$, has the same time marginals as $X_t$ and terminal distribution $p_{\mathrm{data}}$.

\looseness=-1
The reference marginal path also admits an SDE realization~\citep{song2021scorebased,albergo2025stochastic}. We use the memoryless noise schedule of \citet{domingoenrich2025adjoint}, and parameterize its drift directly through the velocity:
\begin{equation}
\diff X_t^v
=\left(2v_t(X_t^v)-\frac{X_t^v}{t}\right)\diff t
+\sigma_t\diff W_t.
\label{eq:velocity_drift_backwards}
\end{equation}
Here $\sigma_t^2=2(1-t)/t$ for $0<t<1$ and $X_0^v\sim p_0$.
For $v=v^{\mathrm{ref}}$, this SDE shares the reference ODE's time marginals
and has independent endpoints. Its backward conditional is the pretraining
corruption kernel,
\[
X_t^{\mathrm{ref}}\mid X_1^{\mathrm{ref}}=x_1
\sim\mathcal N(tx_1,(1-t)^2I),
\]
where $X^{\mathrm{ref}}=X^{v^{\mathrm{ref}}}$ denotes the reference SDE.

\subsection{Reward fine-tuning}
\looseness=-1
Given a pretrained velocity $v^{\mathrm{ref}}$ and a terminal reward $R$,
we seek to increase reward while penalizing departures from the reference.
Following \citet{domingoenrich2025adjoint}, we optimize the stochastic optimal
control (SOC) objective
\begin{equation}
\label{eq:soc_problem}
\mathcal L(v)
=\mathbb E\!\left[
R(X_1^v)-\int_0^1
w(t)\left\|v_t(X_t^v)-v_t^{\mathrm{ref}}(X_t^v)\right\|^2\diff t
\right].
\end{equation}
The process $X^v$ follows \Cref{eq:velocity_drift_backwards} with fixed
initial law and diffusion coefficient. We write $R=\beta R_0$ for a base
reward $R_0$ and strength $\beta>0$.

\looseness=-1
The velocity weight $w(t)=2/\sigma_t^2$ makes the reference penalty the
path-space Kullback--Leibler divergence~\citep{berner2022optimal}.
With this weight, the optimal velocity $v^\star$ induces an endpoint-tilted
path measure with terminal marginal $p_1^{v^\star}(x)\propto
p_{\mathrm{data}}(x)e^{R(x)}$~\citep{domingoenrich2025adjoint}.
The tilt preserves the conditional path law given the endpoint:
\begin{equation}
\operatorname{Law}\!\left((X_t^{v^\star})_{t\in[0,1]}\mid X_1^{v^\star}=x_1\right)
=
\operatorname{Law}\!\left((X_t^{\mathrm{ref}})_{t\in[0,1]}\mid X_1^{\mathrm{ref}}=x_1\right).
\label{eq:same_laws}
\end{equation}
In particular, $X_t^{v^\star}\mid X_1^{v^\star}=x_1$ follows the same corruption kernel as the reference.

\section{Efficient first-order training}
\looseness=-1
We derive the reward-ascent regression from the SOC objective
in \Cref{eq:soc_problem} and use it to motivate \namelong's endpoint
guidance, reference regularization, and independent re-noising.

\looseness=-1
Let $\bar \theta = \mathrm{sg}(\theta)$, where $\mathrm{sg}$ denotes the stop-gradient operator, and let $p_{1}^{\bar\theta}$ be the current ODE endpoint distribution for the frozen velocity $\bar v=v^{\bar\theta}$.
\nameshort first generates a clean sample $X_1\sim p_{1}^{\bar\theta}$.
Its reward gradient $\nabla R(X_1)$ points in a direction that locally
increases its reward. To guide the model at a noisy state $X_t$, we would
ideally account for how a perturbation at $X_t$ propagates to the endpoint.
\nameshort approximates the transport Jacobian by the identity, so the reward correction to the pretrained velocity at $X_t$ is simply $\nabla R(X_1)$. 
We construct $X_t$ by independently re-noising $X_1$, so one endpoint gradient can supervise several noisy states.

This gives the regression loss
\begin{mdframed}[style=MyFrame2]
\setlength{\abovedisplayskip}{2pt}
\setlength{\belowdisplayskip}{2pt}
\begin{align}
\mathcal L_{\nameshort}(\theta; \bar{\theta})
&=
\mathbb E_{t \sim p(t)}
\left\|
v_t^\theta(X_t)-v_t^{\mathrm{ref}}(X_t)
-\nabla R(X_1)
\right\|^2.
\label{eq:SERF_objective}
\end{align}
\end{mdframed}
Here, $X_1 \sim p_{1}^{\bar\theta}$, $t\sim p(t)$, $X_0 \sim \mathcal N(0, I)$, and $X_t = t X_1 + (1 - t) X_0$.

The overall algorithm is summarized in \Cref{alg:serf_algo}.
\Cref{fig:reward-anchor-geometry} details a single \nameshort training step: the reward gradient tilts the velocity while the reference velocity serves as an anchor.

\looseness=-1
\xhdr{Loss decomposition} When $v$ is parameterized by a neural network, directly optimizing
the SOC objective $\mathcal L(v)$ involves costly backpropagation
through the sampled SDE trajectories. Instead, we first decompose the objective into a reward term, \(\mathcal{L}_{\mathrm{reward}}\), and a regularization one \(\mathcal{L}_{\mathrm{regularization}}\), with the aim of deriving
an efficient surrogate for each:
\begin{align}
\label{eq:loss_decomposition}
    \mathcal L(v) = \underbrace{\mathbb E[R(X_1^v)]}_{\mathcal L_{\mathrm{reward}}(v)}
    - \underbrace{\mathbb E \left[ \int_0^1 w(t)
    \left\| v_t(X_t^v) - v_t^{\mathrm{ref}}(X_t^v) \right\|^2 \diff t \right]}_{\mathcal L_{\mathrm{regularization}}(v)}.
\end{align}

\label{sub:decoup_grads}
\xhdr{Reward ascent}
The local ascent direction for $\mathcal L_{\mathrm{reward}}$ can be obtained by differentiating an \(L_2\) loss. Define the expected-reward gradient under the SDE with frozen velocity $\bar v$ by
\[
    G_t^{\bar v}(x)=\nabla_x\mathbb E[R(X_1^{\bar v})\mid X_t^{\bar v}=x].
\]
The reward regression loss is
\begin{align}
    \hat{\mathcal L}_{\mathrm{reward}}(\theta;\bar v)
    = \mathbb E\left[\int_0^1
    \left\|v_t^\theta(X_t^{\bar v})-\bar v_t(X_t^{\bar v})
    -G_t^{\bar v}(X_t^{\bar v})\right\|^2\diff t\right].
    \label{eq:reward_loss}
\end{align}
The derivative of $\bar v$ with respect to $\theta$ is set to $0$, so the sampled states and the regression target are held fixed.

\begin{mdframed}[style=MyFrame2]
\begin{theorem}
\label{thm:reward_surrogate_objective}
    The gradient of
    $\mathcal L_{\mathrm{reward}}(v^\theta)$ with respect to $\theta$ equals
    $-\nabla_\theta\hat{\mathcal L}_{\mathrm{reward}}(\theta;\bar v)$,
    evaluated at the current velocity with $\bar v$ held fixed.
\end{theorem}
\end{mdframed}

We prove this in \S\ref{sub:soc_proofs}. This gives a local reward-ascent update by regressing $v_t^\theta$ in the direction of the gradient of the expected reward. Unfortunately, standard unbiased estimates for the regression target require simulation of the remaining diffusion trajectory and propagation of the endpoint gradient \citep{domingoenrich2025adjoint}.

\begin{algorithm}[t]
\caption{\textbf{Simple Endpoint Reward Finetuning (SERF)}}
\label{alg:serf_algo}
\begin{algorithmic}[1]
\Require Pretrained flow model $v^{\mathrm{ref}}$, reward $R$, outer steps $K$, inner steps $L$, sampling distribution $p(t)$
\State Initialize $v^\theta = v^{\mathrm{ref}}$
\For{$k = 0,\ldots,K-1$}
    \State Sample endpoint minibatch $x_1^i\gets\operatorname{ODESample}(\operatorname{sg}(v^\theta))$
    \State Compute and store reward gradients $g^i = \nabla R(x_1^i)$
    \For{$l = 0, \ldots, L-1$}
        \State Sample intermediate states $t^i \sim p(t)$, $X^i \sim \mathcal N(t^i x^i_1, (1 - t^i)^2 I)$
        \State Compute loss $\mathcal L(\theta) = \sum_i \|v^\theta_{t^i}(X^i) - \operatorname{sg}\!\left(v^{\mathrm{ref}}_{t^i}(X^i) + g^i\right)\|^2$
        \State Update $\theta$ using $\nabla_\theta \mathcal L(\theta)$
    \EndFor
\EndFor
\State \Return $v^\theta$
\end{algorithmic}
\end{algorithm}

\xhdr{Efficient first-order estimation}
We approximate the intractable target in \Cref{eq:reward_loss} as
\begin{align}
G_t^{\bar v}(x)
&=\mathbb E\!\left[(J_{1,t}^{\bar v})^\top\nabla R(X_1^{\bar v})\mid X_t^{\bar v}=x\right]\notag\\
&\approx\mathbb E\!\left[\nabla R(X_1^{\bar v})\mid X_t^{\bar v}=x\right]
=:\hat G_t^{\bar v}(x).
\end{align}
Here \(J_{1,t}^{\bar v}\) is the stochastic-flow Jacobian, which describes how perturbations of \(X_t^{\bar v}\) propagate to \(X_1^{\bar v}\) with the future Brownian increments held fixed (\S\ref{app:stochastic_jacobian}). We approximate it by the identity matrix, as in reward steering \citep{zhou2026qgf} and fine-tuning with non-differentiable rewards \citep{mcallister2026fdfo}.
\begin{mdframed}[style=MyFrame2]
\begin{proposition}\label{positve-aligned}
    Suppose $\mathrm{Law}(X_1\mid X_t=x)=\mathcal N(T(x),\sigma^2 I)$,
    where $\sigma$ is independent of $x$ and $J_T(x)=\nabla_xT(x)$ has
    positive-definite symmetric part. Then
    $\hat G_t^{\bar v}(x)=\mathbb E[\nabla R(X_1)\mid X_t=x]$
    has positive inner product with $\nabla_x\mathbb E[R(X_1)\mid X_t=x]$
    whenever $\hat G_t^{\bar v}(x)\ne0$. For our Gaussian corruption kernel, the Jacobian of the conditional mean $T(x)=\mathbb E[X_1\mid X_t=x]$ is positive definite.
\end{proposition}
\end{mdframed}
\Cref{positve-aligned} motivates replacing $G_t^{\bar v}(x)$ with
$\hat G_t^{\bar v}(x)$ in the reward regression. Under the Gaussian
posterior model in the proposition, which have also been used in~\citet{eijkelboom2024variational},  the conditionally averaged endpoint
gradient has positive alignment with the exact guidance. 
The Gaussian corruption kernel supplies the conditional-mean Jacobian condition.
Regressing on a single endpoint gradient gives the same expected parameter gradient as regressing on its conditional
mean (\S\ref{sub:simp_endpoint_derivation}). This gives the simplified
reward objective
\begin{align}
    \mathcal L_{\mathrm{simple-reward}}(\theta;\bar v)
    = \mathbb E_{t\sim p(t)}
    \left\|v_t^\theta(X_t^{\bar v})-\bar v_t(X_t^{\bar v})
    -\nabla R(X_1^{\bar v})\right\|^2.
    \label{eq:simple_reward}
\end{align}
We measure the directional alignment of the endpoint guidance on
SD3.5-M in \S\ref{app:gradient_alignment}.

\xhdr{Regularization}
We approximate the gradient of the reference penalty in
\Cref{eq:loss_decomposition} by holding the sampled states
$X_t^{\bar v}$ fixed and differentiating only the velocity difference,
as is common in reward fine-tuning
\citep{xue2026advantage,choi2026scalingreinforcementlearningdiffusion,schulman2017proximal}.

Setting $w(t)=2/\sigma_t^2$ makes the reference penalty a path-space
KL divergence~\citep{berner2022optimal}. The resulting KL-regularized
SOC optimum preserves the reference corruption kernel
(\Cref{eq:same_laws}). In the regression loss, we give the reference
penalty unit weight at each sampled time, following prior work on
alternative weightings in diffusion training and fine-tuning
\citep{ho2020denoising,zheng2026diffusionnft,choi2026scalingreinforcementlearningdiffusion}.
This gives
\begin{align}
    \mathcal L_{\mathrm{simple-regularization}}(\theta;\bar v)
    = \mathbb E_{t\sim p(t)}
    \left\|v_t^\theta(X_t^{\bar v})-v_t^{\mathrm{ref}}(X_t^{\bar v})\right\|^2.
\end{align}
At $v^\theta=\bar v$, the gradients of the simplified reward and
regularization losses sum to the gradient of a single regression
objective (\S\ref{sub:combine_targets}),
\begin{align}
    \mathcal L_{\nameshort}^{\mathrm{SDE}}(\theta;\bar v)
    = \mathbb E_{t\sim p(t)}
    \left\|v_t^\theta(X_t^{\bar v})-v_t^{\mathrm{ref}}(X_t^{\bar v})
    -\nabla R(X_1^{\bar v})\right\|^2.
\end{align}

The time integral in \Cref{thm:reward_surrogate_objective} and \Cref{eq:loss_decomposition} corresponds to uniform time sampling. During training, we use $p(t)=2t$ to give more weight to low-noise states near $t=1$. We ablate this choice and other time weightings in \S\ref{app:ablations}.

\xhdr{Sampling through re-noising} The SDE regression loss requires pairs $(X_t^{\bar v},X_1^{\bar v})$, which we can draw by sampling an endpoint and then an intermediate state from $p_{t|1}^{\bar v}$. 
At initialization and at the KL-regularized SOC optimum (\Cref{eq:same_laws}), this conditional is the pretraining corruption kernel \(\mathcal N(tx_1,(1-t)^2I)\).
We use this kernel as a proxy throughout fine-tuning, independently re-noising endpoints from the current ODE (\Cref{alg:serf_algo}).
We compare this procedure with training on SDE rollout states in \S\ref{app:independent_renoising}.
For linear rewards and an isotropic Gaussian reference, the regression has a reward-tilted fixed point with rescaled reward strength
(\S\ref{sub:fixed_point}).

Trajectory-based methods use states visited during rollout
\citep{lee2026reward,liu2025value,domingoenrich2025adjoint}.
In \S\ref{app:independent_renoising}, we isolate the effect of re-noising by keeping the ODE sampler and training update fixed and changing only how the intermediate states are obtained. 

The following proposition compares the gradient covariance of rollout states and independent conditional draws under a common SDE joint law.
\begin{mdframed}[style=MyFrame2]
\begin{proposition}\label{lower-variance}
    Define the single-sample gradients $g_t(X_t,X_1)=\nabla_\theta\|v_t^\theta(X_t)-v_t^{\mathrm{ref}}(X_t)-\nabla R(X_1)\|^2$, with the sample pair held fixed, and assume they have a finite second moment. For IID $t_i\sim p(t)$ drawn independently of the SDE trajectory, consider the two estimates:
    \begin{itemize}[leftmargin=1.2em, itemsep=0em]
        \item $g_{\mathrm{traj}} =  \frac1m\sum_{i=1}^m g_{t_i}(X_{t_i}, X_1)$ where $(X_t)_{t \in [0, 1]}$ is an SDE trajectory under $\bar v$.
        
        \item $g_{\mathrm{corr}} = \frac1m\sum_{i=1}^m g_{t_i}(X'_{t_i}, X_1)$ where $X_1 \sim p_1^{\bar v}$ and $
        X'_{t_i} \sim p_{t_i|1}^{\bar v}(\cdot \mid X_1)
        $ independently given the endpoint and times.
    \end{itemize}
    Then $g_\mathrm{traj}$ and $g_{\mathrm{corr}}$ are unbiased estimators of $\nabla_\theta \mathcal L_{\nameshort}^{\mathrm{SDE}}(\theta;\bar v)$, and $\mathrm{Cov}(g_{\mathrm{corr}}) \preceq \mathrm{Cov}(g_{\mathrm{traj}})$.
\end{proposition}
\end{mdframed}

\Cref{lower-variance} motivates independent re-noising: under the same
SDE joint law, independent conditional draws preserve the expected
regression gradient and yield no larger covariance than draws from
a shared trajectory. We prove this in \S\ref{sub:combine_targets}.

Applying the regression to independently re-noised ODE endpoints gives
\Cref{eq:SERF_objective}. \Cref{alg:serf_algo} reuses each endpoint reward
gradient across inner updates, drawing fresh noisy states at each update.

\begin{figure}[h]
    \centering
    \vspace{-10pt}\includegraphics[width=1.0\linewidth]{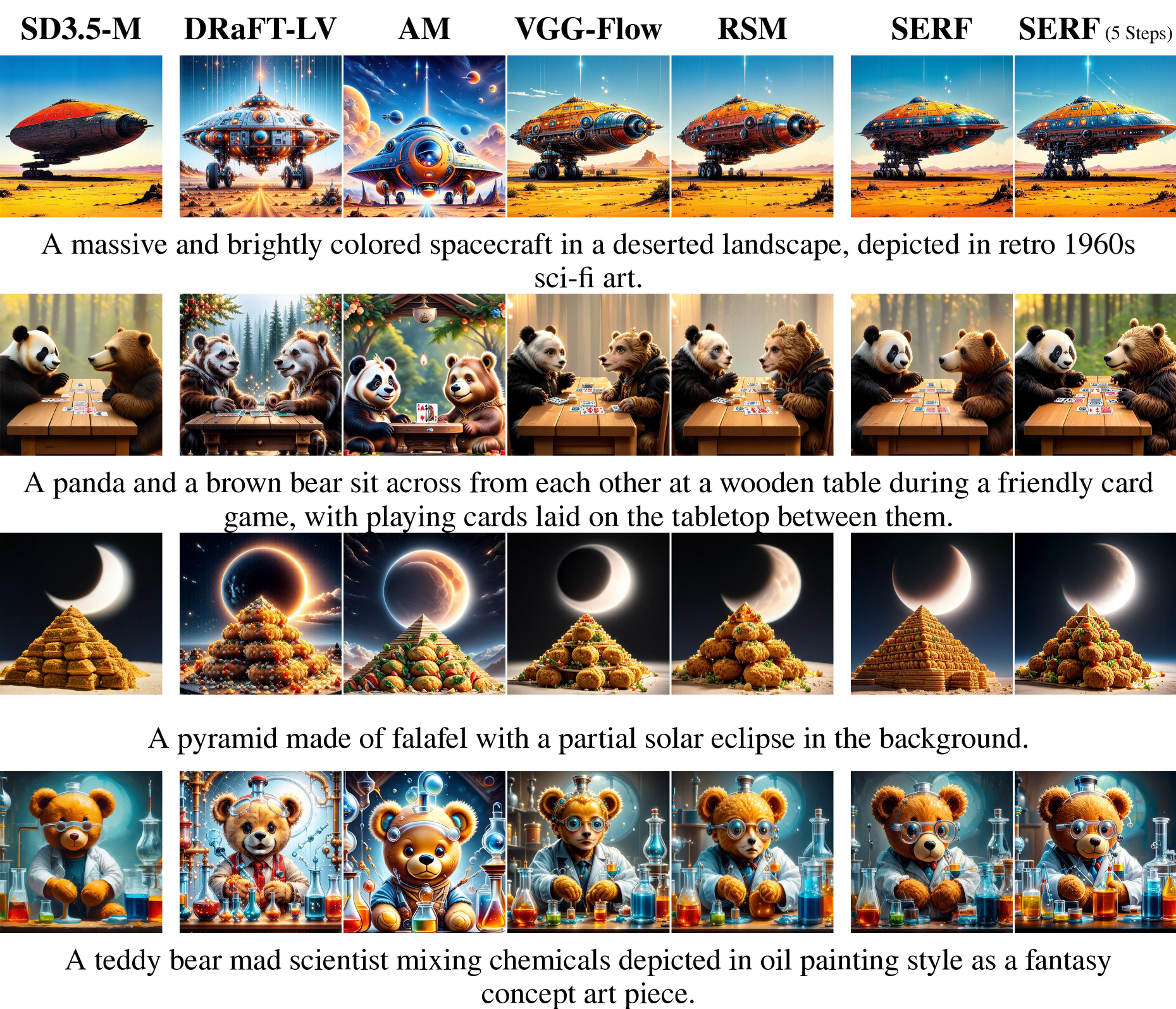}
    \vspace{-15pt}
    \caption{Qualitative comparison across four different prompts. The final column
            shows samples from \nameshort post-trained using five sampling steps
            per training rollout.}
    \label{fig:qualitative}
    \vspace{-15pt}
\end{figure}

\clearpage

\section{Experiments}

\subsection{Experimental Settings}
\label{sec:exp_settings}
\looseness=-1
\xhdr{Setup}
We post-train Stable Diffusion 3.5 Medium (SD3.5-M)~\citep{esser2024scaling} at $512 \times 512$ resolution on two differentiable human-preference rewards: PickScore~\citep{kirstain2023pickapic} and Human Preference Score v2.1 (HPSv2.1)~\citep{wu2023hpsv2}. We use training prompts from Pick-a-Pic~\citep{kirstain2023pickapic} for PickScore and Human Preference Dataset v2 (HPDv2)~\citep{wu2023hpsv2} for HPSv2.1.

\looseness=-1
\noindent \xhdr{Baselines} We compare against four first-order fine-tuning methods: 
VGG-Flow~\citep{liu2025value}, which learns a correction to one-step reward guidance;
RSM~\citep{lee2026reward}, which removes the auxiliary correction network;
Adjoint Matching~\citep{domingoenrich2025adjoint}, which constructs targets through a backward adjoint solve; and
DRaFT-LV~\citep{clark2024directly}, which averages final-step reward
gradients over re-noised samples. \Cref{tab:baseline_comparison} provides a detailed comparison.

\begin{wrapfigure}[42]{r}{0.50\columnwidth}
  \centering
  \vspace{-10pt}
  \includegraphics[width=\linewidth]{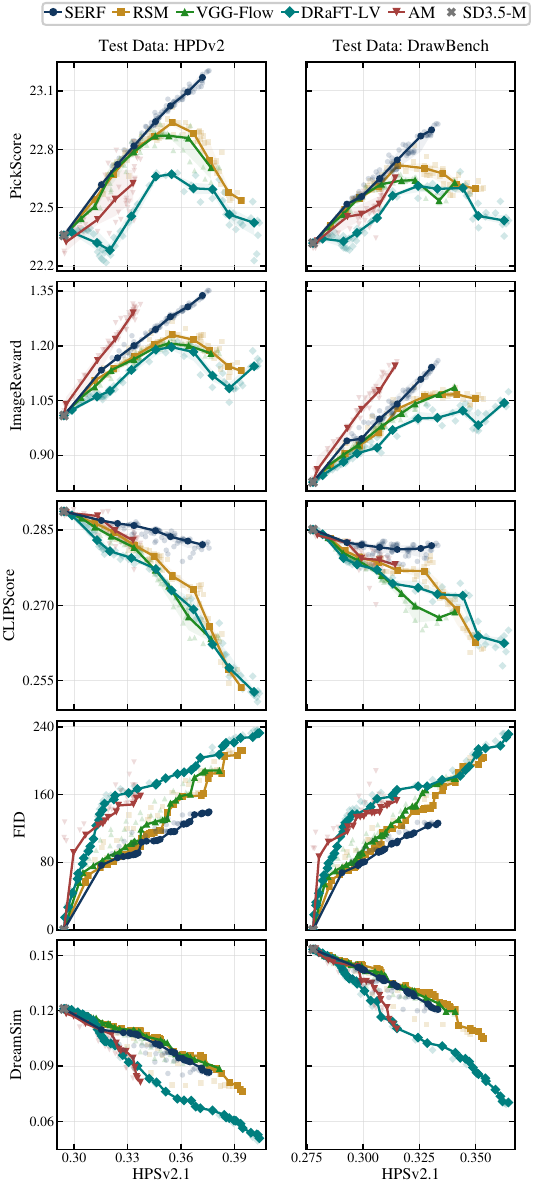}
  \caption{Reward--quality trade-offs for HPSv2.1-trained models, evaluated on HPDv2 (left) and DrawBench (right).}
  \label{fig:pareto}
\end{wrapfigure}

\noindent \xhdr{Training Settings}
\looseness=-1
We train LoRA~\citep{hu2021lora} adapters with rank 32 and scaling parameter 64 for 300 AdamW updates~\citep{loshchilov2017decoupled}, using a default learning rate of $1\times10^{-4}$, $\beta_1=0.9$, $\beta_2=0.99$, and weight decay $1\times 10^{-4}$. Each update uses 64 generated trajectories and four training targets per sample. Training rollouts use 10 Euler steps, except for Adjoint Matching, which uses 10 steps of its stochastic memoryless sampler~\citep{domingoenrich2025adjoint} (see~\Cref{fig:am_rollout_steps}). All methods use CFG $4.5$, \texttt{bfloat16} precision, and an EMA of the LoRA parameters with decay $0.9$. We train on four NVIDIA GH200 GPUs. For each task, we sweep the learning rate for DRaFT-LV and the
reward coefficient $\beta$ for the other methods to obtain runs
spanning different reward levels within the 300-update budget.
The default \nameshort method uses time sampler $p(t)=2t$
(\S\ref{app:time_weighting}). The full search spaces and selected settings are given in
\S~\ref{app:hyperparameter_search}.

\noindent\xhdr{Evaluation Settings}
\looseness=-1 Every 20 updates, we evaluate the EMA model with 20 Euler steps and CFG $4.5$ on 200 held-out prompts from the training task's dataset and 200 DrawBench prompts~\citep{saharia2022photorealistic}.
We generate 16 images per prompt using fixed seeds, giving 3,200 images per evaluation set. 
We report PickScore, HPSv2.1, ImageReward~\citep{xu2023imagereward}, and CLIPScore~\citep{hessel2021clipscore}. 
Diversity is the mean pairwise cosine distance between DreamSim features~\citep{fu2023dreamsim} within each prompt, averaged across prompts. We also compute FID~\citep{heusel2017gans} separately for each prompt against images from the frozen base model, then average these conditional FID values across prompts.
    
\subsection{Main Results}
\label{sec:mainresults}

\noindent\xhdr{Quantitative Results}
\looseness=-1 \Cref{fig:pareto} compares reward quality trade-offs for HPSv2.1 fine-tuning across five hyperparameter settings per method. \nameshort improves held-out human-preference scores, image prompt
similarity, and FID over RSM and VGG-Flow while maintaining comparable
diversity. DRaFT-LV achieves large gains in the optimized reward
but weaker performance on the remaining metrics.
Adjoint Matching remains competitive on ImageReward but does not
reach the higher HPSv2.1 rewards within 300 updates.

\noindent \looseness=-1\Cref{tab:quantitative_pickscore_matched,tab:quantitative_hpsv2_matched}
compare PickScore- and HPSv2.1-trained models at approximately
matched optimized rewards. We select target levels that provide
substantial reward gains without pronounced reward hacking,
allowing the remaining metrics to be compared at similar reward
levels. Results are means $\pm$ sample standard deviations across
three independent training seeds.  Across all four evaluation settings, \nameshort achieves the lowest FID and highest CLIPScore, while maintaining
diversity comparable to RSM and VGG-Flow. \nameshort also improves training efficiency relative to VGG-Flow and
Adjoint Matching, as detailed in \S~\ref{sec:ablations} and
Appendix~\ref{app:training_efficiency}.

\noindent
\looseness=-1\nameshort also ranks first or second in overall GenEval accuracy
under both training rewards
(\S\ref{app:quantitative_results}). Further, \nameshort improves reward rapidly per update while
its velocity deviation stabilizes early
(\S\ref{app:policy_deviation}).

\begin{table*}[t]
\centering
\caption{
\looseness=-1 Comparison of PickScore-trained models.
Selected checkpoints target a PickScore of $23.45$
on Pick-a-Pic and $23.40$ on DrawBench.
Values are means $\pm$ sample standard deviations across three
independent training seeds.
Shaded cells denote the optimized reward.
}
\vspace{-2pt}
\label{tab:quantitative_pickscore_matched}

\resizebox{1\textwidth}{!}{
\setlength{\tabcolsep}{5pt}
\renewcommand{\arraystretch}{1.10}

\begin{tabular}{lcccccc}
\toprule
Method
& PickScore $\uparrow$
& HPSv2 $\uparrow$
& ImageReward $\uparrow$
& CLIPScore $\uparrow$
& FID $\downarrow$
& Diversity $\uparrow$ 
\\
\midrule

\multicolumn{7}{c}{\textit{Evaluation: Pick-a-Pic prompts}} \\

\textcolor{gray}{SD3.5}
& \textcolor{gray}{$19.840$}
& \textcolor{gray}{$0.207$}
& \textcolor{gray}{$-0.532$}
& \textcolor{gray}{$0.247$}
& \textcolor{gray}{--}
& \textcolor{gray}{$0.248$}
\\

SD3.5 + CFG
& $21.543$
& $0.293$
& $0.971$
& $0.290$
& --
& $0.128$
\\
\midrule

DRaFT-LV
& \cellcolor{rewardcell}\pmt{23.418}{0.024}
& \pmt{0.3227}{0.0007}
& \pmt{1.039}{0.009}
& \pmt{0.2742}{0.0010}
& \pmt{221.3}{4.4}
& \pmt{0.0627}{0.0008}
\\

VGGFlow
& \cellcolor{rewardcell}\pmt{23.367}{0.009}
& \pmt{0.3149}{0.0011}
& \pmt{1.088}{0.014}
& \pmt{0.2836}{0.0007}
& \pmt{180.7}{1.3}
& \bpmt{0.1024}{0.0013}
\\

Adjoint Matching
& \cellcolor{rewardcell}\pmt{23.307}{0.080}
& \bpmt{0.3353}{0.0013}
& \bpmt{1.426}{0.030}
& \underline{\pmt{0.2929}{0.0019}}
& \pmt{191.6}{7.0}
& \pmt{0.0588}{0.0011}
\\

RSM
& \cellcolor{rewardcell}\pmt{23.422}{0.031}
& \pmt{0.3207}{0.0005}
& \pmt{1.154}{0.012}
& \pmt{0.2850}{0.0008}
& \pmt{172.6}{5.4}
& \underline{\pmt{0.0974}{0.0011}}
\\

\textbf{SERF}
& \cellcolor{rewardcell}\pmt{23.454}{0.020}
& \underline{\pmt{0.3242}{0.0007}}
& \underline{\pmt{1.213}{0.003}}
& \bpmt{0.2969}{0.0002}
& \bpmt{144.9}{1.3}
& \underline{\pmt{0.0954}{0.0009}}
\\

\midrule
\multicolumn{7}{c}{\textit{Evaluation: DrawBench prompts}} \\

\textcolor{gray}{SD3.5}
& \textcolor{gray}{$20.628$}
& \textcolor{gray}{$0.200$}
& \textcolor{gray}{$-0.611$}
& \textcolor{gray}{$0.239$}
& \textcolor{gray}{--}
& \textcolor{gray}{$0.279$} 
\\

SD3.5 + CFG
& $22.318$
& $0.278$
& $0.826$
& $0.285$
& --
& $0.153$
\\
\midrule

DRaFT-LV
& \cellcolor{rewardcell}\pmt{23.416}{0.030}
& \underline{\pmt{0.2983}{0.0006}}
& \pmt{0.880}{0.021}
& \pmt{0.2645}{0.0002}
& \pmt{190.7}{7.2}
& \pmt{0.0918}{0.0015}
\\

VGGFlow
& \cellcolor{rewardcell}\pmt{23.207}{0.012}
& \pmt{0.2882}{0.0006}
& \pmt{0.953}{0.020}
& \pmt{0.2734}{0.0006}
& \underline{\pmt{151.5}{1.9}}
& \bpmt{0.1328}{0.0005}
\\

Adjoint Matching
& \cellcolor{rewardcell}\pmt{23.625}{0.110}
& \bpmt{0.3117}{0.0018}
& \bpmt{1.283}{0.030}
& \bpmt{0.2799}{0.0049}
& \pmt{168.6}{8.3}
& \pmt{0.0823}{0.0032}
\\

RSM
& \cellcolor{rewardcell}\pmt{23.332}{0.035}
& \pmt{0.2939}{0.0006}
& \pmt{1.018}{0.017}
& \pmt{0.2751}{0.0017}
& \underline{\pmt{148.0}{3.7}}
& \underline{\pmt{0.1271}{0.0012}}
\\

\textbf{SERF}
& \cellcolor{rewardcell}\pmt{23.366}{0.032}
& \underline{\pmt{0.2977}{0.0005}}
& \underline{\pmt{1.081}{0.012}}
& \bpmt{0.2840}{0.0006}
& \bpmt{121.1}{1.6}
& \underline{\pmt{0.1284}{0.0003}}
\\
\bottomrule
\end{tabular}
}
\vspace{-15pt}
\end{table*}

\begin{table*}[h]
\centering
\vspace{-10pt}
\caption{
Comparison of HPSv2-trained models.
Selected checkpoints target an HPSv2 score of $0.365$ on
HPSv2 prompts and $0.325$ on DrawBench.
Values are means $\pm$ sample standard deviations across three
independent training seeds.
Shaded cells denote the optimized reward.
}
\vspace{-2pt}
\label{tab:quantitative_hpsv2_matched}

\resizebox{1\textwidth}{!}{
\setlength{\tabcolsep}{5pt}
\renewcommand{\arraystretch}{1.10}

\begin{tabular}{lcccccc}
\toprule
Method
& PickScore $\uparrow$
& HPSv2 $\uparrow$
& ImageReward $\uparrow$
& CLIPScore $\uparrow$
& FID $\downarrow$
& Diversity $\uparrow$
\\
\midrule

\multicolumn{7}{c}{\textit{Evaluation: HPDv2 prompts}} \\

\textcolor{gray}{SD3.5}
& \textcolor{gray}{$20.517$}
& \textcolor{gray}{$0.206$}
& \textcolor{gray}{$-0.357$}
& \textcolor{gray}{$0.250$}
& \textcolor{gray}{--}
& \textcolor{gray}{$0.241$} 
\\

SD3.5 + CFG
& $22.358$
& $0.294$
& $1.009$
& $0.289$
& --
& $0.121$ \\
\midrule

DRaFT-LV
& \pmt{22.700}{0.098}
& \cellcolor{rewardcell}\pmt{0.3648}{0.0029}
& \pmt{1.214}{0.038}
& \pmt{0.2682}{0.0035}
& \pmt{193.0}{5.7}
& \pmt{0.0711}{0.0021} \\

VGGFlow
& \underline{\pmt{22.858}{0.057}}
& \cellcolor{rewardcell}\pmt{0.3645}{0.0014}
& \pmt{1.204}{0.008}
& \pmt{0.2708}{0.0009}
& \pmt{160.4}{0.2}
& \bpmt{0.0967}{0.0018} \\

Adjoint Matching
& \pmt{22.593}{0.201}
& \cellcolor{rewardcell}\pmt{0.3382}{0.0029}
& \bpmt{1.282}{0.086}
& \underline{\pmt{0.2755}{0.0044}}
& \pmt{189.1}{18.8}
& \pmt{0.0744}{0.0067} \\

RSM
& \underline{\pmt{22.882}{0.021}}
& \cellcolor{rewardcell}\pmt{0.3653}{0.0021}
& \pmt{1.232}{0.015}
& \underline{\pmt{0.2753}{0.0004}}
& \underline{\pmt{148.0}{1.9}}
& \bpmt{0.0988}{0.0005} \\

\textbf{SERF}
& \bpmt{23.121}{0.012}
& \cellcolor{rewardcell}\pmt{0.3639}{0.0007}
& \bpmt{1.316}{0.006}
& \bpmt{0.2838}{0.0008}
& \bpmt{126.5}{1.1}
& \pmt{0.0924}{0.0009} \\

\midrule
\multicolumn{7}{c}{\textit{Evaluation: DrawBench prompts}} \\

\textcolor{gray}{SD3.5}
& \textcolor{gray}{$20.628$}
& \textcolor{gray}{$0.200$}
& \textcolor{gray}{$-0.611$}
& \textcolor{gray}{$0.239$}
& \textcolor{gray}{--}
& \textcolor{gray}{$0.279$} 
\\

SD3.5 + CFG
& $22.318$
& $0.278$
& $0.826$
& $0.285$
& --
& $0.153$ \\
\midrule

DRaFT-LV
& \pmt{22.639}{0.035}
& \cellcolor{rewardcell}\pmt{0.3277}{0.0028}
& \pmt{1.032}{0.006}
& \pmt{0.2709}{0.0024}
& \pmt{174.3}{6.1}
& \pmt{0.1039}{0.0029} \\

VGGFlow
& \pmt{22.632}{0.025}
& \cellcolor{rewardcell}\pmt{0.3231}{0.0014}
& \pmt{1.025}{0.011}
& \pmt{0.2724}{0.0027}
& \pmt{145.5}{1.6}
& \bpmt{0.1316}{0.0022} \\

Adjoint Matching
& \pmt{22.508}{0.103}
& \cellcolor{rewardcell}\pmt{0.3127}{0.0039}
& \bpmt{1.110}{0.034}
& \underline{\pmt{0.2717}{0.0058}}
& \pmt{178.0}{19.6}
& \pmt{0.1011}{0.0100} \\

RSM
& \underline{\pmt{22.664}{0.012}}
& \cellcolor{rewardcell}\pmt{0.3240}{0.0011}
& \pmt{1.034}{0.010}
& \underline{\pmt{0.2786}{0.0013}}
& \underline{\pmt{132.8}{1.0}}
& \bpmt{0.1340}{0.0004} \\

\textbf{SERF}
& \bpmt{22.785}{0.003}
& \cellcolor{rewardcell}\pmt{0.3216}{0.0004}
& \bpmt{1.084}{0.007}
& \bpmt{0.2825}{0.0004}
& \bpmt{114.0}{1.4}
& \pmt{0.1289}{0.0007} \\

\bottomrule

\end{tabular}
}
\vspace{-5pt}
\end{table*}

\noindent \xhdr{Qualitative Results}
\looseness=-1
\Cref{fig:qualitative} shows samples from the HPSv2.1-trained checkpoints in~\Cref{tab:quantitative_hpsv2_matched}. DRaFT-LV and Adjoint Matching
often produce unnatural images, while VGG-Flow and RSM sometimes give animals with
human-like facial features, 
\begin{wrapfigure}[9]{r}{0.48\columnwidth}
  \centering
  \vspace{-5pt}
  \includegraphics[width=\linewidth]{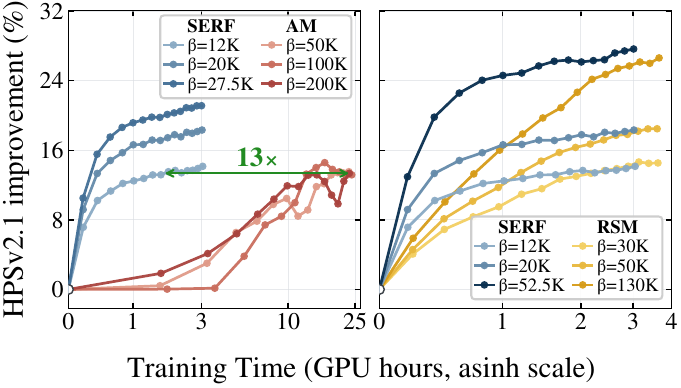}
  \vspace{-15pt}
  \caption{HPSv2.1 improvement over training time for \nameshort,
  AM (left), and RSM (right).}
  \vspace{5pt}
  \label{fig:convergence_speed}
\end{wrapfigure}
consistent with reward hacking. 
\nameshort produces natural,
prompt-consistent images, including when training rollouts use only 5 sampling steps. 
All displayed images are generated with 20 Euler steps. Additional qualitative results are provided in~\S\ref{app:qualitative}.

\begin{figure}[h]
    \vspace{20pt}
    \centering
    \includegraphics[width=\linewidth]{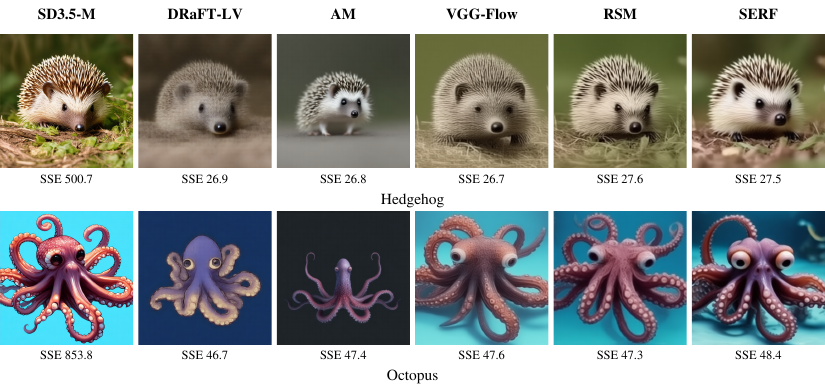}
    \vspace{-15pt}
    \caption{\looseness=-1 Selected JPEG-compressibility comparisons using identical prompts and noise seeds within each row. Reconstruction SSE differs by at most $4\%$ across trained methods, except the base model.}
    \label{fig:jpeg_qualitative_main}
\end{figure}

\noindent\xhdr{JPEG compressibility}
\looseness=-1Following~\citet{clark2024directly}, we minimize JPEG reconstruction
error (quality 75); \nameshort better preserves subject detail at
matched rewards (\Cref{fig:jpeg_qualitative_main};
\S~\ref{app:qualitative}). 

\subsection{Ablations}
\label{sec:ablations}

\begin{figure}[h]
    \centering
    \includegraphics[width=1.0\linewidth]{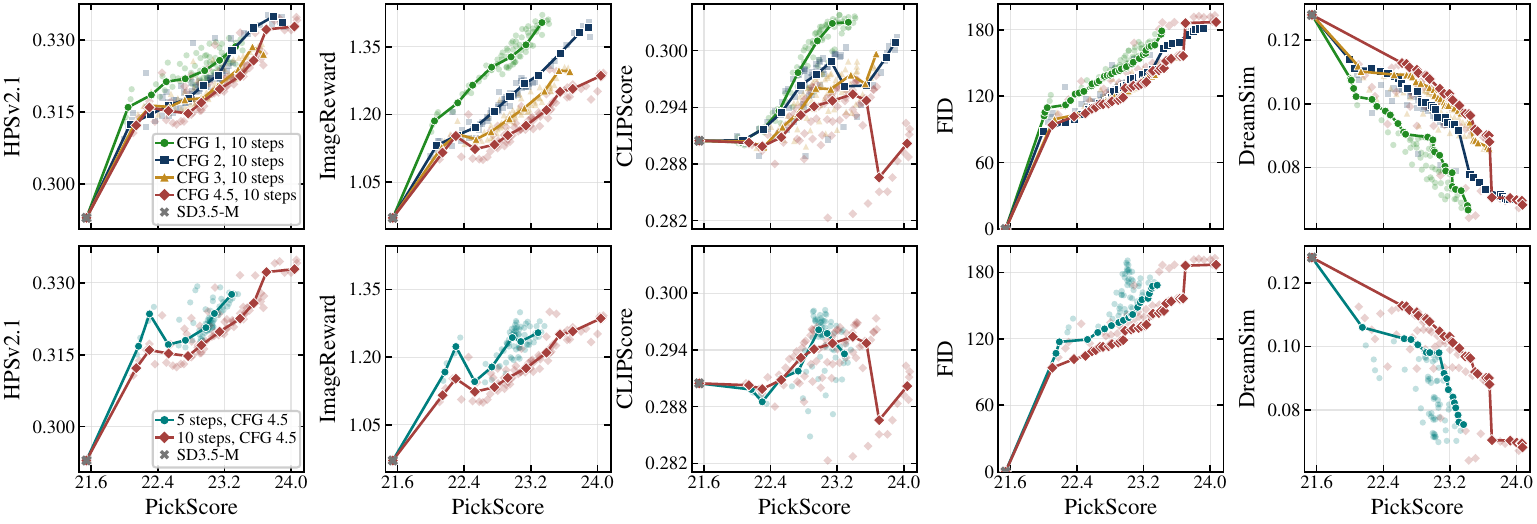}
    \vspace{-15pt}
    \caption{\nameshort trained on PickScore and evaluated on Pick-a-Pic. Top: Comparison for different training CFG scales. Bottom: using five or ten sampling steps per training rollout.}
    \vspace{-10pt}
    \label{fig:cfg_steps}
\end{figure}

\noindent\xhdr{Convergence speed}
\Cref{fig:convergence_speed} compares how quickly \nameshort, RSM,
and Adjoint Matching improve HPSv2.1 during post-training.
Adjoint Matching does not reach high HPSv2.1 rewards within
300 gradient updates and requires roughly $13\times$ more
training time than \nameshort to reach comparable improvement.
\nameshort also improves faster than RSM.
In our setting, \nameshort fine-tunes SD3.5-M in roughly three
GPU hours. Additional timing comparisons and update costs
are reported in \S\ref{app:training_efficiency}.

\noindent\xhdr{Training settings}
\Cref{fig:cfg_steps} ablates the training CFG scale and number of
sampling steps per training rollout. The top row varies training CFG over
$1,2,3,4.5$, with evaluation CFG fixed at $4.5$.
Lower training CFG generally improves reward scores at the cost
of worse diversity and FID. The bottom row compares five and ten
sampling steps per training rollout. \nameshort remains effective
with five steps because re-noising supplies independent training
states whose number is not limited by the sampling rollout.

We report additional ablations in \S~\ref{app:ablations} and \S~\ref{app:e}.

\section{Related Work}

\looseness=-1 
Reward finetuning methods for diffusion and flow models differ in whether they require differentiable rewards. We focus on
training-time methods; see \citet{uehara2025inference} for test-time approaches.

\looseness=-1
\xhdr{Non-differentiable rewards}
\textit{Trajectory} methods such as DDPO, DPOK, Flow-GRPO, and DanceGRPO apply policy gradients to Gaussian denoising transitions \citep{black2024ddpo,flowgrpo, fan2023dpok,xue2025dancegrpo}. 
\textit{ELBO} methods reward complete generations through pretraining-like losses: 
FPO~\citep{mcallister2025flow,choi2026rethinking} uses likelihood bounds,
AWM~\citep{xue2026advantage} advantage weights,
DiffusionNFT~\citep{zheng2026diffusionnft,kveton2026advantageflow,fan2025online} high- and low-reward samples,
and RAM~\citep{bergmeister2026reinforceadjointmatchingscaling} a stochastic-control target.
FDFO~\citep{mcallister2026fdfo} applies an endpoint direction at intermediate states, estimated from paired generations without
reward gradients.

\looseness=-1
\xhdr{Differentiable rewards}
AlignProp backpropagates reward gradients through the sampling trajectory~\citep{prabhudesai2023aligning}, while DRaFT-$K$, DRTune, and LeapAlign limit the backward computation through stop gradients~\citep{clark2024directly,wu2024deep,liang2026leapalign}. Intermediate reward signals can also be constructed through local denoising and reconstruction~\citep{xu2023imagereward,shen2025directlyaligningdiffusiontrajectory,kang2026diffusion,tan2026easytune}. ReFL and EasyTune obtain gradients at intermediate states by evaluating the reward at Tweedie estimates of clean samples. ReFL trains only on late denoising steps, while EasyTune's algorithm is mainly for motion representations. \nameshort evaluates the reward once at generated endpoints and reuses its gradient across noise levels. VGG-Flow learns an auxiliary guidance network~\citep{liu2025value}; Adjoint Matching integrates backward adjoint dynamics, while Efficient Adjoint Matching obtains a closed-form adjoint but requires the pretrained terminal score~\citep{domingoenrich2025adjoint,shin2026efficient}. $\nabla$-GFlowNet uses reward gradients in a detailed-balance objective~\citep{liu2025efficient}, whereas RSM applies endpoint reward gradients at sampled trajectory states~\citep{lee2026reward}. Beyond gradient propagation, Reward Sharpness-Aware Fine-Tuning addresses gradient robustness through perturbations of model parameters and generated samples~\citep{kim2026reward}. We discuss these connections in \S\ref{sec:related-work}.

\section{Conclusion}
\label{sec:conclusion}
\looseness=-1
We introduced \namelong, a first-order method for reward fine-tuning flow models. \nameshort uses reward gradients at fully generated samples to construct regression targets at independently re-noised states, without backpropagating through the sampling trajectory. We motivate this objective through stochastic optimal control and establish conditions under which its guidance aligns with the expected-reward gradient. On Stable Diffusion 3.5 Medium, \nameshort achieves strong reward quality trade-offs while reducing training cost. It remains effective with five-step rollouts and multiple training targets per generated sample, showing that inexpensive sample generation can support effective reward fine-tuning.

\noindent
\looseness=-1\xhdr{Limitations}
\nameshort approximates gradient transport by reusing endpoint reward gradients directly; scalar corrections cannot recover changes in their direction. Reusing endpoints reduces sampling cost but does not remove the need for diverse generated samples. Our experiments focus on text-to-image generation with SD3.5-M; evaluating other backbones and domains remains future work.

\subsubsection*{Acknowledgments}
We thank the EuroHPC Joint Undertaking for providing access to Jupiter at JSC, Germany, under project EHPC-AIF-2026SC03-183. AITHYRA is supported by the Austrian Academy of Sciences and the not-for-profit Boehringer Ingelheim Stiftung. M.B. is partially supported by the EPSRC Turing AI World-Leading Research Fellowship No. EP/X040062/1 and EPSRC AI Hub No. EP/Y028872/1.

\cut{
\subsection*{AI use statement}

In this work, we used generative AI tools to improve the clarity of the author-written text,
including the presentation of mathematical claims, identify potentially relevant literature, assist with implementation of the method, and provide critical feedback on our research ideas and methodology. We did not use generative AI tools to generate synthetic datasets, help develop theoretical models or conceptual frameworks, provide critical ingredients for proving mathematical claims, proposing or refining hypotheses, assist with translation, cleaning and reformatting datasets, supporting qualitative and thematic data analysis, or interpreting results. We have reviewed all AI-assisted work. We take responsibility for the final content of this work,
including text, claims or artifacts produced with the aid of generative AI.

}

\subsection*{Ethics statement}

This paper presents work whose goal is to advance the field of machine learning. There are many potential societal consequences of our work, none of which we feel needs to be specifically highlighted here.

\cut{
\subsection*{Reproducibility statement}

We describe the SERF in the main text and provide implementation details, hyperparameters, and proofs in the appendix. All datasets used are publicly available. We will release our code in a public GitHub repository upon publication.
}

\clearpage
\bibliography{references}
\bibliographystyle{iclr2027_conference}
\newpage
\input{appendix_rebuttal}

\end{document}

%% file: math_commands.tex
\usepackage{amsmath, amsthm, amsfonts, bm}
\usepackage{mathtools}
\usepackage{thmtools}
\usepackage{thm-restate}
\usepackage{xcolor}
\usepackage{xspace}
\usepackage[framemethod=TikZ]{mdframed}

\newtheorem{theorem}{Theorem}
\newtheorem{proposition}{Proposition}

\newtheorem{lemma}{Lemma}

\def\eqref#1{equation~\ref{#1}}
\def\1{\bm{1}}

\DeclareMathAlphabet{\mathsfit}{\encodingdefault}{\sfdefault}{m}{sl}
\SetMathAlphabet{\mathsfit}{bold}{\encodingdefault}{\sfdefault}{bx}{n}

\newcommand{\xhdr}[1]{{\noindent\bfseries #1}.}
\newcommand{\cut}[1]{}

\newcommand{\namelong}{\textsc{Simple Endpoint Reward Finetuning}\xspace}
\newcommand{\nameshort}{\textsc{SERF}\xspace}

\newcommand{\diff}{\mathrm{d}}

\usepackage[framemethod=TikZ]{mdframed}
\mdfdefinestyle{MyFrame}{%
    linecolor=black,
    outerlinewidth=.3pt,
    roundcorner=5pt,
    innertopmargin=1pt, %\baselineskip,
    innerbottommargin=1pt, %\baselineskip,
    innerrightmargin=1pt,
    innerleftmargin=1pt,
    backgroundcolor=black!0!white}

\mdfdefinestyle{MyFrame2}{%
    linecolor=white,
    outerlinewidth=1pt,
    roundcorner=1pt,
    innertopmargin=3pt,
    innerbottommargin=2pt,
    innerrightmargin=7pt,
    innerleftmargin=7pt,
    backgroundcolor=black!50!white}

\mdfdefinestyle{MyFrameEq}{%
    linecolor=white,
    outerlinewidth=0pt,
    roundcorner=0pt,
    innertopmargin=0pt,
    innerbottommargin=0pt,
    innerrightmargin=7pt,
    innerleftmargin=7pt,
    backgroundcolor=black!3!white}

\usepackage[framemethod=TikZ]{mdframed}
\mdfdefinestyle{MyFrame}{%
    linecolor=black,
    outerlinewidth=.3pt,
    roundcorner=5pt,
    innertopmargin=1pt, %\baselineskip,
    innerbottommargin=1pt, %\baselineskip,
    innerrightmargin=1pt,
    innerleftmargin=1pt,
    backgroundcolor=black!0!white}

\mdfdefinestyle{MyFrame2}{%
    linecolor=white,
    outerlinewidth=1pt,
    roundcorner=1pt,
    innertopmargin=3pt,
    innerbottommargin=2pt,
    innerrightmargin=7pt,
    innerleftmargin=7pt,
    backgroundcolor=black!5!white}

\mdfdefinestyle{MyFrameEq}{%
    linecolor=white,
    outerlinewidth=0pt,
    roundcorner=0pt,
    innertopmargin=0pt,
    innerbottommargin=0pt,
    innerrightmargin=7pt,
    innerleftmargin=7pt,
    backgroundcolor=black!3!white}

\mdfdefinestyle{box}{
    backgroundcolor=black!3,
    linecolor=black!35,
    linewidth=0.5pt,
    roundcorner=0pt,
    innerleftmargin=9pt,
    innerrightmargin=9pt,
    innertopmargin=6pt,
    innerbottommargin=6pt,
    skipabove=10pt,
    skipbelow=10pt
}

\mdfdefinestyle{box2}{
    backgroundcolor=black!3,
    linecolor=black!35,
    linewidth=0.4pt,
    roundcorner=0pt,
    innerleftmargin=6pt,
    innerrightmargin=6pt,
    innertopmargin=2pt,
    innerbottommargin=5pt,
    skipabove=5pt,
    skipbelow=5pt
}

%% file: figures/serf_overview-2-2.tex
\begingroup%
\endlinechar=-1\relax
\definecolor{serfink}{HTML}{283541}
\definecolor{serfmuted}{HTML}{727D86}
\definecolor{serfreward}{HTML}{861F38}
\definecolor{serfref}{HTML}{626BB1}
\definecolor{serftarget}{HTML}{167D78}
\definecolor{serfrenoise}{HTML}{497F99}
\definecolor{serfsky}{HTML}{BCDCE5}
\definecolor{serfmountain}{HTML}{638793}
\definecolor{serfforest}{HTML}{315C59}
\definecolor{serfgrass}{HTML}{87A585}

% An illustrative clean image, with a noise overlay for the corruptions.
\newcommand{\serftile}[4]{%
  \begin{scope}[shift={(#1,#2)},scale=#3]
    \begin{scope}
      \clip[rounded corners=2pt] (-.5,-.5) rectangle (.5,.5);
      \fill[serfsky] (-.5,-.5) rectangle (.5,.5);
      \fill[orange!28!yellow!35!white] (.24,.27) circle (.105);
      \fill[serfmountain]
        (-.62,-.17) -- (-.12,.35) -- (.28,-.17) -- cycle;
      \fill[white!90!serfsky]
        (-.235,.23) -- (-.12,.35) -- (-.015,.213)
        -- (-.10,.25) -- (-.15,.20) -- cycle;
      \fill[serfmountain!70!serfsky]
        (-.15,-.22) -- (.29,.20) -- (.65,-.22) -- cycle;
      \fill[serfgrass]
        (-.5,-.16) .. controls (-.1,-.01) and (.12,-.28) .. (.5,-.12)
        -- (.5,-.5) -- (-.5,-.5) -- cycle;
      \fill[serfforest]
        (-.5,-.35) .. controls (-.12,-.18) and (.13,-.42) .. (.5,-.30)
        -- (.5,-.5) -- (-.5,-.5) -- cycle;
      \fill[serfsky!70!white]
        (-.05,-.17) .. controls (.25,-.25) and (-.20,-.28) .. (.14,-.5)
        -- (.27,-.5) .. controls (-.07,-.30) and (.31,-.23) .. (.0,-.17)
        -- cycle;
      \node[inner sep=0pt,transform shape,opacity=#4] at (0,0)
        {\includegraphics[width=1cm,height=1cm]{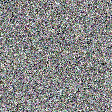}};
    \end{scope}
    \draw[serfink!20,line width=.45pt,rounded corners=2pt]
      (-.5,-.5) rectangle (.5,.5);
  \end{scope}%
}

\begin{tikzpicture}[x=1cm,y=1cm,text=serfink,font=\small,
  flow/.style={
    -{Latex[length=2.8mm,width=2.1mm]},
    line width=1.6pt
  },
  back/.style={
    -{Latex[length=2.2mm,width=1.6mm]},
    line width=1.05pt,
    serfrenoise
  },
  grad/.style={
    -{Latex[length=2mm,width=1.5mm]},
    line width=1.35pt,
    serfreward
  },
  label/.style={font=\normalsize\sffamily},
  note/.style={font=\small\sffamily,text=serfmuted}]

% One irregular reward peak, with nested nonintersecting level sets.
\begin{scope}[shift={(13.2,3.95)},rotate=17]
  \foreach \r/\c in {
    1/FCF2EE,
    .84/F9E4DC,
    .68/F3D1C4,
    .52/E9B9A9,
    .36/DDA28F,
    .20/CF8C7A
  } {
    \definecolor{rewardlevel}{HTML}{\c}
    \fill[rewardlevel]
      plot[smooth cycle,domain=0:360,samples=100,variable=\a]
      ({1.58*\r*(1+.12*sin(3*\a)+.065*cos(5*\a))*cos(\a)},
       {.83*\r*(1+.12*sin(3*\a)+.065*cos(5*\a))*sin(\a)});
  }
\end{scope}

% Forward sampling trajectory.
\draw[flow] (2.12,3.48)
  .. controls (4.65,4.87) and (7.87,4.80) .. (10.20,3.48);
\node[label,anchor=south] at (5.92,4.57) {Sampling};

% Backward arrows terminate at fresh noisy training inputs.
\draw[back] (10.28,3.14)
  .. controls (8.15,3.87) and (4.12,3.82) .. (3.00,2.74)
  node[pos=.55,above=2pt,font=\small\sffamily,text=serfrenoise]
    {Re-noising};

\draw[back] (10.30,3.02)
  .. controls (8.95,3.48) and (6.18,3.69) .. (5.85,3.04);

\draw[back] (10.32,2.92)
  .. controls (9.72,3.13) and (9.00,3.10) .. (8.70,2.74);

\serftile{1.15}{3.48}{1.80}{1}
\serftile{11.15}{3.48}{1.80}{0}

\node[label,anchor=north] at (1.15,2.44) {$X_0$};
\node[label,anchor=north] at (11.15,2.44) {$X_1$};

% Reward gradient at the endpoint.
\fill[serfreward] (12.30,3.66) circle (2.8pt);
\draw[-{Latex[length=2.6mm,width=1.85mm]},
  serfreward,line width=1.65pt]
  (12.30,3.66) -- (13.18,3.929);

\node[text=serfreward,anchor=north west] at (12.13,3.08)
  {$g=\nabla R(X_1)$};

% The same clean image at several noise levels.
\serftile{3.00}{1.86}{1.60}{.80}
\serftile{5.85}{2.16}{1.60}{.49}
\serftile{8.70}{1.86}{1.60}{.22}

\node[anchor=north] at (3.00,.98) {$X_{t_1}$};
\node[anchor=north] at (5.85,1.28) {$X_{t_2}$};
\node[anchor=north] at (8.70,.98) {$X_{t_3}$};

% Head-to-tail vector addition.
\foreach \x/\y/\dx/\dy/\i in {
  3.00/1.86/.408/.55/1,
  5.85/2.16/.552/-.25/2,
  8.70/1.86/.408/-.56/3
} {
  \coordinate (A) at (\x+.87,\y-.06);
  \coordinate (B) at (\x+.87+\dx,\y-.06+\dy);
  \coordinate (C) at (\x+1.446+\dx,\y+.1164+\dy);

  \fill[serftarget,opacity=.07]
    (A) -- (B) -- (C) -- cycle;

  \draw[-{Latex[length=1.65mm,width=1.2mm]},
    line width=1.15pt,serftarget]
    (A) -- (C);

  \draw[-{Latex[length=1.65mm,width=1.2mm]},
    line width=1.15pt,serfref]
    (A) -- (B);

  \draw[-{Latex[length=1.8mm,width=1.3mm]},
    line width=1.25pt,serfreward]
    (B) -- (C);
}

% Label only the central addition.
\node[font=\footnotesize,text=serftarget,anchor=south,inner sep=1pt]
  at (7.27,2.24) {$v^{\mathrm{target}}$};
\node[font=\footnotesize,text=serfref,anchor=north,inner sep=1pt]
  at (6.98,1.78) {$v^{\mathrm{ref}}$};
\node[font=\footnotesize,text=serfreward,anchor=north,inner sep=1pt]
  at (7.61,1.79) {$g$};

% Time axis.
\draw[serfmuted,line width=.65pt]
  (1.15,.37) -- (11.15,.37);
\draw[serfmuted,line width=.65pt]
  (1.15,.29) -- (1.15,.45);
\draw[serfmuted,line width=.65pt]
  (11.15,.29) -- (11.15,.45);

\foreach \x in {3.00,5.85,8.70} {
  \draw[serfmuted,line width=.65pt]
    (\x,.32) -- (\x,.42);
}

\node[anchor=north] at (1.15,.20) {$t=0$};
\node[anchor=north] at (11.15,.20) {$t=1$};

\node[font=\small,align=center,anchor=east,rounded corners=4pt,
  draw=serftarget!30,fill=serftarget!3,
  line width=.65pt,inner sep=6pt]
  at (14.75,1.25)
  {$\textcolor{serftarget}{v^{\mathrm{target}}(X_t)}=$\\[3pt]
   $\textcolor{serfref}{v^{\mathrm{ref}}(X_t)}
    +\textcolor{serfreward}{\nabla R(X_1)}$};

\pgfresetboundingbox
\path[use as bounding box]
  (.23,-.30) rectangle (14.94,5.20);

\end{tikzpicture}
\endgroup

%% file: appendix_rebuttal.tex
\appendix

\startcontents[appendix]

{\LARGE\bfseries Appendix\par}
\vspace{2em}

\printcontents[appendix]{}{1}{
    \setcounter{tocdepth}{2}
}

\newpage

\section{Theoretical analysis}
\label{sec:proofs}

Throughout the proofs, $b$ denotes the SDE drift and
$\bar b=\operatorname{sg}(b)$ its frozen counterpart.
For the velocity parameterization in
\Cref{eq:velocity_drift_backwards}, $b_t(x)=2v_t(x)-x/t$;
thus $X^b$ denotes the same process written as $X^v$
in the main text.

\subsection{Stochastic optimal control proofs}
\label{sub:soc_proofs}
For a diffusion process satisfying
\begin{align}
    \diff X_t^b = b_t(X_t) \diff t + \sigma_t \diff W_t, \qquad X_0^b \sim p_0, \label{appeq:sde}
\end{align}
consider the optimization of a loss of the form
\begin{align}
J(\theta) = \mathbb E\biggl[\int_0^1 L(X_s^b, b_s, s) \diff s + R(X_1^b) \biggr],
\end{align}
where $\theta$ are the parameters underlying the diffusion drift $b_t$. Introducing the value function
\begin{align}
    V(x, t) = \mathbb E \left[ \int_t^1 L(X_s^b, b_s, s) \diff s + R(X_1^b) \bigm| X_t = x \right], \label{appeq:value_fn_def}
\end{align}
we obtain the following:

\begin{mdframed}[style=MyFrame2]
\begin{theorem}
\label{appthm:reward_max_surrogate_target}
    The gradient of the objective $J$ can be exactly written as
    \begin{align}
        \nabla_\theta J(\theta) = \mathbb E \left[ \int_0^1 \nabla_\theta b_t(X_t^b)^{\top} (\nabla_b L(X_t^b, b_t, t) + \nabla_x V(X_t^b, t)) \diff t \right].
        \label{appeq:gradient_exactly_integral}
    \end{align}
    Hence, maximizing $J$ through gradient ascent corresponds to minimizing the following surrogate objective through gradient descent
    \begin{align}
    \hat{\mathcal{L}}(\theta) = \frac 12\mathbb E \left[\int_0^1 \left\| b_t(X_t^{\bar b}) - \bar b_t(X_t^{\bar b}) - \nabla_b L(X_t^{\bar b}, \bar b_t, t) - \nabla_x V(X_t^{\bar b}, t)  \right\|^2 \diff t \right]. \label{appeq:surrogate_target}
    \end{align}
\end{theorem}
\end{mdframed}

\begin{proof}
    The idea is to use the Feynman-Kac equation twice, the first to obtain a PDE which $V$ satisfies, and the second to go from a PDE that $\nabla_\theta V$ satisfies to an equation for $\nabla_\theta V$. Firstly, from the definition of the value function~\Cref{appeq:value_fn_def} and the form of the SDE~\Cref{appeq:sde}, Feynman-Kac tells us that
    \begin{align}
        \frac{\partial}{\partial t} V(x, t) + b_t(x) \nabla_x V(x, t) + \frac 12 \sigma_t^2 \Delta V(x, t) + L(x, b_t(x), t) = 0,
    \end{align}
    with terminal condition $V(x, 1) = R(x)$. We take the gradient of this equation with respect to $\theta$. Let $\nabla_\theta V = U$. Then, we obtain
    \begin{align}
        \frac{\partial}{\partial t} U(x, t) &+ \nabla_\theta b_t(x)^\top \nabla_x V(x, t) + b_t(x) \nabla_x U(x, t) \\
        &+ \frac 12 \sigma_t^2 \Delta U(x, t) + \nabla_\theta b_t(x)^\top \nabla_b L(x, b_t(x), t) = 0.
    \end{align}
    Moreover, $U(x, 1) = \nabla_\theta V(x, 1) = \nabla_\theta R(x) = 0$. From Feynman-Kac, this means we can express $U$ as exactly
    \begin{align}
        U(x, t) = \mathbb E \left[ \int_t^1 \nabla_\theta b_s(X_s^b)^\top ( \nabla_b L(X_s^b, b_s, s) + \nabla_x V(X_s^b, s)) \diff s \bigm| X_t = x \right].
    \end{align}
    Then, we can represent $J(\theta) = \mathbb E_{X_0 \sim p_0}[V(X_0, 0)]$, so since $p_0$ is independent of $\theta$, $\nabla_\theta J(\theta) = \mathbb E_{X_0 \sim p_0}[U(X_0, 0)]$ which gives exactly~\Cref{appeq:gradient_exactly_integral} by plugging in the above expression for $U$.

    To show that the gradient of the surrogate loss $\hat{\mathcal L}$ is the negation of $\nabla_\theta J(\theta)$, we simply differentiate $\hat{\mathcal L}$. Since sampling is done under a stop gradient distribution, we can simply differentiate the interior of the integral, which is
    \begin{align}
        & \qquad \nabla_\theta \tfrac12 \| b_t(X_t^{\bar b}) - \bar b_t(X_t^{\bar b}) - \nabla_b L(X_t^{\bar b}, \bar b_t, t) - \nabla_x V(X_t^{\bar b}, t) \|^2 \\
        &= \nabla_\theta b_t(X_t^b)^\top (b_t(X_t^{b}) - b_t(X_t^b) - \nabla_b L(X_t^b, b_t, t) - \nabla_x V(X_t^b, t)) \\
        &= - \nabla_\theta b_t(X_t^b)^\top (\nabla_b L(X_t^b, b_t, t) + \nabla_x V(X_t^b, t)).
    \end{align}
    Integrating this yields $\nabla_\theta \hat{\mathcal L}(\theta) = - \nabla_\theta J(\theta)$.
\end{proof}

Taking $L=0$ gives the reward-gradient identity in drift space.
By \Cref{eq:velocity_drift_backwards},
$\nabla_\theta b_t^\theta(x)=2\nabla_\theta v_t^\theta(x)$.
At $v^\theta=\bar v$, this factor of two matches the one from
differentiating the squared residual in \Cref{eq:reward_loss},
which proves \Cref{thm:reward_surrogate_objective}.

Moreover, we may derive Adjoint Matching from this formula. In Adjoint Matching, we take $L = - \|b_t -  b_t^{\mathrm{ref}}\|^2/(2\sigma(t)^2)$, which when substituted into~\Cref{appthm:reward_max_surrogate_target} gives
\begin{align}
    \mathbb E_{\theta^-, t} \left\|\frac{(b_t(X_t^{\bar b}) - b_t^\mathrm{ref}(X_t^{\bar b}))}{\sigma(t)^2} - \nabla_x V(X_t^{\bar b}, t)  \right\|^2
\end{align}
This is equivalent to the basic adjoint matching loss, up to reweighting.

\subsection{The stochastic Jacobian}
\label{app:stochastic_jacobian}
Fix $\delta \in (0, 1)$. Assume throughout that
\(\bar b\in C([\delta,1];C^2(\mathbb R^d;\mathbb R^d))\) and $\nabla_x \bar b$  and $\nabla_x^2 \bar b$ are bounded on $[\delta, 1] \times \mathbb R^d$,
\(R\in C_b^1(\mathbb R^d)\), and \(\sigma_s\) is continuous on $[\delta, 1]$ and
independent of the state. Under these assumptions, the SDE generates a
\(C^1\) stochastic flow on $t \in [\delta, 1]$ \citep[Chapter~4]{kunita1990stochastic}.

Given a realization \(\xi\) of the Brownian motion and \(X_t=x\) with $t \in [\delta, 1]$, define
\begin{align}
    T_{s,t}^\xi(x):=X_s^{t,x,\bar b}(\xi),
    \qquad s\in[t,1].
\end{align}
For fixed \(\xi\), \(T_{s,t}^\xi\) is a deterministic function of \(x\).
Its stochastic Jacobian is
\begin{align}
    J_{s,t}^\xi(x):=\nabla_xT_{s,t}^\xi(x).
\end{align}
Since \(\sigma_s\) is independent of the state, the stochastic Jacobian
satisfies
\begin{align}
    \frac{\diff}{\diff s}J_{s,t}^\xi(x)
    =
    \nabla\bar b_s\!\left(T_{s,t}^\xi(x)\right)J_{s,t}^\xi(x),
    \qquad
    J_{t,t}^\xi(x)=I.
\end{align}
Differentiation and expectation can be interchanged under the assumptions
above. Applying the chain rule pathwise gives
\begin{align}
    \nabla_x\mathbb E\!\left[
        R(X_1^{\bar b})\mid X_t^{\bar b}=x
    \right]
    =
    \mathbb E_\xi\!\left[
        J_{1,t}^\xi(x)^\top
        \nabla R\!\left(T_{1,t}^\xi(x)\right)
    \right].
\end{align}

We can furthermore show that the endpoint Jacobian approaches the identity
as \(t\to1\).

\begin{mdframed}[style=MyFrame2]
\begin{proposition}
Under the assumptions above, for all $t \in [\delta, 1]$,
\begin{align}
    J_{1,t}^\xi(x)=I+\mathcal O(1-t)
\end{align}
almost surely.
\end{proposition}
\end{mdframed}

\begin{proof}
The differential equation for the stochastic Jacobian implies
\begin{align}
    J_{s,t}^\xi(x)
    =
    I+\int_t^s
    \nabla\bar b_r\!\left(T_{r,t}^\xi(x)\right)
    J_{r,t}^\xi(x)\,\diff r.
\end{align}
Since $\nabla_x \bar b$ is bounded, there exists \(L<\infty\) such that
\begin{align}
    \sup_{r,x}\|\nabla\bar b_r(x)\|_{\mathrm{op}}\leq L.
\end{align}
Gr\"onwall's inequality implies $\|J_{r, t}^\xi(x)\|_{\mathrm{op}} \leq e^{L(r - t)}$. Consequently,
\begin{align}
    \|J_{1,t}^\xi(x)-I\|_{\mathrm{op}}
    \leq
    L\int_t^1\|J_{r,t}^\xi(x)\|_{\mathrm{op}}\,\diff r
    \leq
    e^{L(1-t)}-1
    =
    \mathcal O(1-t),
\end{align}
which proves the result.
\end{proof}

In practice, the Lipschitz constant for $\bar b_t = 2v_t - x/t$ depends on the Lipschitz constant of the network $v_t$, and of the linear term which is bounded by $1/\delta$. Hence this bound is only informative near the data endpoint $t = 1$, which provides motivation for sampling $t \sim p(t) = 2t$, favoring low-noise states.

\paragraph{Scalar gradient-transport approximations.}
For $b_s(x)=2v_s(x)-x/s$, the stochastic-flow Jacobian satisfies
\[
\partial_s J_{s,t}
=
\left(2\nabla_xv_s(X_s)-s^{-1}I\right)J_{s,t},
\qquad J_{t,t}=I.
\]
We compare three scalar approximations $J_{1,t}\approx c(t)I$.

The identity approximation uses $c(t)=1$.
The factor labelled ``one-step'' follows by dropping
$\nabla_xv_s$ while retaining the known linear drift:
$\partial_sJ_{s,t}=-s^{-1}J_{s,t}$ gives
$J_{s,t}=(t/s)I$ and hence $c(t)=t$.

For a Gaussian reference distribution
$p_{\mathrm{data}}=\mathcal N(\mu,\tau^2I)$, write
$D_s=s^2\tau^2+(1-s)^2$. Its flow-matching velocity satisfies
\[
\nabla_xv_s^{\mathrm{ref}}
=
\frac{s\tau^2-(1-s)}{D_s}I,
\qquad
\nabla_xb_s^{\mathrm{ref}}
=
\left(\frac{D_s'}{D_s}-\frac1s\right)I.
\]
Therefore,
\[
J_{1,t}^{\mathrm{ref}}
=
\exp\left[
\int_t^1\left(\frac{D_s'}{D_s}-\frac1s\right)\diff s
\right]I
=
\frac{t\tau^2}{t^2\tau^2+(1-t)^2}I.
\]
Taking $\tau^2=1$ yields the Gaussian factor used in our ablation.
These factors approximate gradient transport and are distinct
from the fixed-point correction $\Gamma_t$ in
\S\ref{sub:fixed_point}.

\subsection{Derivation of the simple endpoint estimate}
\label{sub:simp_endpoint_derivation}

Here, all laws are under the sampling drift $\bar b$.

\begin{mdframed}[style=MyFrame2]
\begin{proposition}
    Suppose $X_1\mid X_t=x\sim\mathcal N(T(x),\sigma^2 I)$, where
    $\sigma>0$ is independent of $x$ and the symmetric part of
    $J_T(x)=\nabla_x T(x)$ is positive definite. Then
    $\hat R_t(x)=\mathbb E[\nabla R(X_1)\mid X_t=x]$ is positively
    aligned with $\nabla_x\mathbb E[R(X_1)\mid X_t=x]$ whenever
    $\hat R_t(x)\ne 0$. Under Gaussian corruption, the Jacobian
    condition holds whenever the posterior covariance has full rank.
\end{proposition}
\end{mdframed}

\begin{proof}
    If $\mathrm{Law}(X_1 \mid X_t = x) = \mathcal N(T(x), \sigma^2 I)$, then we may write $X_1 = T(X_t) + \sigma Z$ for $Z \sim \mathcal N(0, I)$. Thus,
    \begin{align}
        \nabla_x \mathbb E[R(X_1) \mid X_t = x] &= \nabla_x \mathbb E[R(T(x) + \sigma z)] = \mathbb E [\nabla_x T(x)^\top \nabla R(T(x) + \sigma z) ] \\
        &= J_T(x)^\top  \mathbb E[\nabla R(X_1) \mid X_t = x].
    \end{align}
    This is exactly $\mathbb E[\nabla R(X_1) \mid X_t = x]$ when $J_T(x) = I$, i.e. $T(x)$ is locally a shift. In particular, when $T(x) = x + c$ then the simple endpoint estimate is exact for all $x$.

    In general, when $J_T(x) \neq I$, we can still calculate when the simple endpoint estimate is positively aligned with $\nabla_x \mathbb E[R(X_1) \mid X_t = x]$. The inner product between the two is
    \begin{align}
        &\qquad \mathbb E[\nabla R(X_1) \mid X_t = x]^\top J_T(x)^\top \mathbb E[\nabla R(X_1) \mid X_t = x] \\
        &= \mathbb E[\nabla R(X_1) \mid X_t = x]^\top \left( \frac{J_T(x) + J_T(x)^\top}2 \right) \mathbb E[\nabla R(X_1) \mid X_t = x] > 0
    \end{align}
    when $(J_T + J_T^\top)/2$ is positive definite and $\mathbb E[\nabla R(X_1) \mid X_t = x] \neq 0$, showing the first part.

    We now proceed to show that $J_T$ has positive definite symmetric part if $T(x) = \mathbb E[X_1 \mid X_t = x]$, under Gaussian backwards law. Indeed, under the assumption $\mathrm{Law}(X_t\mid X_1) = \mathcal N(a X_1, s_t^2 I)$ with $a > 0$, Tweedie's gives that $\nabla_x \log p_{1|t}(y|x) = \frac{a}{s_t^2}(y - T(x))$. This gives
    \begin{align}
        \nabla_x \mathbb E[ X_1 \mid X_t = x] &= \nabla_x \int y p_{1|t}(y|x) \diff y \\
        &= \frac{a}{s_t^2}\int y(y - T(x))^\top p_{1|t}(y|x) \diff y \\
        &= \frac{a}{s_t^2} \int (y - T(x))(y - T(x))^\top p_{1|t}(y|x) \diff y \\
        &= \frac{a}{s_t^2} \mathrm{Cov}(X_1 - T(x) \mid X_t = x) \succeq 0.
 \end{align}
 This is symmetric and positive-definite whenever $\mathrm{Cov}(X_1 \mid X_t = x) \succ 0$. From our assumption this is $\sigma^2 I \succ 0$.
\end{proof}
These assumptions imply a jointly Gaussian state--endpoint law; the proposition serves as theoretical motivation for our design, rather than a general guarantee for the implemented algorithm.

Let $\mu_t(x)=\mathbb E[\nabla R(X_1^{\bar v})\mid X_t^{\bar v}=x]$.
Replacing the sampled endpoint gradient in \Cref{eq:simple_reward}
by its conditional mean gives
\begin{align}
    \mathcal L_{\mathrm{simple-expected-reward}}(\theta;\bar v)
    = \mathbb E_{t\sim p(t)}
    \left\|v_t^\theta(X_t^{\bar v})-\bar v_t(X_t^{\bar v})
    -\mu_t(X_t^{\bar v})\right\|^2.
    \label{appeq:simple_expected_reward}
\end{align}

\begin{mdframed}[style=MyFrame2]
\begin{lemma}\label{lem:endpoint_target_variance}
With $\bar v$ and the sampling law held fixed, the losses in
\Cref{eq:simple_reward,appeq:simple_expected_reward} have the same
gradient with respect to $\theta$.
\end{lemma}
\end{mdframed}

\begin{proof}
Let $A=v_t^\theta(X_t^{\bar v})-\bar v_t(X_t^{\bar v})$,
$Y=\nabla R(X_1^{\bar v})$, and
$\mu=\mathbb E[Y\mid t,X_t^{\bar v}]$.
The cross term vanishes because $A-\mu$ is determined by
$(t,X_t^{\bar v})$ and $\mathbb E[Y-\mu\mid t,X_t^{\bar v}]=0$.
Thus
\[
    \mathbb E\|A-Y\|^2
    = \mathbb E\|A-\mu\|^2+\mathbb E\|Y-\mu\|^2.
\]
The final term is independent of $\theta$ under the frozen sampling law.
\end{proof}

We now prove that under \Cref{positve-aligned} optimizing \Cref{eq:simple_reward} increases the reward.

\begin{mdframed}[style=MyFrame2]
\begin{proposition}
Under the assumptions of \Cref{positve-aligned}, the velocity-field
perturbation $h_t(x)=p(t)\hat G_t^{\bar v}(x)$ is a strict local
reward-ascent direction at $\bar v$ whenever $h$ is nonzero on a set
of positive measure under $\diff t\,p_t^{\bar v}(\diff x)$.
\end{proposition}
\end{mdframed}

\begin{proof}
The reward-gradient identity in \S\ref{sub:soc_proofs} gives
\[
\left.\frac{\diff}{\diff\eta}
\mathcal L_{\mathrm{reward}}(\bar v+\eta h)\right|_{\eta=0}
=
2\int_0^1 p(t)\,
\mathbb E\!\left[
\left\langle G_t^{\bar v}(X_t^{\bar v}),
\hat G_t^{\bar v}(X_t^{\bar v})\right\rangle
\right]\diff t.
\]
The integrand is nonnegative by \Cref{positve-aligned} and positive
where $h$ is nonzero. Hence the derivative is strictly positive.
\end{proof}

\subsection{Combining regression targets}
\label{sub:combine_targets}
\begin{mdframed}[style=MyFrame2]
\begin{lemma}
At $v^\theta=\bar v$, the gradients of
$\mathcal L_{\mathrm{simple-reward}}$ and
$\mathcal L_{\mathrm{simple-regularization}}$ sum to the gradient
of $\mathcal L_{\nameshort}^{\mathrm{SDE}}$, with $\bar v$ held fixed.
\end{lemma}
\end{mdframed}

\begin{proof}
All three losses use the same frozen sampling law. At
$v^\theta=\bar v$, the residual of the reward loss is
$-\nabla R(X_1^{\bar v})$. Adding it to the reference residual
$v_t^\theta(X_t^{\bar v})-v_t^{\mathrm{ref}}(X_t^{\bar v})$
gives the residual of the combined loss. Each squared loss has
the same prefactor and is differentiated through the same
$v_t^\theta(X_t^{\bar v})$, so the gradients add as claimed.
\end{proof}

\begin{mdframed}[style=MyFrame2]
\begin{proposition}
Define the single-sample gradient
\[
    g_t(x,x_1)=\nabla_\theta
    \left\|v_t^\theta(x)-v_t^{\mathrm{ref}}(x)-\nabla R(x_1)\right\|^2,
\]
with $x$ and $x_1$ held fixed, and assume it has a finite second
moment under the SDE joint law. Draw IID times $t_i\sim p(t)$
independently of the SDE trajectory. Consider the two estimators:
\begin{itemize}[leftmargin=1.2em, itemsep=0em]
    \item $g_{\mathrm{traj}}=\frac1m\sum_{i=1}^m
    g_{t_i}(X_{t_i},X_1)$, where $(X_t)_{t\in[0,1]}$ is an SDE
    trajectory under $\bar b$.

    \item $g_{\mathrm{corr}}=\frac1m\sum_{i=1}^m
    g_{t_i}(X'_{t_i},X_1)$, where $X_1\sim p_1^{\bar b}$ and
    $X'_{t_i}\sim p_{t_i|1}^{\bar b}(\cdot\mid X_1)$ are drawn
    independently given the endpoint and times.
\end{itemize}
Both estimators are unbiased for
$\nabla_\theta\mathcal L_{\nameshort}^{\mathrm{SDE}}(\theta;\bar v)$,
and $\mathrm{Cov}(g_{\mathrm{corr}})\preceq
\mathrm{Cov}(g_{\mathrm{traj}})$.
\end{proposition}
\end{mdframed}

\begin{proof}
Each summand in either estimator has the law of $g_t(X_t,X_1)$
for $t\sim p(t)$ and $(X_t,X_1)\sim p_{t,1}^{\bar b}$. Thus both
estimators are unbiased and their individual summands have the same
covariance.

Let
\[
    M=\int_0^1 g_s(X_s,X_1)p(s)\,\diff s.
\]
Conditional on the full SDE trajectory, the summands of
$g_{\mathrm{traj}}$ are independent and have common mean $M$.
Conditional on $X_1$, the summands of $g_{\mathrm{corr}}$ are
independent and have common mean $\mathbb E[M\mid X_1]$.
By the law of total covariance, the cross-covariance between any
two distinct summands is therefore $\mathrm{Cov}(M)$ for
$g_{\mathrm{traj}}$ and $\mathrm{Cov}(\mathbb E[M\mid X_1])$
for $g_{\mathrm{corr}}$. The individual covariance terms cancel,
leaving
\begin{align}
    \mathrm{Cov}(g_{\mathrm{traj}})-\mathrm{Cov}(g_{\mathrm{corr}})
    &=\frac{m-1}{m}
    \left(\mathrm{Cov}(M)-\mathrm{Cov}(\mathbb E[M\mid X_1])\right)
    \notag\\
    &=\frac{m-1}{m}\mathbb E[\mathrm{Cov}(M\mid X_1)]
    \succeq 0.
    \label{appeq:random_term}
\end{align}
\end{proof}

For $m\geq 2$, the independent-conditional estimator has strictly
smaller total variance, measured by the trace of its covariance,
whenever $M$ is not almost surely determined by $X_1$.

\subsection{Fixed-point analysis}
\label{sub:fixed_point}

Throughout, $R$ refers to the reward after absorbing any constant factors, and we assume $\mathbb E\|\nabla R(X_1)\|^2<\infty$. We call $v$ a \emph{fixed point} of \nameshort if $v$ minimizes \Cref{eq:SERF_objective} over all measurable velocity fields when the sampling model is $\bar v = v$, i.e.\ if one exact \nameshort update leaves $v$ unchanged.

\begin{mdframed}[style=MyFrame2]
\begin{lemma}[Self-consistency]
\label{lem:fixed_point}
Let $q^v$ denote the law of $(X_t, X_1)$ with $X_1\sim p_1^v$,
$X_0\sim\mathcal N(0,I)$ independent, and $X_t = tX_1+(1-t)X_0$.
Then $v$ is a fixed point of \nameshort if and only if, for
$p$-almost every $t\in(0,1)$ and almost every $x$,
\begin{align}
    v_t(x) = v_t^{\mathrm{ref}}(x) + \mathbb E_{q^v}\!\left[\nabla R(X_1)\mid X_t = x\right].
    \label{appeq:fixed_point}
\end{align}
\end{lemma}
\begin{proof}
For fixed $\bar v = v$, \Cref{eq:SERF_objective} is a weighted least-squares
regression of $v_t(X_t)$ onto $v_t^{\mathrm{ref}}(X_t)+\nabla R(X_1)$ with
$(t,X_t,X_1)\sim p(t)\,q^v$. Its minimizer is the conditional expectation of
the target given $(t, X_t)$, which is \Cref{appeq:fixed_point}. Since $p(t)>0$
on $(0,1]$ and $X_t$ has a positive density for $t<1$, this determines $v$
almost everywhere.
\end{proof}
\end{mdframed}

\Cref{appeq:fixed_point} is a self-consistency equation since the sampling distribution $q^v$ depends on $v$ itself. The fixed point is independent of our choice of $p(t)$; the sampling distribution instead serves to place emphasis on learning the velocity over certain parts of the trajectory.

\xhdr{Linear rewards}
In the linear case with $R(x)=a^\top x$, $\nabla R(x) = a$ and \Cref{appeq:fixed_point} gives
\begin{align}
    v_t(x) = v_t^{\mathrm{ref}}(x) + a.
\end{align}
This is true for any distribution and any re-noising kernel used. We show that if the data distribution is an isotropic Gaussian, then the endpoint distribution is a reward tilt under $\gamma R$, for some constant $\gamma$.

\begin{mdframed}[style=MyFrame2]
\begin{proposition}
\label{prop:fixed_point_gaussian}
Let $p_{\mathrm{data}}=\mathcal N(\mu, \sigma^2 I)$ and $R(x)=a^\top x$. Then the
fixed point of \nameshort has endpoint distribution
\begin{align}
    p_1^v = \mathcal N\!\left(\mu + \sigma^2 \gamma a,\; \sigma^2 I\right)
    \;\propto\; p_{\mathrm{data}}(x)\, e^{\gamma R(x)},
    \qquad
    \gamma = \frac{1}{\sigma} \int_0^1 \frac{\diff r}{s_r},
\end{align}
where $s_t^2 = t^2\sigma^2+(1-t)^2$. That is, $p_1^v$ is the reward-tilted data distribution with modified reward strength $\gamma$.
\end{proposition}
\end{mdframed}

\begin{proof}
For a linear interpolant, $X_t\sim\mathcal N(t\mu, s_t^2 I)$ and
\[
    v_t^{\mathrm{ref}}(x) = \mathbb E[X_1-X_0\mid X_t=x]
    = \mu + \frac{\dot s_t}{s_t}\,(x - t\mu),
\]
since $\mathrm{Cov}(X_1-X_0, X_t) = (t\sigma^2-(1-t))I = s_t\dot s_t I$. The \nameshort fixed point $v_t = v_t^{\mathrm{ref}} + a$ is affine in $x$. Consider $X_0^v \sim \mathcal N(0, I)$, $\diff X_t^v = v_t(X_t^v) \diff t$. Then $m_t = \mathbb E[X_t^v]$ satisfies
\[
\frac{\diff m_t}{\diff t} = \frac{\diff}{\diff t}\mathbb E[X_t^v] = \mathbb E[v_t(X_t^v)] = v_t(m_t)
\]
since $v_t$ is affine. Letting $\delta_t = X_t^v - m_t$, $\tfrac{\diff}{\diff t} \delta_t = \tfrac{\dot s_t}{s_t} \delta_t$. This is true both under the reference velocity and the \nameshort fixed point, hence their endpoint distributions are simply shifted versions.

To calculate the displacement of the \nameshort endpoint distribution, we solve the ODE for $m_t$, starting from $m_0 = 0$. This can be done by noticing $\partial_t(m_t - t \mu) = a+\tfrac{\dot s_t}{s_t}(m_t - t \mu)$, so $\partial_t \big( (m_t - t \mu)/s_t \bigr) = a/s_t$. Integrating gives
\[
\frac{m_1 - \mu}{\sigma} = a \int_0^1 \frac{\diff r}{s_r},
\]
This gives $m_1 = \mu + \sigma^2 \gamma a$, as in our expression for $p_1^v$. To see $p_1^v$ is the reward tilt with weight $\gamma R(x)$, it suffices to calculate the likelihood, which is proportional to
\[
e^{-\|x - \mu - \sigma^2 \gamma a\|^2/(2 \sigma^2)} \propto e^{-\|x - \mu\|^2/(2 \sigma^2)} e^{\gamma a^{\top}x} \propto p_{\mathrm{data}}(x) e^{\gamma R(x)}.
\]
\end{proof}

For an anisotropic covariance $\Sigma\neq \sigma^2I$, the same computation applies to each eigendirection of
$\Sigma$ with its own strength $\gamma_i$. We still obtain a tilt by a linear reward, but not by a multiple of $R$.

\xhdr{General Gaussian target} For a Gaussian target $p_{\mathrm{data}} = \mathcal N(\mu, \sigma^2 I)$, we can obtain a regression target that differs only from the \nameshort target by a time reweighting, and which has the exact reward tilt as a fixed point.

\begin{mdframed}[style=MyFrame2]
\begin{proposition}\label{prop:gamma_exact}
Let $p_{\mathrm{data}}=\mathcal N(\mu,\sigma^2I)$, $p^\star = p_1^{b^{\star}}\propto p_{\mathrm{data}}\,e^{R}$, and
\[
\Gamma_t=\frac{(1-t)\,\sigma^2}{(1-t)^2+t^2\sigma^2}.
\]
Let $v^\star$ be the flow-matching velocity for $p^\star$. Then, for every reward $R$ with $\nabla Re^{R}$ integrable,
\[
v^\star_t(x)=v^{\mathrm{ref}}_t(x)+\Gamma_t\,\mathbb E_{p^\star}\!\left[\nabla R(X_1)\mid X_t=x\right].
\]
Hence $v^\star$ is a fixed point of \nameshort with target $v^{\mathrm{ref}}_t(X_t)+\Gamma_t\nabla R(X_1)$.
\end{proposition}
\end{mdframed}

\begin{proof}
Let $p_{1|t}$ and $p_{1|t}^\star$ be the laws of $X_1$ given $X_t=x$ under $X_1\sim p_{\mathrm{data}}$ and $X_1\sim p^\star$, with means $m_t(x)$ and $m^\star_t(x)$.
The linear interpolant $X_t = tX_1 + (1 - t)X_0$ shows that any flow-matching velocity satisfies $v_t(x)=\mathbb E[X_1-X_0\mid X_t=x]=(\mathbb E[X_1\mid X_t=x]-x)/(1-t)$, hence
\begin{align}
v^\star_t(x)-v^{\mathrm{ref}}_t(x)=\frac{m^\star_t(x)-m_t(x)}{1-t}. \label{appendix:v_in_terms_of_m}
\end{align}
We can relate the two distributions $p_{1|t}$ and $p_{1|t}^\star$ through
\begin{align}
p_{1|t}^\star(x_1 \mid x_t) \propto p_{t|1}^\star(x_t \mid x_1) p_1^\star(x_1) \propto p_{t|1}(x_t \mid x_1) p_1(x_1) e^{R(x_1)} \propto p_{1|t}(x_1 \mid x_t) e^{R(x_1)},
\end{align}
hence the latter is a reward tilt of the former, since the re-noising kernels $p_{t|1}$ and $p_{t|1}^\star$ are identical. Hence we can write
\begin{align}
m_t^\star(x) - m_t(x) &= \frac{\mathbb E_{p_{1|t}}[X_1 e^{R(X_1)} \mid X_t = x]}{\mathbb E_{p_{1|t}}[e^{R(X_1)} \mid X_t = x]} - m_t(x) \\
&= \frac{\mathbb E_{p_{1|t}}[(X_1 - m_t(x)) e^{R(X_1)} \mid X_t = x]}{\mathbb E_{p_{1|t}}[e^{R(X_1)} \mid X_t = x]}.
\end{align}
For $p_1$ Gaussian, the conditional law $p_{1|t}$ is also Gaussian since $(X_t, X_1)$ are jointly Gaussian, with mean defined to be $m_t(x)$ and variance $(1 - t)^2\sigma^2/((1 - t)^2 + t^2 \sigma^2)$. Applying Stein's lemma to the term in the numerator thus gives
\begin{align}
m_t^{\star}(x) - m_t(x) &= (1 - t) \Gamma_t \frac{\mathbb E_{p_{1|t}}[\nabla R(X_1) e^{R(X_1)} \mid X_t = x]}{\mathbb E_{p_{1|t}}[e^{R(X_1)} \mid X_t = x]} \\
&= (1 - t)\Gamma_t\mathbb E_{p_{1|t}^\star}[\nabla R(X_1) \mid X_t = x].
\end{align}
Substituting \Cref{appendix:v_in_terms_of_m} and dividing by $(1 - t)$ yields the claim for the relation between $v_t^\star$ and $v_t^{\mathrm{ref}}$. Regressing against $v_t^{\mathrm{ref}}(X_t) + \Gamma_t \nabla R(X_1)$ with $X_1 \sim q^v$ has the conditional expectation as the minimizer. If $v = v^\star$, then the endpoint is distributed as $p^\star$, so $v^\star$ is a minimizer.

\end{proof}

Our choice to use a constant time weighting in front of the reward is motivated by the observation that the weighting $\Gamma_t$ vanishes at $t = 1$, the data end. For rewards based on perceptual quality this suppresses the reward signal at times near $1$, where finer features are generated~\citep{liu2025value,lee2026reward}.

\section{Connections to prior work}
\label{sec:related-work}
Reward fine-tuning methods for diffusion and flow models either
use reward values, as in DDPO, Flow-GRPO, and FPO, or exploit
reward gradients, as in DRaFT, Adjoint Matching, and RSM.
They share the objective of maximizing expected terminal reward,
$J(\theta)=\mathbb E_{X_1\sim p_1^\theta}[R(X_1)]$,
typically with reference regularization as
in~\Cref{eq:soc_problem}.

For non-differentiable rewards, REINFORCE uses likelihood gradients.
For a discrete stochastic sampler on
$0=t_0<\cdots<t_N=1$, the path
$\tau=(X_{t_0},\ldots,X_{t_N})$ has joint density
\[
    p_\theta(\tau)
    =
    p_0(X_0)\prod_{k=0}^{N-1}
    p_\theta(X_{t_{k+1}}\mid X_{t_k}).
\]
Its transition densities can be tractable even when its endpoint
marginal $p_1^\theta$ is not.
The reward gradient admits both endpoint and path representations:
\[
\begin{aligned}
    \nabla_\theta J(\theta)
    &=
    \mathbb E_{X_1\sim p_1^\theta}\!\left[
        R(X_1)\nabla_\theta\log p_1^\theta(X_1)
    \right] \\
    &=
    \mathbb E_{\tau\sim p_\theta}\!\left[
        R(X_1)\sum_{k=0}^{N-1}
        \nabla_\theta\log
        p_\theta(X_{t_{k+1}}\mid X_{t_k})
    \right].
\end{aligned}
\]
Trajectory methods such as DDPO and Flow-GRPO exploit tractable
stochastic-transition likelihoods, whereas FPO approximates the
endpoint log-likelihood using an evidence lower bound.

For differentiable rewards, the reward-ascent target in the SOC
decomposition~\Cref{eq:reward_loss} instead admits the
gradient-transport representation
\[
\begin{aligned}
    g_t^{\bar b}(x)
    &:=
    \nabla_x\mathbb E\!\left[
        R(X_1^{\bar b})\mid X_t^{\bar b}=x
    \right] \\
    &=
    \mathbb E\!\left[
        (J_{1,t}^{\bar b})^\top\nabla R(X_1^{\bar b})
        \mid X_t^{\bar b}=x
    \right],
\end{aligned}
\]
where $J_{1,t}^{\bar b}$ is the stochastic Jacobian defined
in~\S\ref{app:stochastic_jacobian}.
Computing this transport requires differentiating through the
dynamics. Efficient methods approximate it in different ways;
\nameshort uses $J_{1,t}^{\bar b}\approx I$ and independently
re-noised training states.

We discuss these method families below.
\Cref{tab:baseline_comparison} summarizes reward evaluation,
gradient construction, and training-state construction across
differentiable methods.

\paragraph{Non-differentiable rewards}
Trajectory methods cast the generative SDE as a Markov decision
process with Gaussian transitions and apply policy gradients to
its steps~\citep{black2024ddpo,fan2023dpok,flowgrpo,xue2025dancegrpo}.
In contrast, endpoint and forward-process methods use complete
generations to construct pretraining-like losses.
FPO uses an evidence lower bound to estimate the intractable
endpoint likelihood~\citep{mcallister2025flow,choi2026rethinking};
other methods derive their losses from stochastic optimal
control~\citep{bergmeister2026reinforceadjointmatchingscaling}
or online reward-weighted regression~\citep{fan2025online}.
Related approaches use advantage-weighted
matching~\citep{xue2026advantage}, contrastive forward-process
training~\citep{zheng2026diffusionnft}, rollout-policy
regularization~\citep{kveton2026advantageflow}, or reward-weighted
velocity regression~\citep{choi2026scalingreinforcementlearningdiffusion}.

In text-to-image experiments, RAM reaches a matched reward in
up to $50\times$ fewer training steps than
Flow-GRPO~\citep{bergmeister2026reinforceadjointmatchingscaling}.
Methods that re-noise generated samples to construct training
states are particularly similar to \nameshort.
FDFO also applies an endpoint reward direction to intermediate
velocity updates without propagating it through the sampling
Jacobian, but estimates that direction from paired generations
rather than reward gradients~\citep{mcallister2026fdfo}.

\paragraph{Differentiable rewards}
AlignProp and DRaFT optimize differentiable rewards by
backpropagating through sampling~\citep{prabhudesai2023aligning,
clark2024directly}.
DRaFT-$K$ restricts backpropagation to the final $K$ steps,
while DRaFT-LV averages gradients over independently noised
copies of a generated sample~\citep{clark2024directly}. ReFL evaluates the reward on a predicted clean image $\hat x_1 = x_t + (1 -t) v_t^\theta(x_t)$ at an
intermediate step and differentiates through that
prediction~\citep{xu2023imagereward}.
These methods still differentiate through at least one sampling
or prediction step.
ReFL further restricts training to the final 10 of 40 denoising steps \citep{xu2023imagereward}.

EasyTune \citep{tan2026easytune} targets motion representations, and evaluates the reward at intermediate noised states directly. For image generation, they note noisy states are difficult for the reward model to interpret, and recommend evaluating the reward at $\hat x_1$. Such one-step estimates can be blurry at high noise and lead to unreliable reward gradient predictions \citep{tao2025lpo, kang2026diffusion}. \nameshort instead evaluates the reward gradient at generated endpoints, and uses these gradients at re-noised states. 

Other methods simplify gradient propagation or construct shorter
backward surrogates~\citep{wu2024deep,liang2026leapalign,
wang2026exploring}, or use local denoising and
reconstruction~\citep{shen2025directlyaligningdiffusiontrajectory}.
Reward Sharpness-Aware Fine-Tuning studies reward-gradient
robustness~\citep{kim2026reward}.

VGG-Flow derives guidance from ODE optimal control and trains an
auxiliary network to estimate it~\citep{liu2025value}.
SQDF approximates soft-$Q$ guidance by evaluating the reward on
a clean-image estimate, without training a
critic~\citep{kang2026diffusion}.
Adjoint Matching constructs its training target by integrating
adjoint dynamics along sampled
trajectories~\citep{domingoenrich2025adjoint}.
\nameshort uses stochastic optimal control to motivate a local
endpoint-gradient target that avoids adjoint integration.
\Cref{tab:baseline_comparison} summarizes the computational
differences between these methods.

\paragraph{Reward Score Matching}
RSM unifies several reward fine-tuning objectives as score matching
against a value-guided target, highlighting guidance estimation and
time weighting as central design choices~\citep{lee2026reward}.
Its first-order endpoint-gradient variant is particularly close to
\nameshort: both use reward gradients evaluated at fully generated
samples to supervise local model updates.
RSM applies this guidance at states collected along the sampling
trajectory.
\nameshort instead independently re-noises each endpoint, allowing
the same reward gradient to supervise multiple training states without
requiring additional sampling steps.
\Cref{lower-variance} shows that, when both follow the conditional law $p_{t|1}$, re-noising based gradient estimators have no larger covariance than those constructed using shared SDE trajectories.
RSM's flow-matching implementation instead collects states from ODE trajectories, making $X_t$ a deterministic function of $X_1$. This gives a different conditional coupling, and so \Cref{lower-variance} does not apply directly.
\nameshort instead constructs fresh training states at independently sampled times without additional rollout steps. \S\ref{app:independent_renoising} compares the two methods with all other settings matched.

\citet{domingoenrich2026unified} study conditional regression-target
variance at a single training state. For \nameshort, the
corresponding unweighted term is
\[
\mathbb E\left\|
\nabla R(X_1)-\mathbb E[\nabla R(X_1)\mid X_t]
\right\|^2,
\]
which is finite for bounded $\nabla R$. With the sampling law
held fixed, this term contributes no expected gradient and is
the conditional-variance constant removed in
\Cref{lem:endpoint_target_variance}.
\Cref{lower-variance} instead concerns correlation between
multiple training states per endpoint, comparing independent
conditional draws with shared SDE trajectories.

Adjoint Sampling~\citep{havens2025adjoint} also re-noises generated
endpoints, amortizing sampling and energy-gradient evaluation
across multiple updates. \nameshort uses this construction for
fine-tuning pretrained flow models, approximating gradient
transport by directly reusing endpoint reward gradients.

\paragraph{Efficient Adjoint Matching}
EAM shares \nameshort's use of endpoint re-noising and its
stochastic-control perspective~\citep{shin2026efficient}.
It obtains a closed-form adjoint by introducing a linear base drift
and compensating through a modified terminal cost.
This correction requires an estimate of the pretrained terminal score,
which EAM approximates using the pretrained model.
\nameshort retains the reference dynamics in its control formulation
and instead approximates gradient transport by reusing the clean
endpoint reward gradient.
The resulting target is the reference velocity plus this gradient.

\begin{table*}[h]
\centering
\caption{Comparison of differentiable reward fine-tuning methods.}
\label{tab:baseline_comparison}
\small
\setlength{\tabcolsep}{4pt}
\renewcommand{\arraystretch}{1.15}
\begin{tabularx}{\textwidth}{
@{}l*{3}{>{\raggedright\arraybackslash}X}@{}
}
\toprule
Method & Reward input & Gradient construction & Training states \\
\midrule
\shortstack[l]{AlignProp\\\citep{prabhudesai2023aligning}}
& Endpoint & Sampling backprop. & Trajectory \\[4pt]

\shortstack[l]{DRaFT-$K$\\\citep{clark2024directly}}
& Endpoint & Last-$K$ backprop. & Trajectory \\[4pt]

\shortstack[l]{DRTune\\\citep{wu2024deep}}
& Endpoint & Scalar transport & Trajectory \\[4pt]

\shortstack[l]{ReFL\\\citep{xu2023imagereward}}
& One-step estimate & One-step backprop. & Trajectory (late steps) \\[4pt]

\shortstack[l]{EasyTune\\\citep{tan2026easytune}}
& Noise-aware or one-step estimate & One-step backprop. & Trajectory \\[4pt]

\shortstack[l]{SQDF\\\citep{kang2026diffusion}}
& Clean estimate & Soft-$Q$ backprop. & Trajectory replay \\[4pt]

\shortstack[l]{Direct-Align\\
\citep{shen2025directlyaligningdiffusiontrajectory}}
& Recovered image & Noise-corrected backprop.
& Multi-time re-noising \\[4pt]

\shortstack[l]{LeapAlign\\\citep{liang2026leapalign}}
& Endpoint & Two-step surrogate & Trajectory \\[4pt]

\shortstack[l]{FlowBP-Lagrange\\\citep{wang2026exploring}}
& Endpoint & Lagrange surrogate & Trajectory \\[4pt]

\shortstack[l]{VGG-Flow\\\citep{liu2025value}}
& One-step estimate & Learned correction & Trajectory \\[4pt]

\shortstack[l]{RSM\\\citep{lee2026reward}}
& Endpoint & Direct gradient & Trajectory \\[4pt]

\shortstack[l]{Adjoint Matching\\
\citep{domingoenrich2025adjoint}}
& Endpoint & Lean adjoint &  Trajectory \\[4pt]

\shortstack[l]{EAM\\\citep{shin2026efficient}}
& Endpoint & Closed-form adjoint & Multi-time re-noising \\
\midrule
\textbf{\nameshort}
& Endpoint & Direct gradient & Multi-time re-noising \\
\bottomrule
\end{tabularx}
\end{table*}

\section{Experimental details and additional results}
\label{app:results}

\subsection{Experimental protocol}
\label{app:hyperparameter_search}
We tune each method separately for PickScore and HPSv2.1,
first conducting a broad sweep and then evaluating additional
settings near the target reward levels used in
\Cref{tab:quantitative_pickscore_matched,tab:quantitative_hpsv2_matched}.
We vary the learning rate for DRaFT-LV and the reward coefficient
$\beta$ for the other methods, keeping all other training settings
fixed. \Cref{tab:hyperparameter_search} lists the examined hyperparameters;
bold entries identify the settings used for the three-seed comparisons reported in \Cref{tab:quantitative_pickscore_matched,tab:quantitative_hpsv2_matched}.

\begin{table}[h]
\centering
\small
\setlength{\tabcolsep}{6pt}
\caption{Hyperparameter values evaluated across the initial sweeps and
subsequent targeted runs. Bold entries identify the settings used for \Cref{tab:quantitative_pickscore_matched,tab:quantitative_hpsv2_matched}. Reward coefficients are expressed in units of $10^3$;
DRaFT-LV learning rates are expressed in units of $10^{-5}$.}
\label{tab:hyperparameter_search}
\begin{tabular}{llcc}
\toprule
Method & Parameter & PickScore & HPSv2.1 \\
\midrule
SERF
& Reward coefficient
& $0.75,\,1.1,\,\mathbf{1.5},\,2.2,\,7$
& $12,\,20,\,27.5,\,\mathbf{36},\,52.5$ \\

RSM
& Reward coefficient
& $2.75,\,4.3,\,\mathbf{7},\,10,\,30$
& $\,50,\,\mathbf{100},\,130,\,290,\,1000$ \\

DRaFT-LV
& Learning rate
& $2,\,\mathbf{5},\,8,\,10,\,20$
& $2,\,\mathbf{4},\,5,\,8,\,20$ \\

VGG-Flow
& Reward coefficient
& $2,\,5,\,10,\,15,\,\mathbf{25}$
& $50,\,75,\,100,\,\mathbf{200},\,500$ \\

Adjoint Matching
& Reward coefficient
& $2,\,5,\,10,\,15,\,\mathbf{25}$
& $50,\,100,\,200,\,500,\,\mathbf{1000}$ \\
\bottomrule
\end{tabular}
\end{table}

For \Cref{tab:quantitative_pickscore_matched,tab:quantitative_hpsv2_matched},
we evaluate each selected hyperparameter setting with three
independent training seeds. Within each run, we select checkpoints
separately for each evaluation set, choosing the evaluated checkpoint
closest to the target reward: PickScore $23.45$ on Pick-a-Pic and
$23.40$ on DrawBench, or HPSv2.1 $0.365$ on HPDv2 and $0.325$
on DrawBench. If all checkpoints remain below the target, we select
the highest-reward checkpoint. All metrics for each evaluation set
use its selected checkpoint. We report means $\pm$ sample standard
deviations across the three runs.

\newpage
\subsection{Additional quantitative results}
\label{app:quantitative_results}
\begin{figure}[h]
    \centering
    \includegraphics[width=0.95\linewidth]{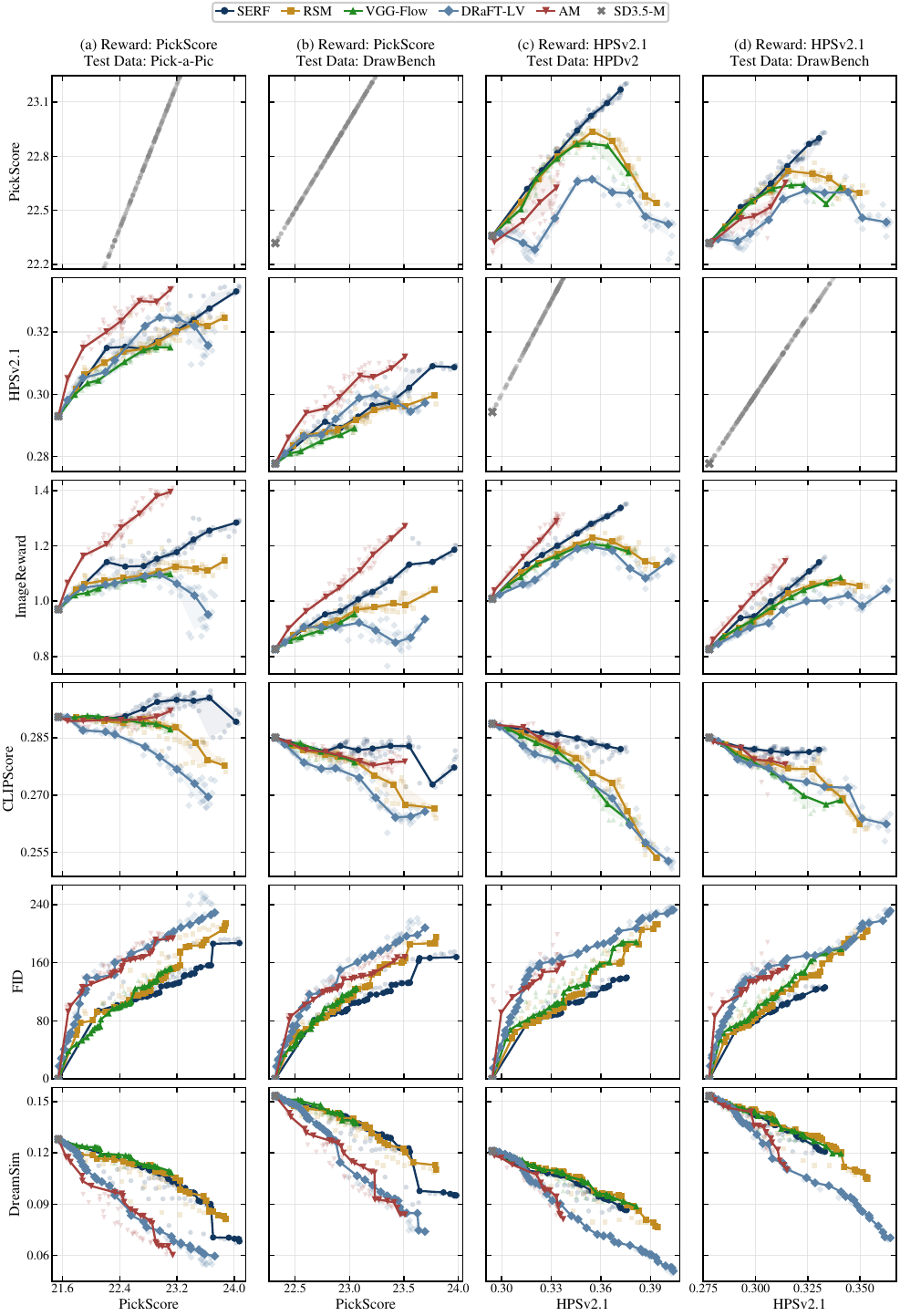}
    \caption{Reward--quality trade-offs.}
    \label{fig:pareto_additional}
\end{figure}

We report a complete trade-off panel in~\Cref{fig:pareto_additional}.

\newpage
\xhdr{Compositional benchmark} We evaluate the PickScore- and HPSv2.1-trained checkpoints selected for
\Cref{tab:quantitative_pickscore_matched,tab:quantitative_hpsv2_matched}
on \textit{GenEval-test-200}, a fixed subset of 200 unique GenEval test
prompts~\citep{ghosh2023geneval}. The checkpoints are fixed before this
evaluation. For each prompt, we generate 16 images using 20 Euler steps
and CFG $4.5$, with image seeds shared across methods and the frozen
base. Values are means and sample standard deviations across
three training seeds per method and training reward.

\Cref{tab:geneval_breakdown} reports overall and per-category GenEval
accuracy, with overall accuracy computed as the unweighted mean of
the six category accuracies. Among fine-tuned methods, \nameshort
achieves the highest mean overall accuracy after HPSv2.1 training
($67.37\%$) and the second-highest after PickScore training
($69.00\%$, versus $70.75\%$ for Adjoint Matching), both exceeding
the frozen base ($63.34\%$). It ranks first or second in every
category under both training rewards.

These gains on an unoptimized compositional benchmark provide
further evidence of reduced reward hacking: \nameshort's preference
improvements also translate into better prompt adherence in a compositional task.

\begin{table*}[h]
\centering
\caption{
GenEval-test-200 binary accuracy (\%) by category, grouped by
training reward. Overall is the unweighted mean of the six category
accuracies. Values are mean $\pm$ sample standard deviation over three
training seeds. Bold and underline denote the best and second-best
fine-tuned means within each group. The shared frozen base is shown once.
}
\label{tab:geneval_breakdown}
\begingroup
\setlength{\tabcolsep}{5pt}
\renewcommand{\arraystretch}{1.10}
\resizebox{\textwidth}{!}{%
\begin{tabular}{lccccccc}
\toprule
Method
& \shortstack{Single\\object}
& \shortstack{Two\\objects}
& Counting
& Colors
& Position
& \shortstack{Color attribute\\binding}
& Overall \\
\midrule

\textcolor{gray}{SD3.5-M + CFG}
& \textcolor{gray}{$99.35$}
& \textcolor{gray}{$78.75$}
& \textcolor{gray}{$49.33$}
& \textcolor{gray}{$77.96$}
& \textcolor{gray}{$22.71$}
& \textcolor{gray}{$51.94$}
& \textcolor{gray}{$63.34$} \\

\midrule
\multicolumn{8}{c}{\textit{Training reward: PickScore}} \\
\midrule

DRaFT-LV
& \underline{\pmt{99.28}{0.25}}
& \pmt{75.12}{0.57}
& \pmt{52.31}{4.36}
& \pmt{58.77}{3.47}
& \pmt{18.19}{1.08}
& \pmt{33.12}{6.19}
& \pmt{56.13}{2.04} \\

VGG-Flow
& \pmt{97.84}{0.86}
& \pmt{74.88}{1.44}
& \pmt{52.08}{4.13}
& \pmt{78.89}{1.12}
& \pmt{23.22}{0.99}
& \pmt{52.30}{0.87}
& \pmt{63.20}{0.80} \\

Adjoint Matching
& \pmt{98.71}{0.86}
& \bpmt{88.39}{1.56}
& \bpmt{60.42}{2.45}
& \bpmt{85.91}{1.85}
& \bpmt{27.18}{1.45}
& \bpmt{63.86}{2.48}
& \bpmt{70.75}{1.39} \\

RSM
& \underline{\pmt{99.28}{0.33}}
& \pmt{78.27}{1.05}
& \pmt{52.60}{1.23}
& \pmt{79.71}{0.68}
& \pmt{23.88}{0.78}
& \pmt{55.53}{1.96}
& \pmt{64.88}{0.76} \\

\textbf{SERF}
& \bpmt{99.86}{0.12}
& \underline{\pmt{85.36}{0.64}}
& \bpmt{57.07}{1.29}
& \underline{\pmt{82.57}{0.87}}
& \bpmt{27.39}{0.98}
& \bpmt{61.78}{1.40}
& \underline{\pmt{69.00}{0.20}} \\

\midrule
\multicolumn{8}{c}{\textit{Training reward: HPSv2.1}} \\
\midrule

DRaFT-LV
& \bpmt{99.71}{0.12}
& \underline{\pmt{81.37}{1.52}}
& \bpmt{54.39}{0.85}
& \pmt{60.14}{3.62}
& \underline{\pmt{23.27}{0.23}}
& \pmt{44.25}{4.57}
& \pmt{60.52}{1.64} \\

VGG-Flow
& \pmt{99.21}{0.12}
& \pmt{76.19}{2.06}
& \pmt{51.86}{0.56}
& \underline{\pmt{78.07}{0.94}}
& \pmt{22.97}{0.62}
& \pmt{48.06}{0.57}
& \pmt{62.73}{0.53} \\

Adjoint Matching
& \pmt{97.05}{1.86}
& \pmt{78.87}{5.81}
& \pmt{48.36}{4.27}
& \pmt{71.93}{3.42}
& \pmt{23.17}{3.32}
& \pmt{45.11}{9.91}
& \pmt{60.75}{4.50} \\

RSM
& \pmt{99.57}{0.22}
& \pmt{75.83}{1.36}
& \pmt{51.04}{0.68}
& \pmt{77.63}{0.59}
& \pmt{23.17}{0.26}
& \underline{\pmt{51.51}{1.12}}
& \underline{\pmt{63.13}{0.19}} \\

\textbf{SERF}
& \bpmt{99.78}{0.00}
& \bpmt{84.70}{1.32}
& \bpmt{55.95}{1.01}
& \bpmt{80.43}{1.03}
& \bpmt{26.27}{1.10}
& \bpmt{57.11}{0.99}
& \bpmt{67.37}{0.43} \\

\bottomrule
\end{tabular}%
}
\endgroup
\end{table*}

\clearpage

\subsection{Additional qualitative results}
\label{app:qualitative}
\begin{figure}[h]
    \centering
    \includegraphics[width=1.0\linewidth]{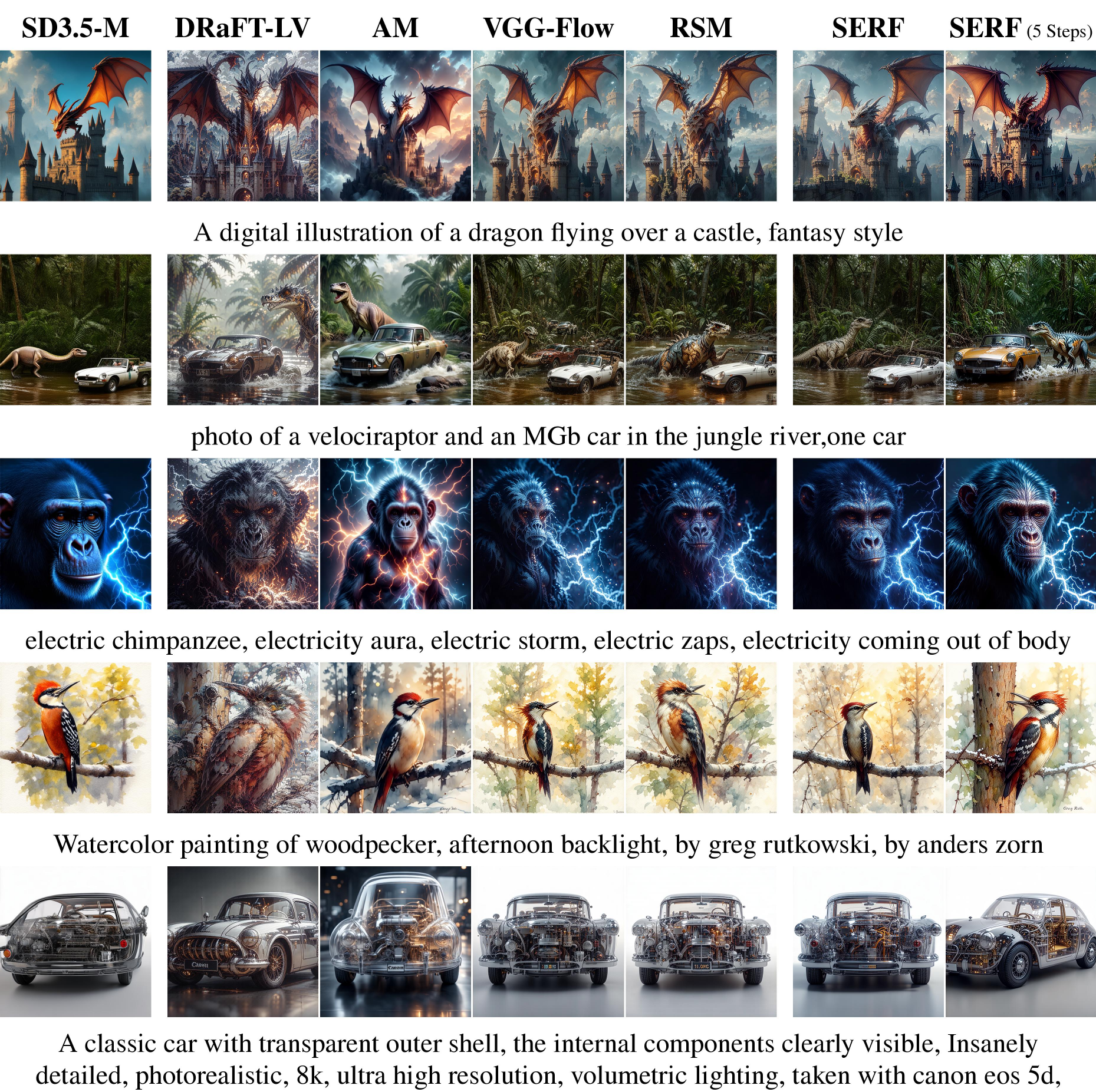}
    \caption{Qualitative comparison of PickScore-trained models on held-out Pick-a-Pic prompts. All methods use the same initial noise, CFG $4.5$, and $20$ Euler steps at inference. SERF uses $10$ training sampling steps unless otherwise indicated.}
    \label{fig:qualitative_1}
\end{figure}

\begin{figure}[h]
    \centering
    \includegraphics[width=1.0\linewidth]{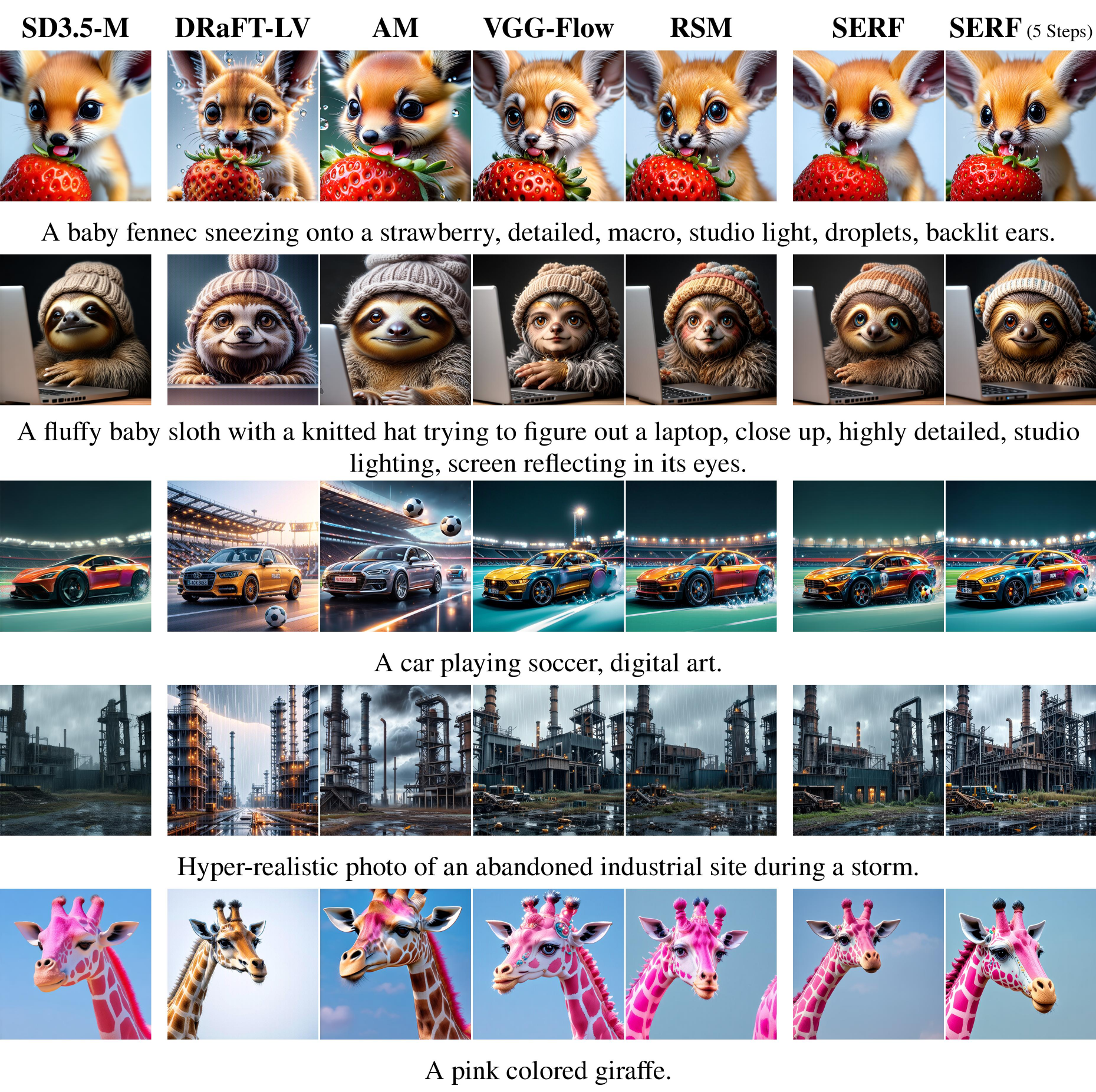}
\caption{Qualitative comparison of HPSv2.1-trained models on DrawBench prompts. All methods use the same initial noise, CFG $4.5$, and $20$ Euler steps at inference. SERF uses $10$ training sampling steps unless otherwise indicated.}
    \label{fig:qualitative_2}
\end{figure}

\begin{figure}[h]
    \centering
    \includegraphics[width=1.0\linewidth]{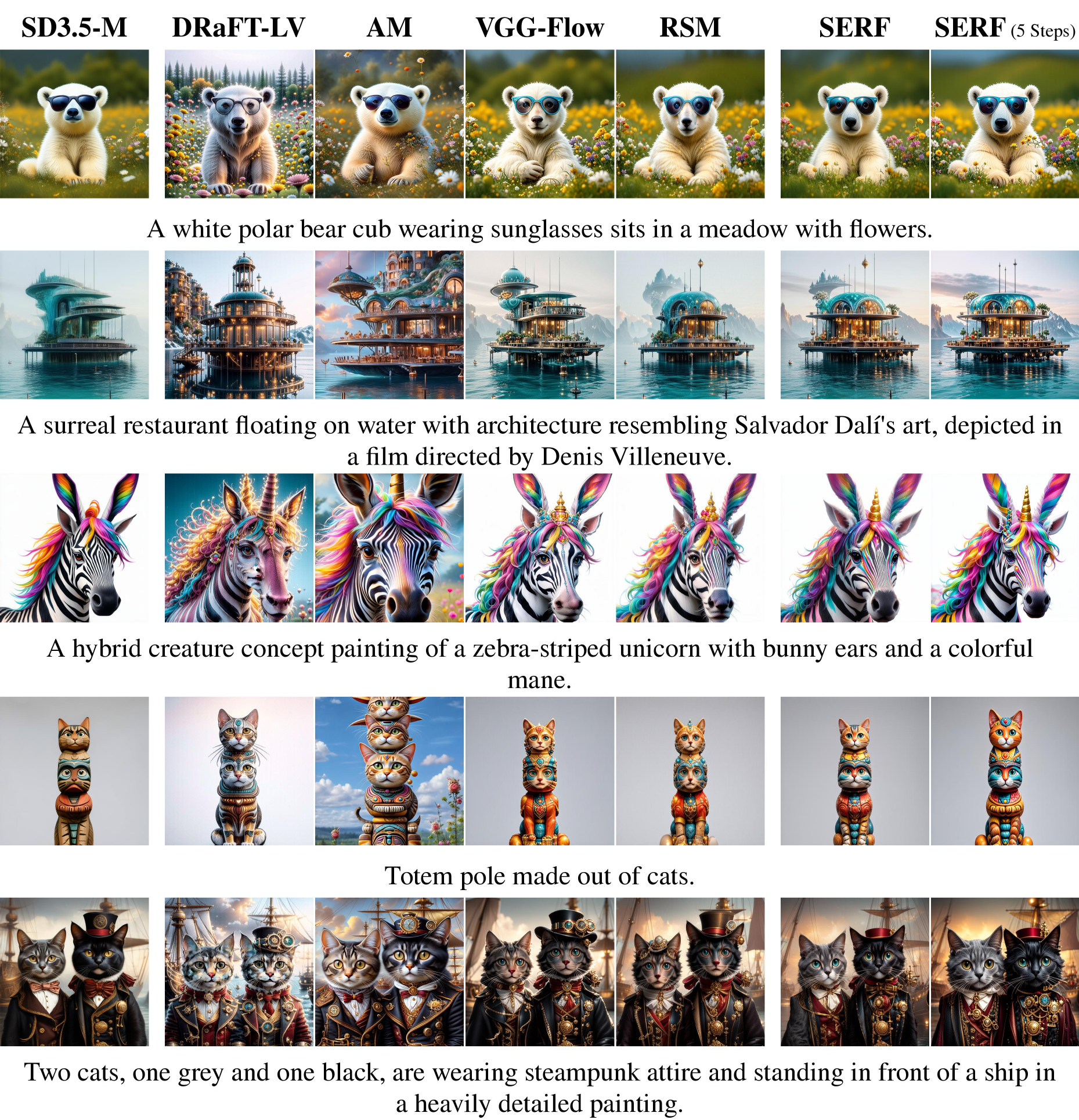}
\caption{Qualitative comparison of HPSv2.1-trained models on held-out HPDv2 prompts. All methods use the same initial noise, CFG $4.5$, and $20$ Euler steps at inference. SERF uses $10$ training sampling steps unless otherwise indicated.}
    \label{fig:qualitative_3}
\end{figure}

\clearpage
\paragraph{JPEG Compressibility}
Following~\citet{clark2024directly}, we optimize
\[
R_{\mathrm{JPEG}}(x)
=
-\|x-\operatorname{JPEG}_{75}(x)\|_2^2,
\]
where $\operatorname{JPEG}_{75}(x)$ is the decoded JPEG
reconstruction at quality 75 and the norm sums squared pixel
errors. This favors images whose appearance changes little
under JPEG compression.

\begin{figure}[h]
    \centering
    \includegraphics[width=\linewidth]{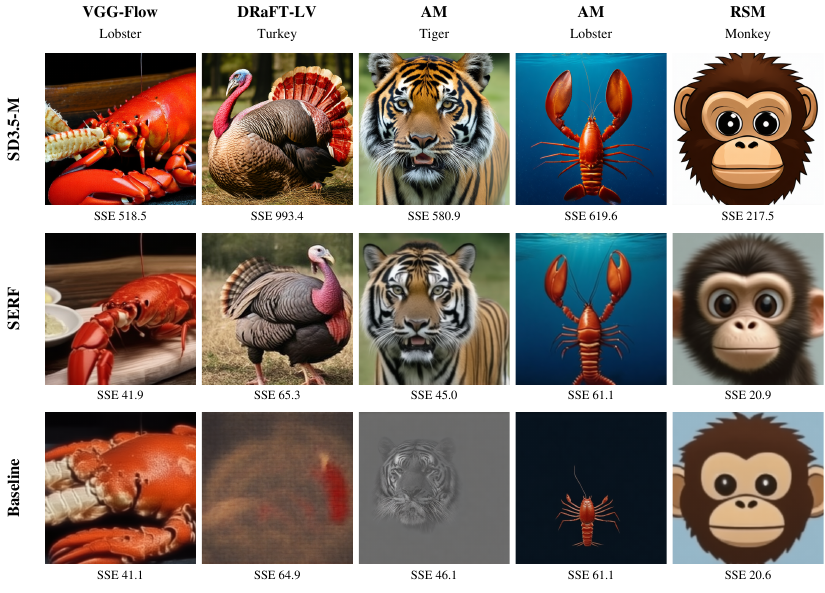}
    \caption{Additional selected comparisons. Each column uses the same
    prompt and noise seed. Trained pairs are matched within $3\%$
    reconstruction SSE; base SSE is unmatched.}
    \label{fig:jpeg_qualitative_appendix}
\end{figure}

At matched JPEG reconstruction error, \nameshort preserves subject
structure and detail where competing methods exhibit severe
blurring, texture collapse, or subject shrinkage
(\Cref{fig:jpeg_qualitative_appendix}).
This extends its observed reward--quality advantage beyond
human-preference objectives.

\begin{figure}[h]
    \centering
    \includegraphics[width=\linewidth]{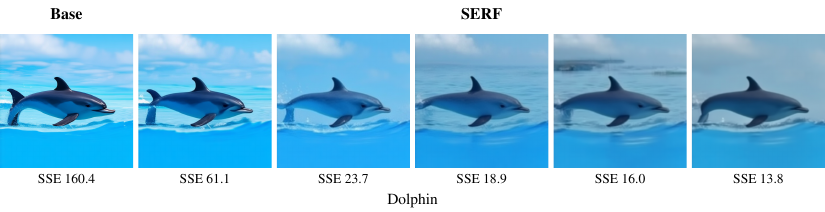}
    \caption{Base model followed by SERF checkpoints at updates
    $10$, $20$, $30$, $40$, and $50$ ($c=3000$).
    The dolphin prompt and noise seed $14$ remain fixed.
    Reconstruction SSE decreases as backgrounds and fine details simplify.}
    \label{fig:serf_jpeg_image_progression}
\end{figure}

\Cref{fig:jpeg_qualitative_appendix} provides additional selected comparisons.
The fixed-seed dolphin progression shows increasingly smooth backgrounds
and reduced fine detail (\Cref{fig:serf_jpeg_image_progression}).
Table~\ref{tab:jpeg-heldout-matched-reward} compares methods at matched reward: SERF leads on HPSv2.1, PickScore, ImageReward and FID, despite a slight disadvantage from stronger JPEG optimization than RSM, VGG-Flow and DRaFT-LV.

\begin{table}[t]
\centering
\resizebox{\linewidth}{!}{%
\begin{tabular}{lrrrrrrr}
\toprule
Method & JPEG SSE $\downarrow$ & HPSv2.1 $\uparrow$
& PickScore $\uparrow$ & FID$_{\rm base}$ $\downarrow$
& ImageReward $\uparrow$ & Diversity $\uparrow$ & CLIPScore $\uparrow$ \\
\midrule
SERF & \underline{40.21} & \textbf{0.2912} & \textbf{21.835}
& \textbf{65.32} & \textbf{1.015} & 0.1011 & 0.2591 \\
RSM & 41.36 & \underline{0.2860} & \underline{21.808}
& 65.84 & \underline{0.981} & 0.1047 & \textbf{0.2636} \\
VGG-Flow & 42.97 & 0.2791 & 21.653
& \underline{65.68} & 0.932 & 0.1162 & \underline{0.2617} \\
DRaFT-LV & 41.34 & 0.2566 & 21.292
& 96.23 & 0.660 & \underline{0.1503} & 0.2557 \\
AM & \textbf{36.40} & 0.2289 & 20.976
& 143.61 & -0.074 & \textbf{0.1936} & 0.2391 \\
\bottomrule
\end{tabular}%
}
\caption{Held-out JPEG results at approximately matched reward.
Best in bold; second best underlined.}
\label{tab:jpeg-heldout-matched-reward}
\end{table}

\section{Ablations}
\label{app:ablations}

\subsection{Independent re-noising}
\label{app:independent_renoising}

\begin{figure}[h]
    \centering
    \includegraphics[width=\linewidth]{
        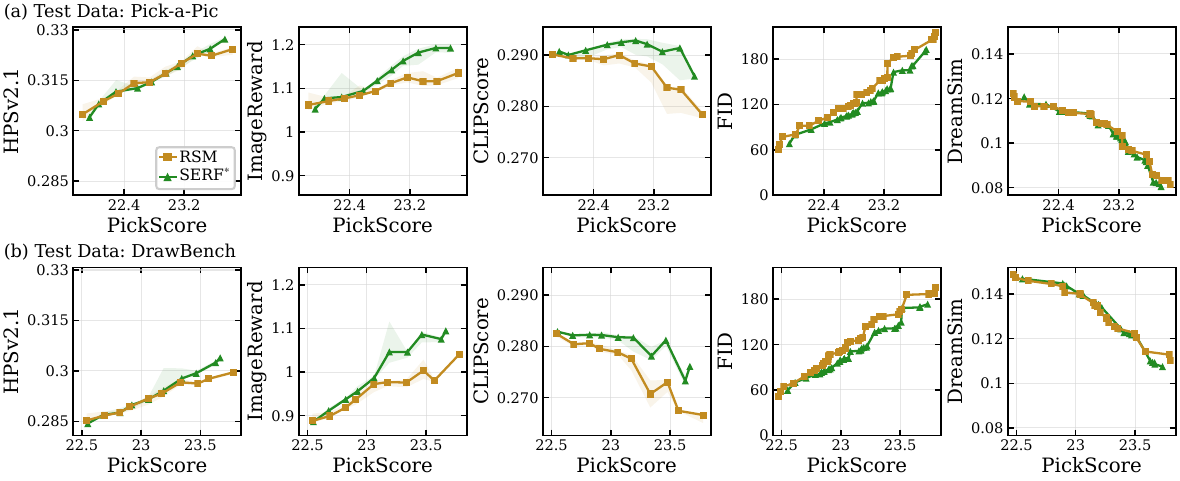
    }
    \caption{
        Independent re-noising ablation under PickScore optimization:
        RSM versus \nameshort$\!^\ast$ with matched training settings.
        Top: Pick-a-Pic. Bottom: DrawBench.
    }
    \label{fig:renoising_pickscore}
\end{figure}

\begin{figure}[h]
    \centering
    \includegraphics[width=\linewidth]{
        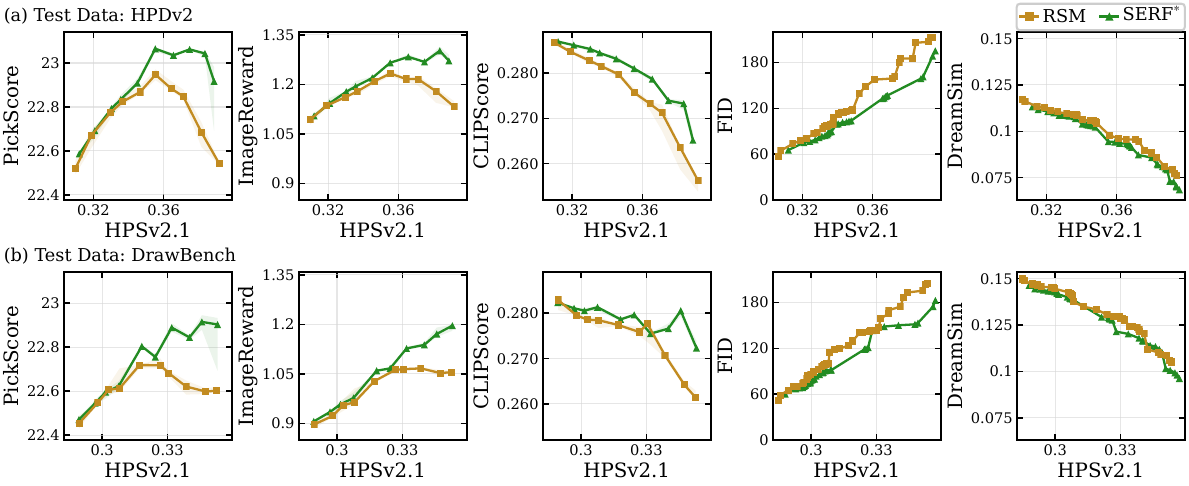
    }
    \caption{
        Independent re-noising ablation under HPSv2.1 optimization,
        following the same protocol as
        Figure~\ref{fig:renoising_pickscore}.
    }
    \label{fig:renoising_hps}
\end{figure}

\paragraph{Controlled comparison with RSM}
We isolate the effect of independent re-noising by comparing RSM with \nameshort$\!^\ast$. Both use uniform timestep sampling and $t$ weighting of the endpoint reward gradient, changing only trajectory states to independently re-noised states while keeping all other settings fixed. \nameshort$\!^\ast$ differs from \nameshort only in using uniform time sampling instead of the default $p(t) = 2t$, and by weighting the reward-gradient target by $t$. We plot the trade-offs using the same reward-coefficient sweep specified in \S\ref{app:hyperparameter_search}.
\Cref{fig:renoising_pickscore,fig:renoising_hps} establish a substantial
reward--quality advantage over RSM from independent re-noising
alone, with time sampling and reward weighting held fixed.
We hypothesize that re-noising supplies better training states by
using the same Gaussian corruption construction as flow-matching
pretraining, keeping supervision closer to the regime in which the
pretrained velocity field was learned.

\paragraph{Stochastic training trajectories.}
We investigate whether stochastic training trajectories can improve
\nameshort by exploring a broader range of intermediate states.
The SERF-SDE variant uses AM's memoryless SDE with $\epsilon=0.1$,
selects intermediate training states along the stochastic trajectory,
and uses AM's endpoint projection for reward evaluation.
This comparison therefore evaluates a complete stochastic
trajectory-based variant of \nameshort.

We sweep the reward coefficient for each variant and evaluate
checkpoints every 20 updates through update 300.
The SERF-SDE coefficient grids are rescaled relative to the default
SERF grids by factors of $6$ for PickScore and $5.5$ for HPSv2.1.
Both configurations use ten training rollout steps.
Evaluation uses the same 20-step Euler ODE sampler with CFG $4.5$.

\begin{figure}[h]
  \centering
  \includegraphics[width=\linewidth]
    {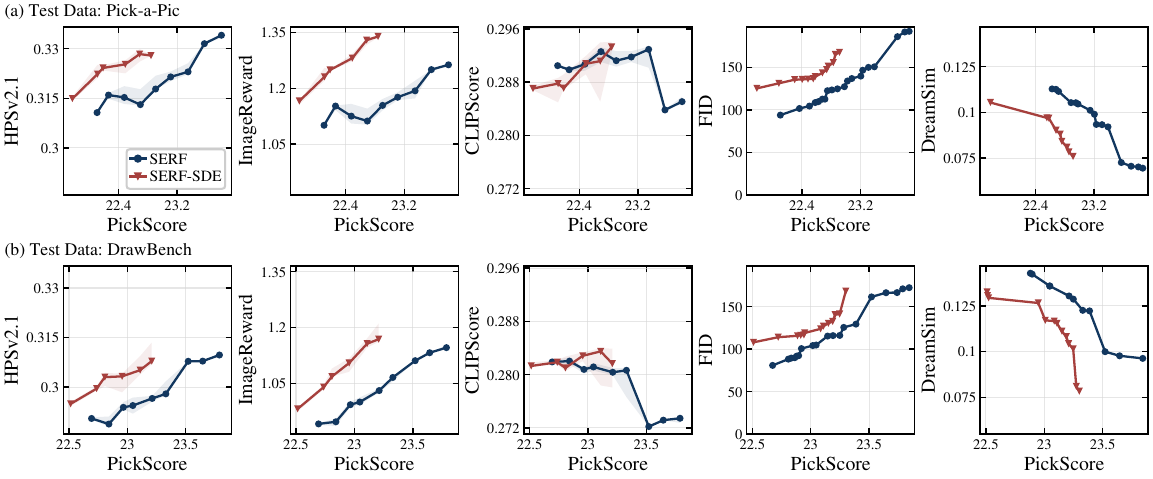}
  \caption{Reward--quality trade-offs for SERF and SERF-SDE under
  PickScore optimization. Top: Pick-a-Pic. Bottom: DrawBench.
  Comparisons include five reward coefficients per variant and
  checkpoints through update 300. Scalar metrics use reward-bin
  summaries; FID and diversity show Pareto frontiers.
  SERF-SDE denotes stochastic training; both variants use
  20-step Euler ODE evaluation with CFG $4.5$.}
  \label{fig:serf_sde_pickscore}
\end{figure}

\begin{figure}[h]
  \centering
  \includegraphics[width=\linewidth]
    {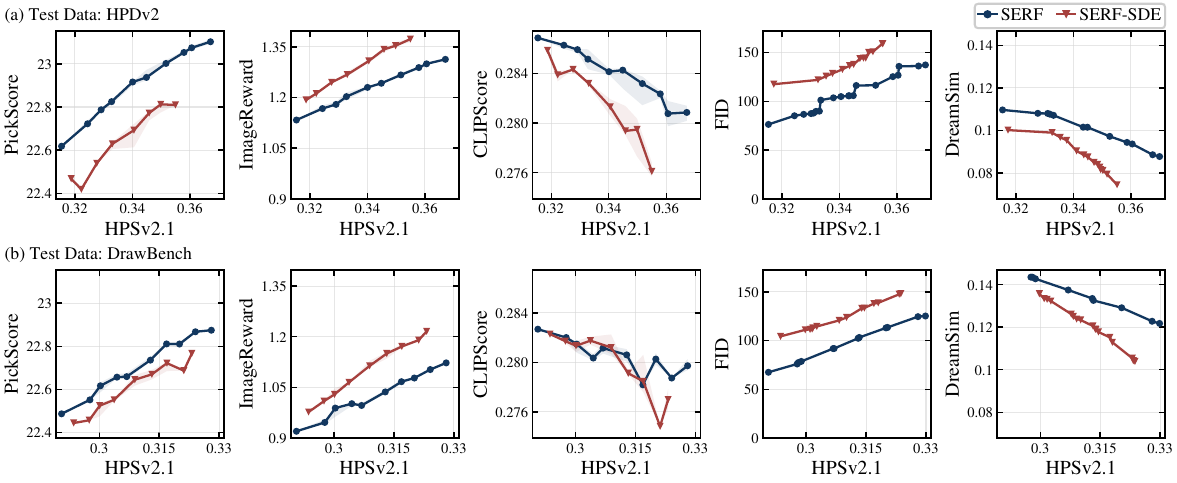}
  \caption{Reward--quality trade-offs for SERF and SERF-SDE under
  HPSv2.1 optimization. Top: HPDv2. Bottom: DrawBench.
  Comparisons include five reward coefficients per variant and
  checkpoints through update 300. Scalar metrics use reward-bin
  summaries; FID and diversity show Pareto frontiers.
  SERF-SDE denotes stochastic training; both variants use
  20-step Euler ODE evaluation with CFG $4.5$.}
  \label{fig:serf_sde_hps}
\end{figure}

Figures~\ref{fig:serf_sde_pickscore} and~\ref{fig:serf_sde_hps}
report the resulting reward--quality trade-offs.
The stochastic variant yields worse reward–quality trade-offs overall.
In particular, at comparable HPSv2.1 rewards, SERF-SDE exhibits
higher FID and lower diversity, although it achieves higher
ImageReward. \nameshort achieves substantially better reward--FID--diversity
trade-offs than SERF-SDE, empirically validating the practical
advantage of independent re-noising. \Cref{lower-variance} provides
a conditional variance guarantee; this experiment does not isolate
variance reduction as the cause of the gap.
We hypothesize that the large gap additionally reflects better
training states: Gaussian re-noising uses the pretraining corruption
construction and may better match the regime in which the pretrained
velocity field was learned.

\subsection{CFG and rollout length}
\begin{figure}[h]
    \centering
    \includegraphics[width=1.0\linewidth]{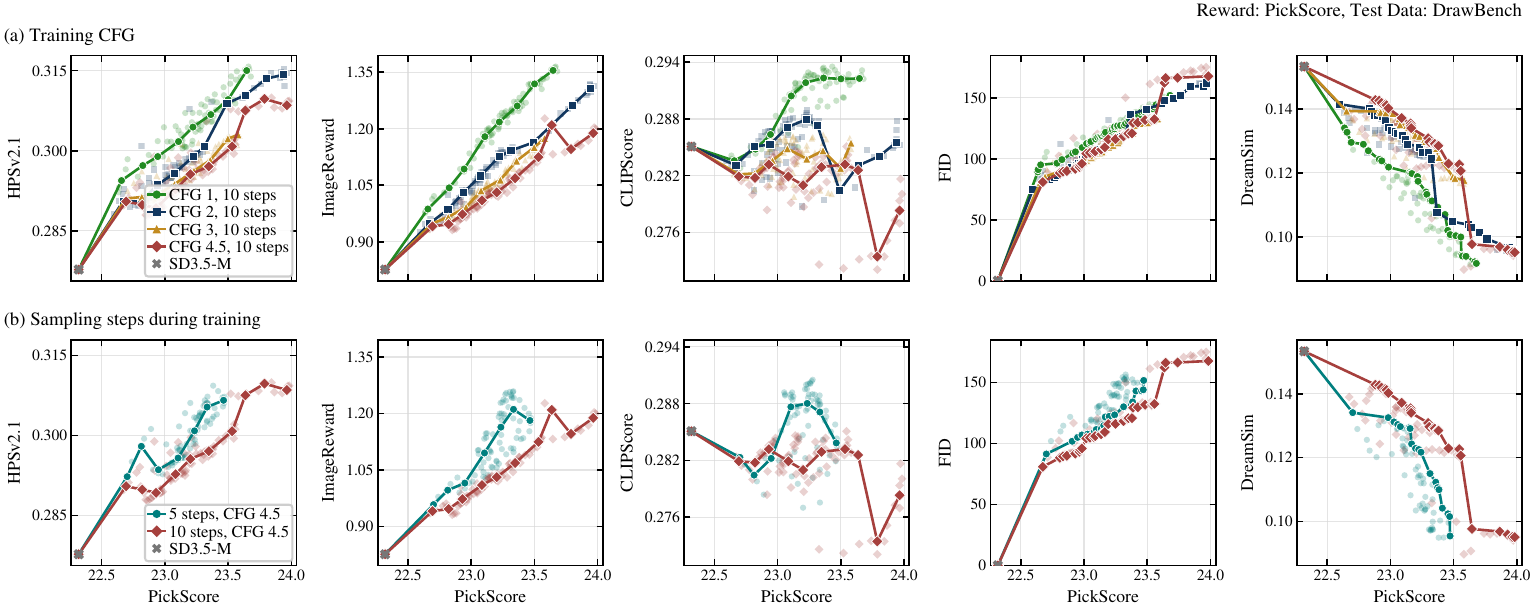}
    \caption{\nameshort trained on PickScore and evaluated on DrawBench. Top: Comparison for different training CFG scales. Bottom: using five or ten sampling steps per training rollout.}
    \label{fig:cfg_steps_b}
\end{figure}

\begin{figure}[h]
    \centering
    \vspace{-15pt}
    \includegraphics[width=1.0\linewidth]{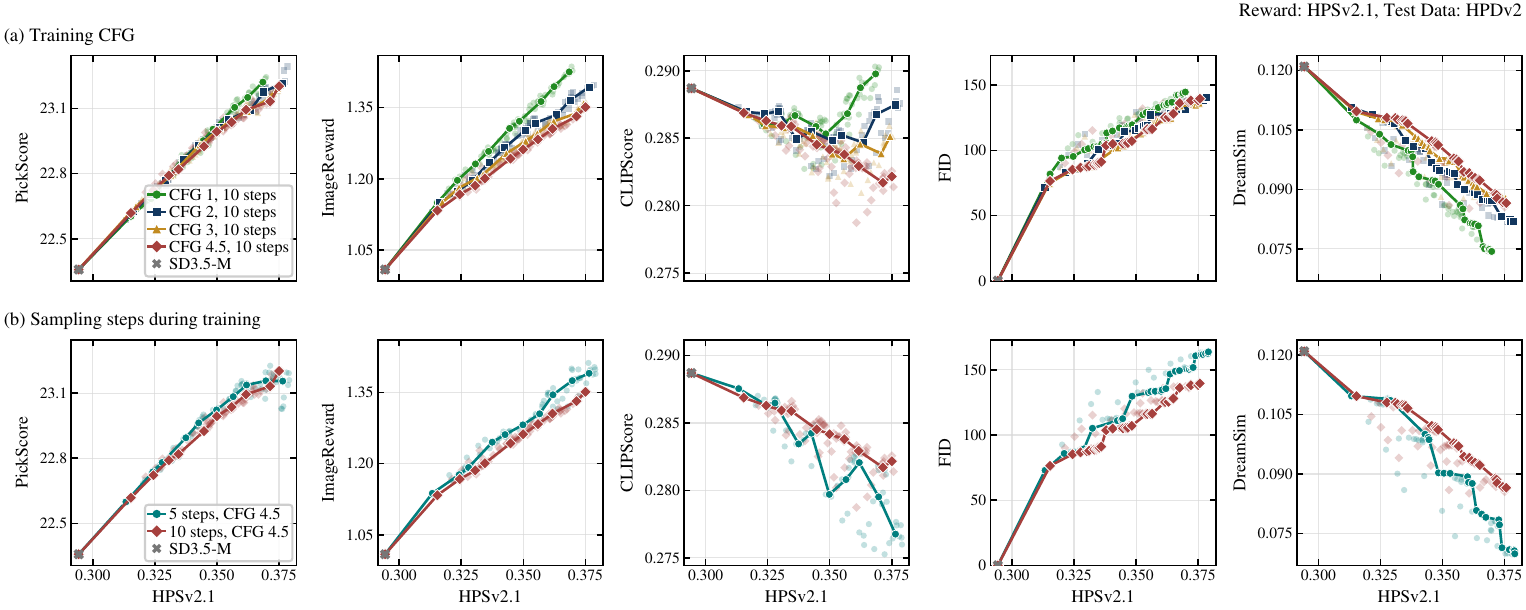}
    \caption{\nameshort trained on HPSv2.1 and evaluated on HPDv2. Top: Comparison for different training CFG scales. Bottom: using five or ten sampling steps per training rollout.}
    \label{fig:cfg_steps_c}
\end{figure}

\begin{figure}[h]
    \centering
    \vspace{-15pt}
    \includegraphics[width=1.0\linewidth]{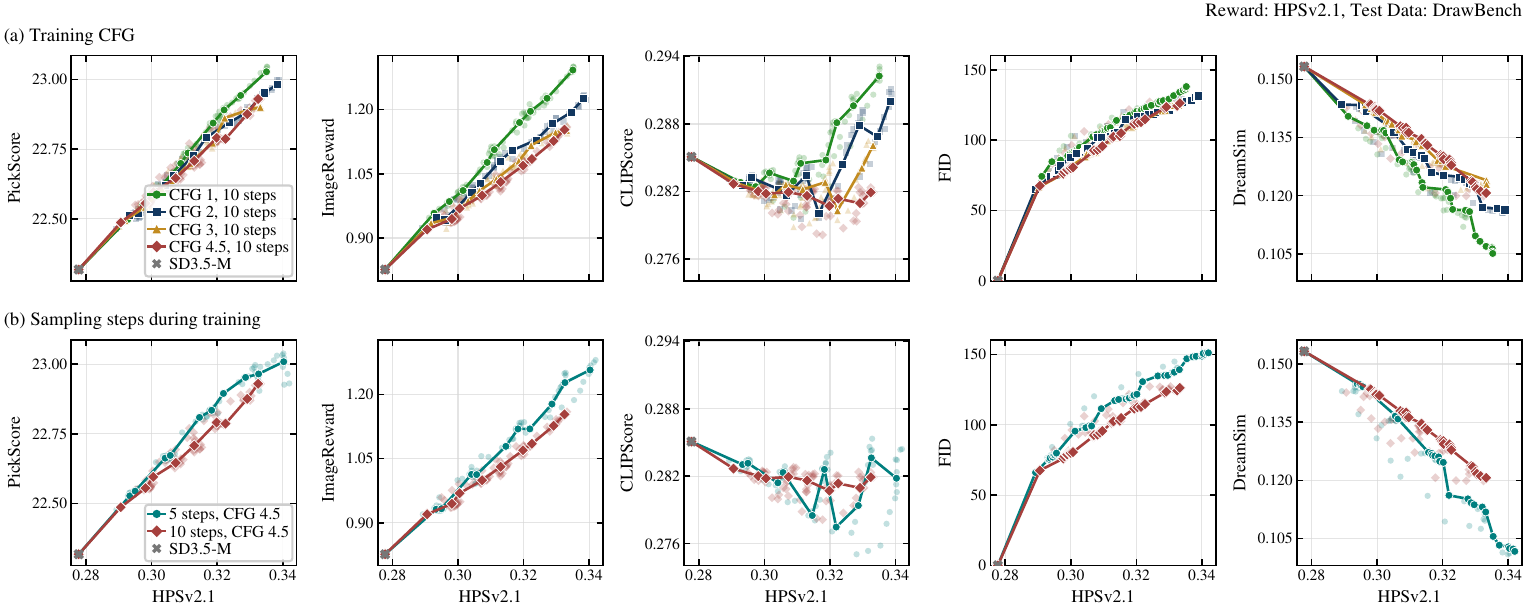}
    \caption{\nameshort trained on HPSv2.1 and evaluated on DrawBench. Top: Comparison for different training CFG scales. Bottom: using five or ten sampling steps per training rollout.}
    \label{fig:cfg_steps_d}
\end{figure}

We report in~\Cref{fig:cfg_steps_b},~\Cref{fig:cfg_steps_c}, and~\Cref{fig:cfg_steps_d} training settings ablations for different training rewards and evaluation datasets. 

The consistent reward improvements across the tested training CFG
scales demonstrate that \nameshort remains stable and effective
across these hyperparameter choices.

Five-step rollouts retain substantial reward improvements while
reducing update time by $26\%$ (\Cref{tab:update_time}).
Independent re-noising supplies fresh training states beyond those
visited during generation, decoupling the number of training targets
from rollout length, as further examined
in~\Cref{fig:serf_target_count}.

\subsection{Endpoint reuse}
\begin{figure}[h]
    \centering
    \includegraphics[width=\linewidth]{
        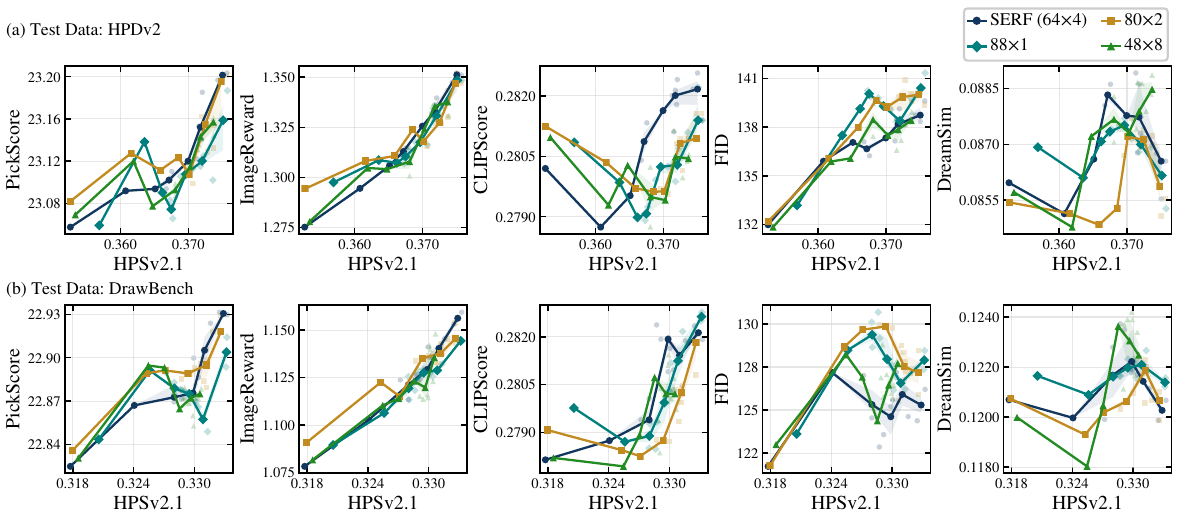
    }
    \caption{
        Endpoint/re-noising allocation under HPSv2.1 optimization.
        Legend entries denote endpoints per update times re-noised
        targets per endpoint. Top: HPDv2 test. Bottom: DrawBench.
        Lines and shading show binned medians and interquartile
        ranges for all five metrics.
        Displayed checkpoints span updates $60$--$300$ for
        $88\times1$ and $40$--$300$ for the other configurations.
    }
    \label{fig:serf_target_count}
\end{figure}

We compare endpoint/re-noising allocations
$(B,M)\in\{(88,1),(80,2),(64,4),(48,8)\}$,
adjusting $B$ to approximately match training cost.
All configurations optimize HPSv2.1 with $\beta=52{,}500$,
ten Euler rollout steps, and $300$ optimizer updates. The $64\times4$ default generally achieves stronger held-out
preference scores at high optimized rewards than allocations with
fewer re-noisings, while performing comparably to $48\times8$
(\Cref{fig:serf_target_count}).
This flexibility allows the allocation to be chosen according to
the relative costs of endpoint sampling, reward evaluation, and
model updates.

\begin{table}[h]
    \centering
    \caption{Training time per gradient update.
    Values are means $\pm$ sample standard deviations over
    20 updates after two warm-up updates, excluding validation
    and checkpoint saving. Timings from the eight-GPU runs
    are multiplied by two.}
    \label{tab:renoising_update_time}
    \small
    \setlength{\tabcolsep}{8pt}
    \begin{tabular}{ccc}
        \toprule
        Images/update & Re-noisings/image
        & Seconds/update $\downarrow$ \\
        \midrule
        88 & 1 & $9.547 \pm 0.231$ \\
        80 & 2 & $10.097 \pm 0.122$ \\
        64 & 4 & $8.865 \pm 0.012$ \\
        48 & 8 & $8.993 \pm 0.095$ \\
        \bottomrule
    \end{tabular}
\end{table}

\subsection{Timestep sampling and weighting}
\label{app:time_weighting}
\label{app:time_sampling}

\paragraph{Timestep sampling.}
By default, \nameshort samples $p(t)=2t$, emphasizing low-noise
states near $t=1$. Prior work identifies timestep-dependent
bias in local predicted-image guidance~\citep{lee2026reward};
\nameshort instead evaluates reward gradients at generated
endpoints. We compare the default sampling choice with uniform
sampling below.

\begin{figure}[h]
    \centering
    \includegraphics[width=0.95\linewidth]{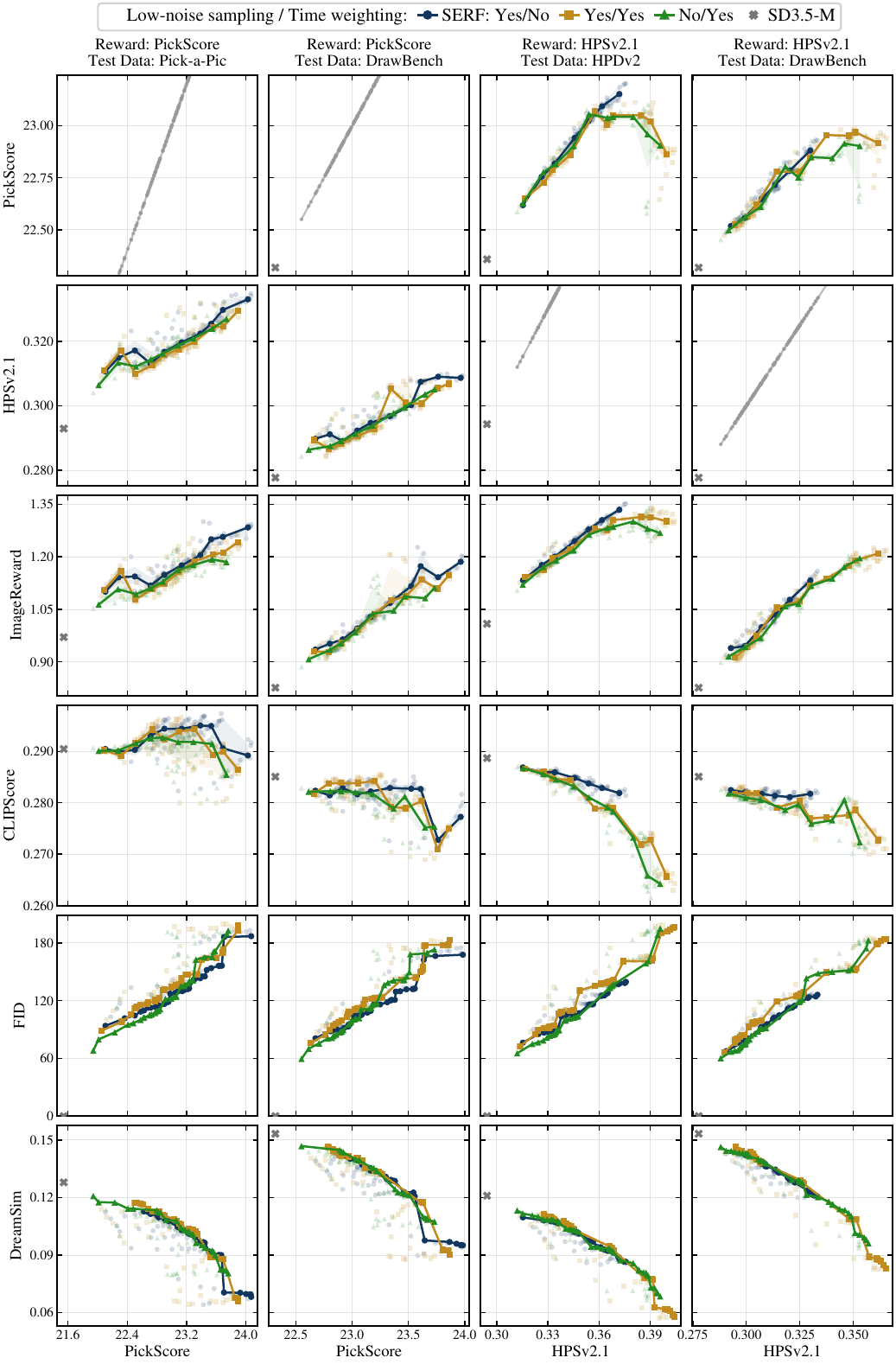}
\caption{Ablation of timestep sampling and target weighting in \nameshort. All variants use the common training horizon of $300$ updates.}
    \label{fig:time_sampling}
\end{figure}

\Cref{fig:time_sampling} examines timestep sampling and
reward-gradient target weighting. Slanted sampling with
$p(t)=2t$ favors low-noise states near clean data, whereas
uniform sampling treats all timesteps equally. Target weighting
multiplies the reward-gradient target by $t$, suppressing it
near pure noise ($t=0$) and preserving it near clean data ($t=1$).
We compare three configurations: slanted sampling without target
weighting (our default), slanted sampling with target weighting,
and uniform sampling with target weighting. In the legend,
``No'' for low-noise sampling denotes uniform sampling. Default \nameshort achieves better overall reward--quality
trade-offs than both target-weighted alternatives
(\Cref{fig:time_sampling}), supporting our simpler constant
reward-gradient target and echoing the benefits of simplified
objectives observed by~\citet{ho2020denoising}.
Our choice $p(t)=2t$ emphasizes low-noise states near clean data,
where fine perceptual refinements are particularly relevant to
preference rewards, consistent with the timestep-dependent guidance
analysis of~\citet{lee2026reward}.

\clearpage

\clearpage
\paragraph{Reference and reward-gradient weighting.}
We vary the reference-penalty weighting $k(t)$ and the scalar
factor $c(t)$ multiplying the endpoint reward gradient.
The identity, one-step, and Gaussian factors are
\[
    c_{\mathrm{id}}(t)=1,\qquad
    c_{\mathrm{step}}(t)=t,\qquad
    c_{\mathrm{Gauss}}(t)=\frac{t}{t^2+(1-t)^2},
\]
respectively; their derivations are given in
\S\ref{app:stochastic_jacobian}.

\begin{figure}[h]
    \centering
    \includegraphics[width=\linewidth]{
        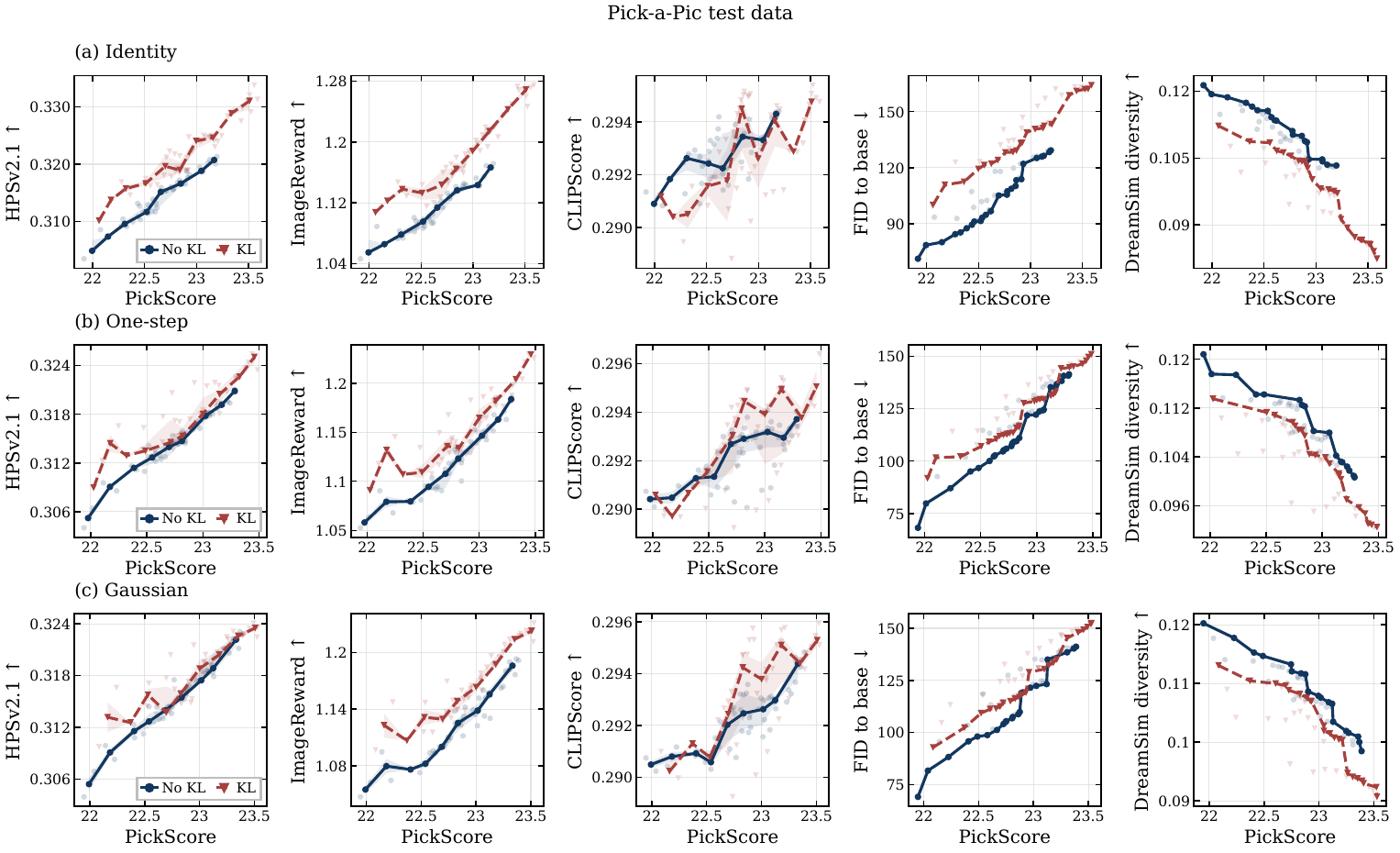
    }
    \caption{
        Effect of reference weighting under PickScore optimization
        on Pick-a-Pic test data.
        Rows use identity, one-step, and Gaussian guidance factors.
        The horizontal axis is achieved test PickScore.
        The first three columns show binned medians and
        interquartile bands; FID and DreamSim show Pareto frontiers.
        ``One-step'' names the guidance factor; all variants
        use ten-step training rollouts.
    }
    \label{fig:serf_kl_vs_constant}
\end{figure}

\Cref{fig:serf_kl_vs_constant} compares constant reference
weighting, $k(t)=1$, with the smoothed KL-derived weighting
$k(t)=(t+0.1)/(1-t+0.1)$, for each guidance factor.
The legend label ``No KL'' denotes constant reference weighting;
the reference penalty remains active.
KL-derived weighting generally gives worse FID and diversity at
comparable PickScore, favoring our simple constant reference weight.
This agrees with earlier observations that removing
variationally derived time weights can improve sample
quality~\citep{ho2020denoising}.

All six variants use the same setting as in \ref{sec:exp_settings}.

\begin{figure}[t]
    \centering
    \includegraphics[width=\linewidth]{
        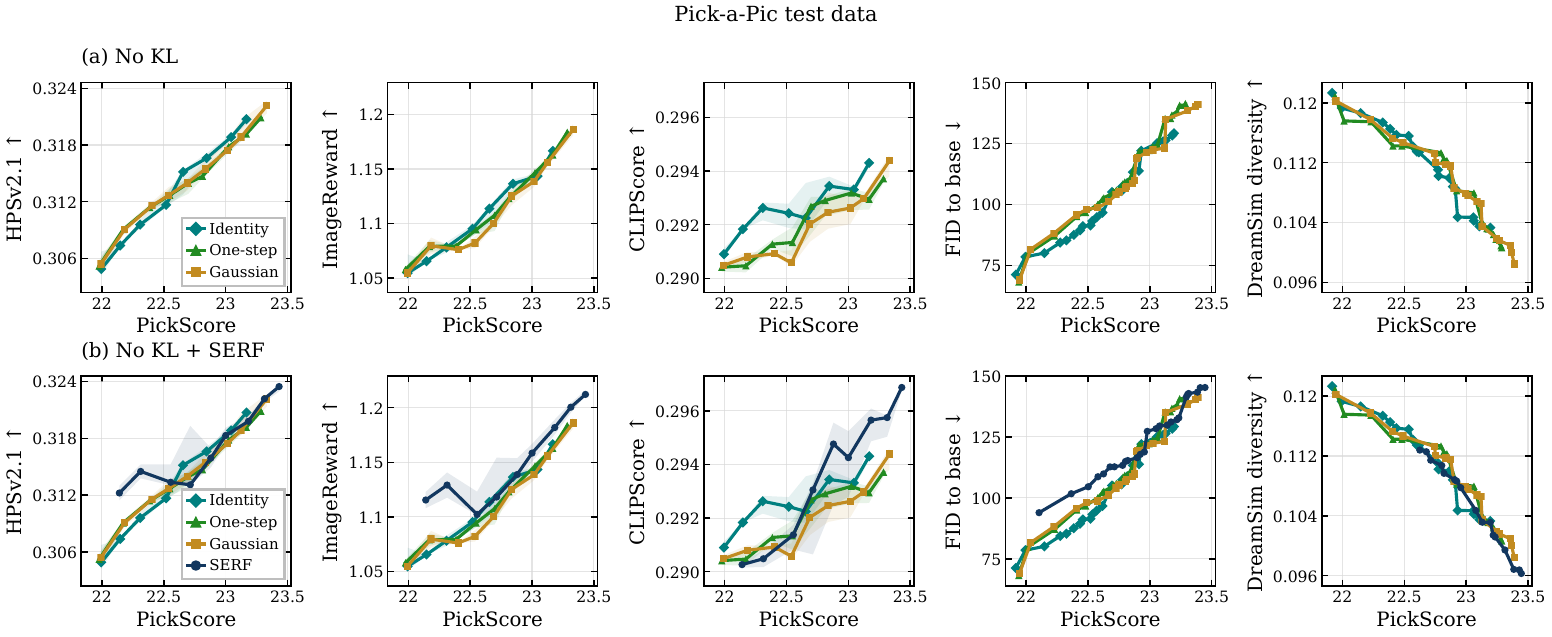
    }
    \caption{
        Guidance-weighting comparison with constant reference
        weighting on Pick-a-Pic.
        Top: identity, one-step, and Gaussian factors.
        Bottom: the same curves with default \nameshort added.
        Default \nameshort uses $p(t)=2t$; the other variants
        sample the interior scheduler grid uniformly.
        Plot summaries follow \Cref{fig:serf_kl_vs_constant}.
    }
    \label{fig:serf_constant_weightings}
\end{figure}

\Cref{fig:serf_constant_weightings} compares the three guidance
factors with constant reference weighting and adds default
\nameshort as a reference. The similar trade-offs demonstrate that the identity approximation
is competitive with the more elaborate scalar corrections,
supporting our simple constant guidance coefficient.
This further echoes the practical benefits of simplified
weighting observed by~\citet{ho2020denoising}.
Default \nameshort uses $p(t)=2t$, whereas the other variants
sample the interior scheduler grid uniformly.

\section{Training dynamics and efficiency}
\label{app:e}

\subsection{Training efficiency}
\label{app:training_efficiency}

\paragraph{Cost per update.}
We distinguish the cost of each gradient update from the training
time needed to reach comparable rewards.
\Cref{tab:update_time} reports update times on four NVIDIA GH200 GPUs,
using 64 trajectories and four training targets per trajectory.
We measure 20 updates after two warm-up updates, excluding validation
and checkpoint saving. Standard deviations describe variation across
timed updates.

\begin{figure}[t]
    \centering
    \vspace{-25pt}
    \includegraphics[width=\linewidth]{
        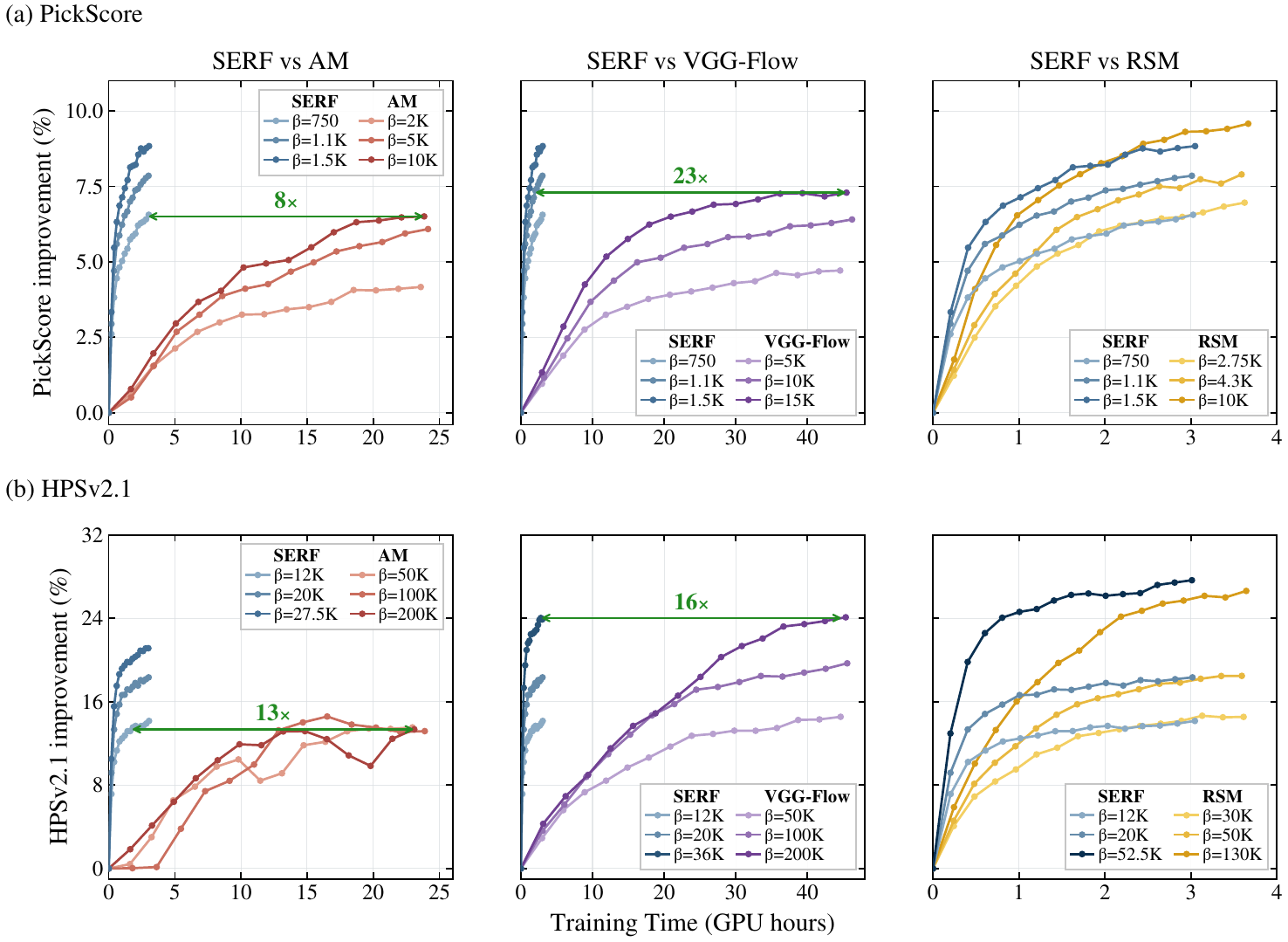}
    \caption{Training efficiency for (a) PickScore on Pick-a-Pic and
    (b) HPSv2.1 on HPDv2. Curves show selected reward coefficients.
    The horizontal axes use linear training GPU-hours, excluding
    validation, with different ranges across columns.
    Green arrows compare training times at matched rewards; ratios are rounded to integers.}
    \label{fig:training_efficiency}
\end{figure}

\begin{table}[t]
    \centering
    \caption{Training time per gradient update on four GH200 GPUs.
    Values are means $\pm$ sample standard deviations over 20 updates.}
    \label{tab:update_time}
    \small
    \setlength{\tabcolsep}{8pt}
    \begin{tabular}{lcc}
        \toprule
        Method & Training rollout steps & Seconds/update $\downarrow$ \\
        \midrule
        DRaFT-LV         & 10 & $32.634 \pm 0.263$ \\
        VGG-Flow         & 10 & $127.097 \pm 0.239$ \\
        Adjoint Matching & 10 & $73.318 \pm 0.202$ \\
        Adjoint Matching & 40 & $181.929 \pm 0.180$ \\
        RSM              & 10 & $8.976 \pm 0.020$ \\
        \nameshort       & 10 & \underline{$8.865 \pm 0.012$} \\
        \nameshort       & 5  & $\mathbf{6.566 \pm 0.004}$ \\
        \bottomrule
    \end{tabular}
\end{table}
\paragraph{Time to matched rewards.}
\Cref{fig:training_efficiency} reports reward improvement against
training GPU hours. At the annotated reward levels, \nameshort
requires $8\times$ and $23\times$ less training compute than
Adjoint Matching and VGG-Flow, respectively, for PickScore, and
$13\times$ and $16\times$ less for HPSv2.1. These comparisons concern the observed training budgets and
do not establish asymptotic convergence.

\begin{figure}[h]
    \centering
    \vspace{-25pt}
    \includegraphics[width=0.48\linewidth]{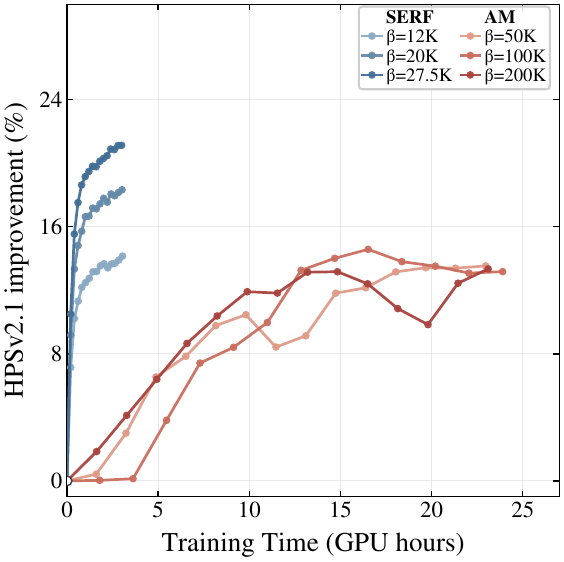}
    \caption{HPSv2.1 improvement on HPDv2 against training GPU-hours
    for SERF and Adjoint Matching. Each run uses 300 gradient updates;
    validation time is excluded.}
    \label{fig:am_training_progress}
\end{figure}

\paragraph{Adjoint Matching}
We further examine Adjoint Matching's training progress and sensitivity to the rollout discretization.
\Cref{fig:am_training_progress} reports reward improvement against training GPU-hours on a linear scale, complementing the convergence comparison in the main text. These results compare progress within the common budget of 300 updates.
Reward improvement largely plateaus after approximately 200
updates, supporting the 300-update budget for these settings.

\begin{figure}[h]
    \centering
    \includegraphics[width=\linewidth]{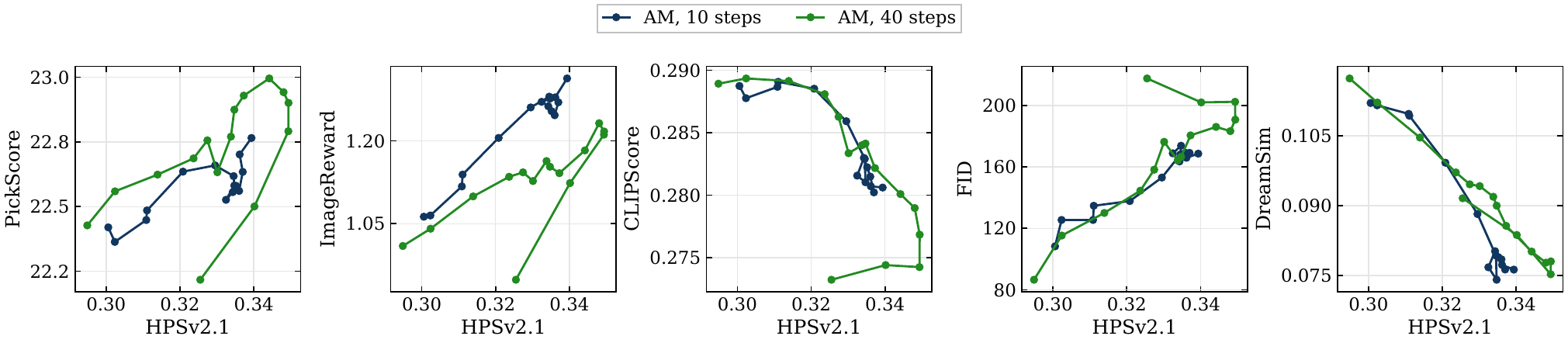}
    \caption{Reward--quality trade-offs on HPSv2.1 for Adjoint Matching
    with 10 or 40 rollout steps. Both runs use reward coefficient
    $\beta=10^6$ and seed 0. Points represent checkpoints evaluated
    every 20 updates through update 300, connected in training order.}
    \label{fig:am_rollout_steps}
\end{figure}

\Cref{fig:am_rollout_steps} compares 10 and 40 steps for both forward sampling and backward adjoint integration. The 40-step run reaches a higher peak HPSv2.1 score but deteriorates later in training. Across the remaining evaluation metrics, neither configuration shows a consistent advantage. Thus, quadrupling the integration steps does not consistently
improve AM's reward--quality trade-offs, despite increasing update
cost by approximately $2.5\times$ (\Cref{tab:update_time}).
These results argue against coarse rollout discretization being
the primary explanation for AM's weaker trade-offs in our comparison.

\clearpage
\subsection{Policy deviation}
\label{app:policy_deviation}

\begin{figure}[h]
  \centering
  \begin{subfigure}[t]{0.48\linewidth}
    \centering
    \includegraphics[width=\linewidth]
      {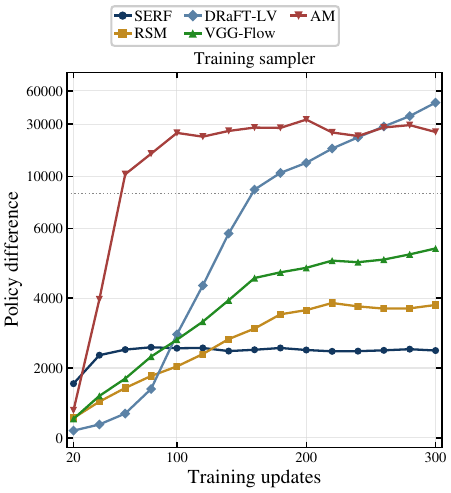}
    \caption{Evolution during training.}
    \label{fig:policy_deviation_training}
  \end{subfigure}\hfill
  \begin{subfigure}[t]{0.48\linewidth}
    \centering
    \includegraphics[width=\linewidth]
      {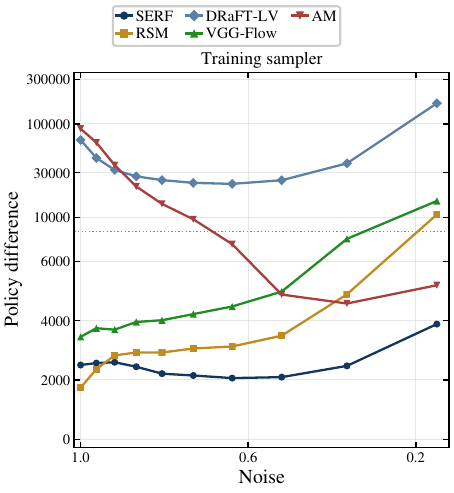}
    \caption{Noise-level profile at update 300.}
    \label{fig:policy_deviation_noise}
  \end{subfigure}
  \caption{Velocity deviation using each method's training rollout
  sampler. The diagnostic uses 20 sampling steps; training used ten.
  Deviation axes are linear below 7000 and logarithmic above,
  with the transition marked by a dotted line.}
  \label{fig:policy_deviation_rollout}
\end{figure}

\paragraph{Training evolution and noise profiles.}
\Cref{fig:policy_deviation_rollout} uses each method's training
rollout sampler: ODE for \nameshort, RSM, VGG-Flow, and DRaFT-LV;
memoryless SDE for AM. \nameshort's velocity deviation stabilizes
early. Separately, the saved reward evaluations first reach HPSv2.1
$0.36$ at update 160 for \nameshort, versus 240 for RSM, 260 for
VGG-Flow, and 280 for DRaFT-LV. AM remains below this reward level
within 300 updates.
At update 300, \nameshort also has the smallest measured deviation
in the mid-noise region ($\sigma \approx 0.5$--$0.8$).

\citet{jena2025elucidating} connect earlier denoising steps
to mode recovery and show improved reward--diversity trade-offs
by retaining more pretrained dynamics there.
\nameshort therefore achieves faster reward improvement with
substantially smaller and more stable changes to the pretrained
velocity field, especially at intermediate noise levels.
This supports preservation of pretrained dynamics as an explanation
for its stronger reward--quality trade-offs.

\begin{figure}[t]
  \centering
  \includegraphics[width=\linewidth]
    {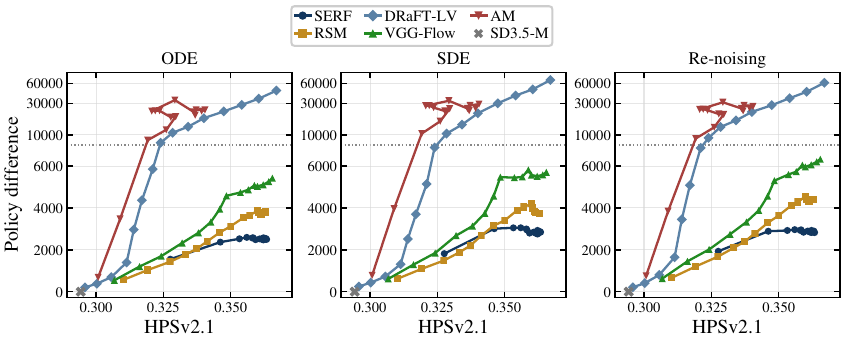}
  \caption{HPSv2.1 reward on HPDv2 versus velocity deviation
  measured on ODE, SDE, and re-noised states. Points represent
  checkpoints every 20 updates, connected in training order.
  Every panel uses the same saved ODE image rewards; the state
  sampler changes only the deviation measurement.
  The grey marker denotes the pretrained model.}
  \label{fig:reward_policy_deviation}
\end{figure}

\paragraph{Reward--deviation trade-offs.}
\Cref{fig:reward_policy_deviation} compares reward and deviation
separately for ODE, SDE, and re-noised states. Near HPSv2.1 $0.36$,
\nameshort achieves comparable rewards with less deviation than
RSM, VGG-Flow, and DRaFT-LV. The advantage persists across ODE, SDE, and re-noised diagnostic
states, establishing a robust improvement in the measured
reward--deviation trade-off.
Thus, \nameshort obtains substantial reward gains while better
preserving the pretrained velocity field, complementing the
reward--quality results in~\S\ref{sec:mainresults}.

\begin{figure}[t]
    \centering
    \includegraphics[width=\linewidth]{
        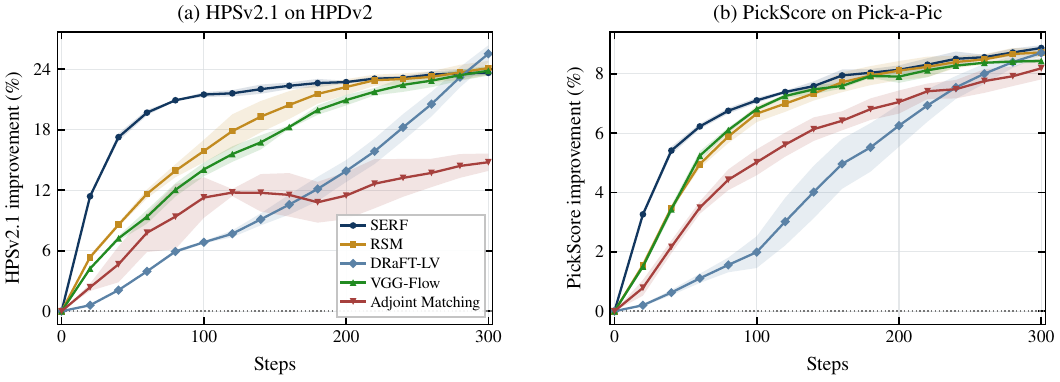
    }
    \caption{Reward improvement relative to the pretrained model
    against optimizer updates, optimizing HPSv2.1 on HPDv2
    (left) and PickScore on Pick-a-Pic (right).
    \nameshort achieves faster early reward improvement
    under both objectives.}
    \label{fig:reward_per_update}
\end{figure}

\paragraph{Reward improvement per update.}
\Cref{fig:reward_per_update} compares reward improvement against
optimizer updates, separating progress per update from the cost
of each update. \nameshort achieves faster initial reward
improvement under both training objectives. Together with
\Cref{fig:policy_deviation_training}, this shows that rapid
reward improvement is compatible with early stabilization
of velocity deviation: reward continues to improve without
sustained growth in deviation from the reference.

\subsection{Endpoint-gradient alignment}
\label{app:gradient_alignment}

We assess whether replacing the trajectory Jacobian by the identity
preserves the reward-gradient direction, and compare SERF-style
conditional endpoint guidance with RSM.
We focus on directional alignment, since a global guidance rescaling
can be absorbed into $\beta$, while the tested time-dependent
guidance weightings yield similar trade-offs
(\S\ref{app:time_weighting}).

We evaluate at initialization using frozen SD3.5-M and HPSv2.1.
Since the model coincides with the reference policy, the control
cost vanishes and its fixed-policy value reduces to
$V_t^{\mathrm{ref}}(x)=
\mathbb E_{\mathrm{ref}}[r(X_1)\mid X_t=x]$,
isolating the reward-gradient approximation.
We use eight prompts with four seeds each, 40 Euler steps,
and FP32 computations.
At eight denoising times, we retain ODE states and construct
independent re-noisings of each endpoint.
From each starting state, we sample 32 SDE continuations and
compute both the endpoint reward gradient and its full pullback,
differentiating through the sampler.

\begin{figure}[t]
    \centering
    \begin{subfigure}[t]{0.48\linewidth}
        \centering
        \includegraphics[width=\linewidth]
            {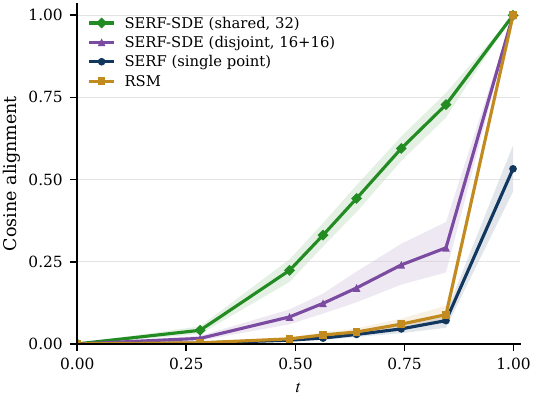}
        \caption{Conditional-gradient alignment.}
        \label{fig:conditional-endpoint-approximation}
    \end{subfigure}\hfill
    \begin{subfigure}[t]{0.48\linewidth}
        \centering
        \includegraphics[width=\linewidth]
            {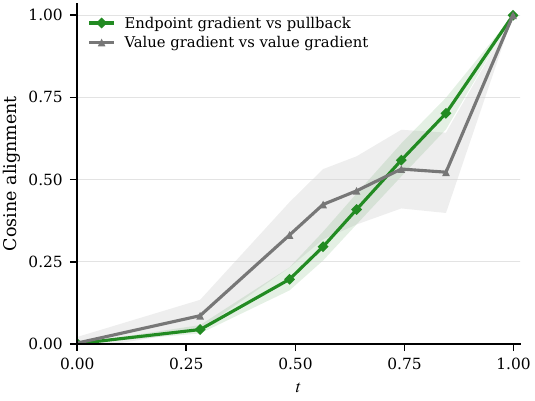}
        \caption{Endpoint--pullback agreement.}
        \label{fig:independent-endpoint-check}
    \end{subfigure}
    \caption{Left: alignment with the forward value-gradient estimate,
    using shared (32 paths) or disjoint ($16+16$ paths) continuations
    for the averaged endpoint-gradient and pullback estimates.
    Right: paired endpoint--pullback alignment and agreement between
    disjoint 16-path value-gradient estimates.
    Bands: 95\% prompt-bootstrap intervals.
    Final points are projection-only controls.}
\end{figure}

\Cref{fig:conditional-endpoint-approximation} reports cosine
alignment with the forward value-gradient estimate, obtained
by averaging full pullbacks.
Here, SERF-SDE denotes the endpoint-gradient surrogate under
the SDE state--endpoint coupling, before introducing independent
re-noising.
We estimate its conditional guidance by averaging endpoint
gradients over forward SDE continuations from each fixed
re-noised starting state.
The shared estimate uses the same 32 continuations for both
means; the disjoint estimate uses 16 continuations for each
mean with no overlap, testing alignment without shared
continuation noise.
Single-endpoint SERF and RSM instead use the original endpoint
gradient, compared with the value-gradient estimate at re-noised
and ODE states, respectively.

SERF-SDE achieves substantially stronger alignment than either
single-endpoint construction.
At $t\approx0.85$, alignment remains approximately $0.3$ with
disjoint continuations, compared with below $0.1$ for RSM and
single-endpoint SERF; the shared estimate reaches approximately
$0.7$.
The advantage therefore persists when guidance and the
value-gradient reference are estimated independently, even with
only 16 continuations per estimate.
This provides evidence that conditional averaging recovers an
informative reward-gradient direction without explicit
Jacobian transport.

\Cref{fig:independent-endpoint-check} provides complementary
diagnostics at re-noised starting states.
The green curve reports the mean cosine between each endpoint
gradient and its own full pullback.
The grey curve compares two value-gradient estimates obtained
from disjoint sets of 16 continuations.
This $16$-versus-$16$ control shows poor agreement at high noise,
indicating substantial Monte Carlo uncertainty in the reference
direction.
Low alignment with this reference at high noise therefore cannot
be attributed solely to the endpoint-gradient approximation;
the later-denoising comparisons are more informative as the
reference estimates become more reproducible.
Together, these diagnostics support the identity approximation
at the level of reward-gradient direction, particularly later
in denoising.
Even modest positive alignment is notable in the approximately
$65{,}000$ dimensional latent space.

Near the data endpoint, the SDE noise vanishes and the remaining
sampling map approaches the identity, so the other constructions
compare essentially the same gradient and approach unit alignment.
Single-endpoint SERF instead compares the original endpoint
gradient with the value-gradient estimate after independently
re-noising the endpoint and sampling forward again.
At the final evaluated time $t<1$, this perturbation remains
nonzero, so its alignment need not equal one.

Crucially, the single-endpoint SERF curve evaluates a sampled
training target, rather than the conditional-mean guidance
targeted by its regression loss.
Under the re-noising joint law $q$, population squared-loss
regression targets
$\mathbb E_q[\nabla r(X_1)\mid X_t=x]$ implicitly.
Although additional endpoints conditional on a fixed re-noised
state are not directly available, SERF can learn this average
from sampled training pairs without explicitly sampling that
posterior.
For RSM's deterministic ODE coupling, however, the intermediate
state and prompt determine the endpoint, so the corresponding
conditional average collapses to a single endpoint gradient.

The disjoint-SDE experiment demonstrates the potential benefit
of this averaging.
Although its continuations follow a different conditional law
from SERF's re-noising posterior, they provide evidence for the
averaging mechanism that SERF's regression exploits implicitly.
This supplies a concrete motivation for independent re-noising,
complementing its measured performance advantage in
\S\ref{app:independent_renoising}.

